\pdfoutput=1

\documentclass[11pt]{article}

\usepackage[]{ACL2023}

\usepackage{times}
\usepackage{latexsym}

\usepackage[T1]{fontenc}

\usepackage[utf8]{inputenc}

\usepackage{microtype}

\usepackage{inconsolata}

\usepackage{graphicx}
\usepackage{svg}
\usepackage{float}
\usepackage[shortlabels]{enumitem}
\usepackage{mathtools}
\usepackage{tabularx, booktabs, subcaption, multirow}
\usepackage{adjustbox}
\newcommand{\ra}[1]{\renewcommand{\arraystretch}{#1}}
\usepackage{hyperref}
\usepackage{longtable, multicol}
\usepackage{bm}

\usepackage{array}
\newcolumntype{C}[1]{>{\centering\arraybackslash}m{#1}}

\usepackage{pdfpages}

\usepackage{tcolorbox}

\usepackage{amsmath}
\newcommand\norm[1]{\left\lVert#1\right\rVert}

\usepackage{alphalph}
\usepackage{tikz}

\usepackage{subcaption}

\makeatletter
\newcommand\notsotiny{\@setfontsize\notsotiny\@viiipt\@ixpt}
\newcommand\notnotsotiny{\@setfontsize\notnotsotiny\@vipt\@viipt}
\makeatother

\title{Set-Theoretic Compositionality of Sentence Embeddings}

\author{
  Naman Bansal$^1$ \quad Yash Mahajan$^1$ \quad Sanjeev Sinha$^1$ \quad Santu Karmaker$^2$ \\
  $^1$Auburn University \quad $^2$University of Central Florida \\
  \texttt{\{nzb0040,yzm0034,sks0099\}@auburn.edu, santu@ucf.edu}
}

\begin{document}
\maketitle

\begin{abstract}
Sentence encoders play a pivotal role in various NLP tasks; hence, an accurate evaluation of their compositional properties is paramount. However, existing evaluation methods predominantly focus on goal task-specific performance. This leaves a significant gap in understanding how well sentence embeddings demonstrate fundamental compositional properties in a task-independent context. Leveraging classical set theory, we address this gap by proposing six criteria based on three core ``set-like'' compositions/operations: \textit{TextOverlap}, \textit{TextDifference}, and \textit{TextUnion}.  
We systematically evaluate $7$ classical and $9$ Large Language Model (LLM)-based sentence encoders to assess their alignment with these criteria. Our findings show that SBERT consistently demonstrates set-like compositional properties, surpassing even the latest LLMs.  
Additionally, we introduce a new dataset of $\sim192$K samples designed to facilitate future benchmarking efforts on set-like compositionality of sentence embeddings.

\end{abstract}

\section{Introduction}
\label{sec:introduction}

In the realm of NLP, acquiring meaningful sentence representations/embeddings is a fundamental pursuit. Over the past few years, researchers have spent significant research efforts to develop powerful sentence embeddings by leveraging both classical encoding models like Sentence-Bert (SBERT)~\citep{reimers-gurevych-2019-sentence}, Universal Sentence Encoder (USE)~\citep{cer-etal-2018-universal}, etc., as well as Large Language Models (LLMs) such as GPT-3~\citep{brown2020language} and LLaMA~\citep{touvron2023llama}. These sentence encoders have demonstrated their utility across various downstream tasks, including \textit{Semantic Similarity Estimation}, \textit{Text Classification}, and \textit{Information Retrieval }\citep{hamann2019hamming,muennighoff2022sgpt,gupta2023improved}. However, while task-specific performance has been extensively studied, there remains a gap in understanding how well these embeddings capture fundamental semantic compositionality in a task-independent context.

Semantic compositionality—the principle that the meaning of a complex expression is determined by the meanings of its parts and the rules used to combine them—is an essential aspect of natural language understanding. While modern sentence encoders have achieved notable success in estimating basic semantic similarity, their ability to capture compositional semantics in more complex scenarios remains understudied. To address this gap, we build upon the work of \citet{karmaker2018sofsat}, who introduced set-theoretic compositional operators for text analysis. More specifically, we investigate three  ``set-like'' sentence level compositions in this work: \textit{TextOverlap}, \textit{TextDifference}, and \textit{TextUnion}, analogous to the classic set operators: \textit{Intersection}, \textit{Difference}, and \textit{Union}, respectively. These operations offer a structured way to analyze how semantic content is transformed and combined, providing a systematic framework for evaluating the compositional properties of sentence encoders through a set-theoretic lens.

This work proposes six criteria derived from the aforementioned ``set-like'' operations in order to evaluate a set of novel compositional properties of sentence embeddings. We systematically analyze 16 sentence encoders, including both classical models such as Sentence-BERT (SBERT)\citep{reimers-gurevych-2019-sentence} and Universal Sentence Encoder (USE)\citep{cer-etal-2018-universal}, and LLMs like GPT-3~\citep{brown2020language} and LLaMA~\citep{touvron2023llama}. Our experiments aim to determine how well these encoders capture complex semantic transformations, where the transformations are based on set-inspired compositional operations. We focus exclusively on sentence-level analysis, deferring document-level analysis as future work. For evaluation, we introduce a benchmark dataset of $\sim$\textbf{$\mathbf{192}$K} synthetic samples simulating sentence-level set-like compositions. This dataset is designed to facilitate the evaluation of set-like compositionality across various sentence encoding models and encourage future research in this area\footnote{All data and annotations will be released upon acceptance.}. 

\section{Related Works}
\label{sec:related_works}

Numerous techniques have been proposed to generate embeddings for sentences, ranging from early unsupervised methods to modern transformer-based approaches.
\citet{le2014distributed} introduced Doc2Vec, an unsupervised technique that generates embeddings based on variable-length textual segments, providing unique representations for each paragraph. 
Following this, sentence embeddings were explored through auto-encoders \citep{hill-etal-2016-learning,hu2017toward,montero-etal-2021-sentence}, and methods based on predicting surrounding sentences given a target sentence \citep{kiros2015skip,logeswaran2018an}. \citet{pagliardini-etal-2018-unsupervised} extended the word2vec approach \citep{mikolov2013distributed} by incorporating n-gram embeddings, achieving robust results in unsupervised contexts.

More recent approaches have adopted contrastive objectives, leveraging different perspectives of the same sentence or document through data augmentation or multiple model copies~\citep{zhang-etal-2020-unsupervised,giorgi-etal-2021-declutr,wu2020clear,meng2021coco,carlsson2021semantic,kim-etal-2021-self,yan-etal-2021-consert}. 
Furthermore, supervised sentence embeddings, which achieve a better performance than their unsupervised counterparts, have garnered increasing attention. \citet{conneau-etal-2017-supervised} proposed fine-tuning of Siamese models on Natural Language Inference (NLI) datasets, a methodology extended by subsequent sentence encoding techniques~\citep{cer-etal-2018-universal, reimers-gurevych-2019-sentence}.
Another direction focused on regularizing embeddings to mitigate representation degeneration, resulting in substantial improvements over pre-trained language models \citep{li-etal-2020-sentence, Su2021WhiteningSR, huang-etal-2021-whiteningbert-easy}.
Furthermore, LLMs such as GPT-3 \citep{brown2020language}, LLaMA \citep{touvron2023llama}, and BLOOM \citep{workshop2022bloom} have showcased superior performance over classical models, instigating more interest in studying LLM-induced embeddings~\cite{mahajan-etal-2024-align, freestone2024word}. However, their extensive computational requirements pose accessibility challenges to many users.

In terms of compositionality, earlier research has demonstrated compositional properties at the word \citep{mikolov2013distributed, pennington2014glove} and phrase \citep{liu2022representations, yu2020assessing, dankers-etal-2022-transformer} level. Recently, sentence compositionality—the ability to combine simple sentences into larger, coherent meanings—has emerged as a critical area of exploration~\cite {zhu-de-melo-2020-sentence}. Recent work by \citet{huang-etal-2023-bridging} introduced InterSent, an end-to-end framework supporting compositional sentence operations in the embedding space.
Additionally, \citet{liu2024setcse} proposes a set-theory-based retrieval framework that uses contrastive loss to enhance sentence embeddings for complex information retrieval tasks.

Our work differs from the previous work in three ways: 1) We analyze sentence encoders through a novel lens inspired by classical set theory, focusing on compositional semantics using set-theoretic operations. More specifically, we build on the previous work by~\citet{karmaker2018sofsat} in order to evaluate sentence embeddings through three set-like operations: \textit{TextOverlap}, \textit{TextDifference}, and \textit{TextUnion}; 2) Our framework operates in a task-independent setting, offering a broader understanding of sentence embeddings beyond traditional downstream task-based performances; and 3) We evaluate nine LLM-induced embeddings, a relatively under-explored area, alongside seven classical models, providing a comprehensive comparison of their ``set-like'' compositional abilities.


\section{Background on Set-like Operators}
\label{sec:set_operators}


\begin{figure*}[!t]
    \centering
    \includegraphics[width=0.93\textwidth, keepaspectratio=true]{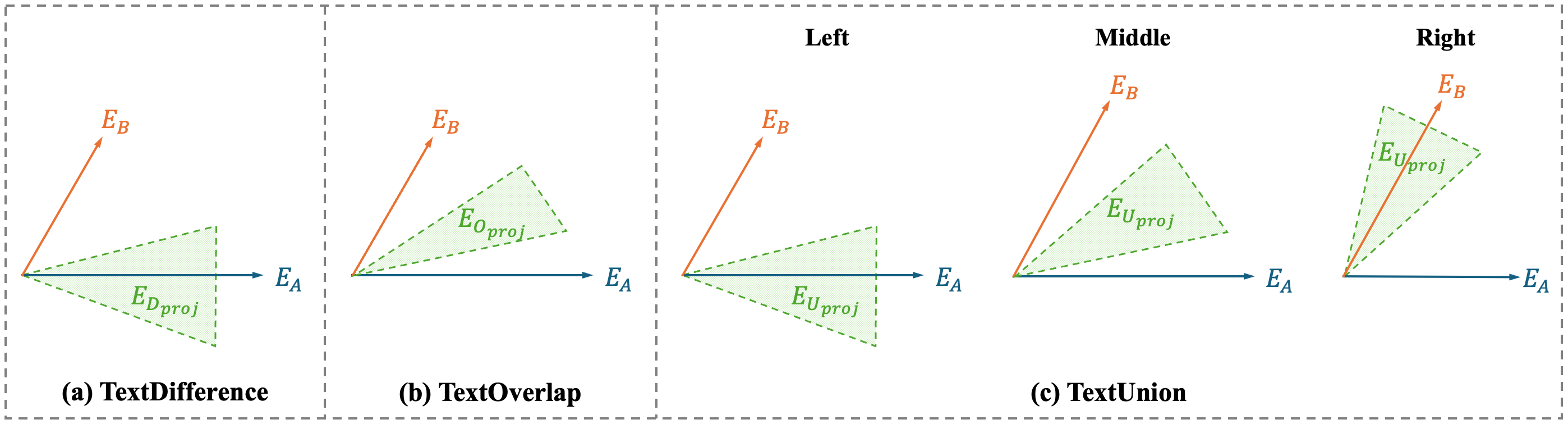}
    \vspace{-1.5mm}
    \caption{Expected results of the projection of the embedding of the TextOverlap, TextDifference and TextUnion operators onto the plane of the embeddings of the input sentences $A$ and $B$.
    Refer to Section \ref{sec:hypothesis} for details. 
    }
    \vspace{-3.5mm}
    \label{fig:projection_expectation}
    
\end{figure*}


\citet{karmaker2018sofsat} originally proposed SOFSAT, a set-like operator-based framework for composing text at both document and sentence levels. However, the scope of this paper is limited to sentence-level compositionality only.
To illustrate these operators, consider a pair of sentences, $\textbf{A}$: \textit{``Emma enjoyed her walk in the park and took many photographs.''}, and $\textbf{B}$: \textit{``Emma walked in the park and admired the blooming flowers.''} Now, we explain the set-like operators using this example:

\smallskip
\noindent \textbf{TextOverlap:} 
The TextOverlap operator represents the overlapping information between two input sentences. It can be formally defined as $(\{A, B\}, O)$, where $\{A, B\}$ is an unordered pair of the input sentences, and $O$ denotes their corresponding overlap in terms of shared semantic content. This operator is commutative, which means that $TextOverlap(A, B)$ is equivalent to $TextOverlap(B, A)$, reflecting the inherent symmetry of the intersection operation in set theory when applied to natural language sentences as well.
In our example, the expected overlap sentence would be \textit{``Emma walked in the park''.}

\vspace{1mm}
\noindent \textbf{TextUnion:} 
Similar to the mathematical union operator, the TextUnion operator
aims to combine the semantic content of the two input sentences into a single, coherent representation, capturing the collective information they convey.
It can be formally defined as $( \{ A, B \}, U)$, where the unordered pair $\{ A, B \}$ denotes the input sentences, and the output $U$ represents their corresponding ``union''. 
This concept resembles the notion of \textit{Sentence Fusion} introduced by \citet{marsi-krahmer-2005-explorations}. 
Similar to the set-theoretic union, the order of the input sentences is irrelevant.
In our example, the output ``union'' sentence would be \textit{``Emma admired the blooming flowers and took a lot of photographs while enjoying her walk in the park''.}

\vspace{1mm}
\noindent \textbf{TextDifference:} 
The TextDifference operator identifies the information present in one sentence but absent in the other.
Analogous to its set theory counterpart, it is non-commutative, allowing the specification of left and right ``difference'' operators. Formally, TextDifference can be denoted as $((A, B), D)$, where the ordered pair $(A, B)$ corresponds to the input sentences and $D$ corresponds to the output containing the information present in $A$ but not in $B$. This non-commutativity property implies that TextDifference$(A, B) \neq \text{TextDifference}(B, A)$, as the order of the input sentences determines which sentence serves as the reference for the difference computation.
In our example, the output TextDifference would be \textit{``Emma enjoyed her walk and took many photographs''} (LeftDifference) and \textit{``Emma admired the blooming flowers in the park.''} (RightDifference).\footnote{These examples are for illustration purposes only and haven't been utilized in our study.}
Throughout this work, when employing the TextDifference operator, the left difference is implied by default (without loss of generality) unless stated otherwise.

\section{Semantic Compositionality Criteria}
\label{sec:hypothesis}
We propose the following six criteria (C1 - C6) to evaluate how well various sentence encoders capture set-like compositional properties. 

\subsection{\textbf{Criteria for TextOverlap}}
\phantomsection
\label{para:h1}
\noindent \textbf{C1}:
Consider the tuple $ ( \{ A, B \}, O) $ where the unordered pair $\{ A, B \} $ are the input sentences, and $O$ is the reference overlap sentence. 
We argue that an effective sentence encoder, capable of capturing set-like semantic operations, would yield embeddings such that the similarity of the sentence pairs $(A, O)$ and $(B, O)$ will be equal to or higher than the similarity of the pair $(A, B)$. 
More specifically, $Sim(A, O) - Sim(A, B) \ge \epsilon_{O_1} $ (denoted as condition $C_{O_1}$), where $\epsilon_{O_1}$ denotes a non-negative expected minimum margin.  
Similarly, $Sim(B, O) - Sim(A, B) \ge \epsilon_{O_2} $ (denoted as condition $C_{O_2}$). 
The underlying rationale is rooted in the expectation that a sentence encoder should adhere to this property because sentence $O$ encapsulates solely the information shared between $A$ and $B$, whereas  $A$ and $B$ may each contain additional, unique information not present in the other.


\smallskip
\phantomsection
\label{para:h2}
\noindent \textbf{C2}: Consider again the tuple $(\{ A, B \},O)$. C2 postulates that the projection of the embedding of sentence $O$ onto the plane defined by the embeddings of input sentences $A$ and $B$ (details in Appendix~\ref{app:sec:projection}) will lie somewhere in the ``middle'' of the input embeddings as shown in Figure~\ref{fig:projection_expectation}b.
Let $E_O$ be the embedding of sentence $O$ and $E_{O_{proj}}$ its projection onto the plane formed by the input sentences' embeddings, $E_A$ and $E_B$. 
By `somewhere in the middle', we mean that $E_{O_{proj}}$ lies between $E_A$ and $E_B$ such that $\angle(E_{O_{proj}}, E_A) + \angle(E_{O_{proj}}, E_B) \approx \angle(E_A, E_B)$.
This relation intuitively places $E_{O_{proj}}$ between $E_A$ and $E_B$ based on their (acute) angular distances. 

\subsection{Criteria for TextDifference}
\phantomsection
\label{para:h3}
\noindent \textbf{C3}: Consider the tuple $ ( (A, B ), D) $ where the ordered pair $(A, B)$ are the input sentences and $D$ is the reference ``difference'' sentence.
We postulate that the similarity between the pair of sentences $(A, D)$ and $(A, B)$ should be greater than the similarity of the pair $(B, D)$. 
Mathematically, we expect $Sim(A, D) - Sim(B, D) \ge \epsilon_{D_1} $ (denoted as condition $C_{D_1}$), where $\epsilon_{D_1}$ denotes a non-negative expected minimum margin.  
Similarly, $Sim(A, B) - Sim(B, D) \ge \epsilon_{D_2} $ (denoted as condition $C_{D_2}$). 
The underlying argument is that sentence $D$ is the TextDifference of ordered pair $(A, B)$, and its embedding should ideally contain no information from $B$, making their embeddings very different. 

\smallskip
\phantomsection
\label{para:h4}
\noindent \textbf{C4}: 
Consider again the ``Difference'' sample $ ( (A, B ), D) $. 
This criterion posits that the algebraic difference of embeddings of the input sentences ($E_{A} - E_{B}$, denoted as $\Delta E_{A,B}$) would exhibit greater similarity to the embedding of ``Difference'' sentence $D$ than to the embedding of sentence $B$.
Specifically, we anticipate $Sim(\Delta E_{A,B}, E_D) - Sim(\Delta E_{A,B}, E_{B}) \ge \epsilon_{D_3}$ where $\epsilon_{D_3}$ denotes the non-negative expected minimum margin. 
The key intuition is that the algebraic difference between the input embeddings could serve as a potential approximation of the embedding of the ``Difference'', reflecting whether the model captures the semantic transformation.
This hypothesis is inspired by the analogy task, where an abstract concept (in this case, TextDifference) is represented as the linear relationship between two embeddings \citep{mikolov2013distributed,wang2019evaluating,wijesiriwardene-etal-2024-relationship}.

\smallskip
\phantomsection
\label{para:h5}
\noindent\textbf{C5}:
Considering tuple $ ( (A, B ), D) $ again, this criterion postulates that the projection vector of the embedding of sentence $D$ onto the plane defined by the embeddings of sentences $A$ and $B$ will be bounded by a small angle around the embedding of sentence $A$. 
Specifically, $\angle (E_A, E_{D_{proj}}) < \theta_D $ where $\theta_D$ is the expected margin and $E_{D_{proj}}$ is the projection of embedding vector of sentence $D$ on the plane of embeddings of sentences $A$ and $B$. This criterion is illustrated in 
Figure~\ref{fig:projection_expectation}a.

\subsection{Criteria for TextUnion}
\phantomsection
\label{para:h6}
\noindent \textbf{C6}: 
Consider the tuple $ ( \{A, B \}, U) $ where $U$ is the ``union''  of input sentences $A$ and $B$. 
Additionally, let us denote the projection of the embedding of $U$ onto the plane of the embeddings of the input sentences as $E_{U_{proj}}$.
We hypothesize that the projection can be either ``inside'' or ``outside'' the two input embeddings. 
Specifically, consider the following cases: 
\\
a) When the norm of  $E_A$ exceeds the norm of $E_B$, we expect the projection to be centred around the embedding of sentence $A$ i.e. $\angle (E_A, E_{U_{proj}}) < \theta_{U_1} $, as illustrated in Figure~\ref{fig:projection_expectation}c-(left)
\\
b) Similarly, when the norm of $E_B$ is larger than the norm of $E_A$, we expect the projection to be centered on $E_B$ i.e $\angle (E_B, E_{U_{proj}}) < \theta_{U_1} $  (Figure~\ref{fig:projection_expectation}c-right)
\\
c) When both the norms of the input embeddings are comparable, we expect the projection to lie in the  ``middle'' (Figure~\ref{fig:projection_expectation}c-middle). Here $\theta_{U_1}$ and $\theta_{U_2}$ denote expected margins.

\section{Dataset}
\label{sec:dataset}

The availability of comprehensive datasets for set-like operators, such as TextOverlap, TextDifference and TextUnion, is limited. 
With regards to the TextOverlap, \citet{bansal-etal-2022-semantic} presented the first benchmark dataset in the news domain, sourced from \href{https://www.allsides.com}{AllSides.com}. 
However, this dataset predominantly consists of entire documents and does not encompass samples corresponding to TextDifference and TextUnion operators.

\begin{figure}[!t]
    \centering
    \includegraphics[width=
    0.9\columnwidth, keepaspectratio=true
    ]{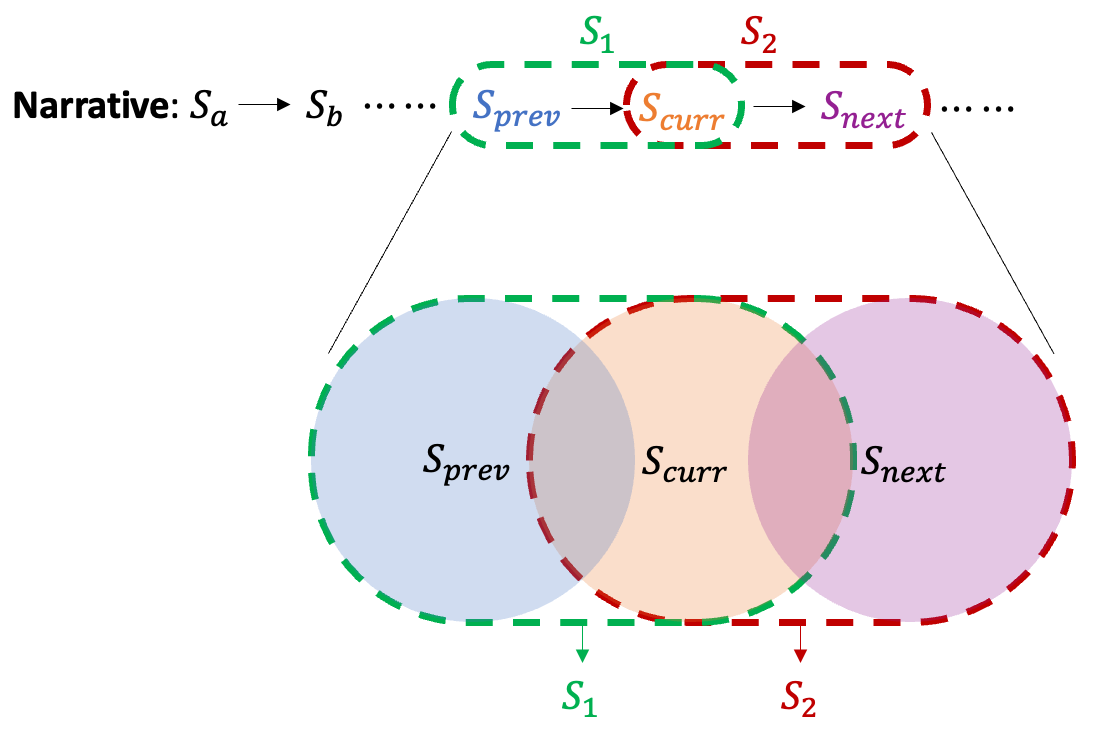}
    \vspace{-3mm}
    \caption{
    Illustration of synthetic data generation for various set-like operators. 
    Three consecutive sentences, denoted as $S_{prev}$, $S_{curr}$, and $S_{next}$, are extracted from a document. 
    Subsequently, sentence pairs $(S_{prev}, S_{curr})$ and $(S_{curr}, S_{next})$ are fused or combined to form $S_1$ and $S_2$ respectively. 
    These five base sentences serve as the foundation for creating samples for different set operators, as elaborated in Section \ref{sec:dataset} and qualitative samples are presented in Tables \ref{tab:qualitative_input_samples}, \ref{tab:qualitative_overlap}, \ref{tab:qualitative_union}, \ref{tab:qualitative_difference1} and \ref{tab:qualitative_difference2}.
    }
    \label{fig:dataset_creation}
    \vspace{-2mm}
\end{figure}

To address this gap, we leverage excerpts from the CNN-DailyMail dataset \citep{see-etal-2017-get} and synthetically generate samples to enable granular analysis at the sentence level. 
Our methodology leverages the fusion (``union'') of sentences to create diverse samples for various set-like operators.
Computationally, given a document $N$, we extract three consecutive sentences: $S_{prev}$, $S_{curr}$ and $S_{next}$ from it. Next, we prompt the GPT3.5~\cite{GPT3} (detailed in Appendix \ref{app:sec:fusion_prompt}) to ``unionize''/fuse information from pairs of sentences ($S_{prev}, S_{curr}$) and ($S_{curr}, S_{next}$), yielding synthesized sentences $S_1$ and $S_2$, respectively (refer to Figure \ref{fig:dataset_creation} for a visual representation)\footnote{We selected three consecutive sentences to generate our samples, thereby increasing the likelihood of a natural connection between the input sentences, where set-theoretic properties would be more intuitive and meaningful. 
Applying set operations to completely unrelated random sentences, while theoretically valid, would not yield particularly insightful or interesting results.}.

These five base sentences $S_{prev}$,  $S_{curr}$, $S_{next}$, $S_1$ and $S_2$ (see Appendix Table \ref{tab:qualitative_input_samples} for a qualitative example) serve as the foundation for creating samples for different set operators:

\smallskip
\noindent \textbf{TextOverlap:} Synthetic samples are created in the form of $( \{S_1, S_2 \}, S_{curr} )$, with the unordered pair $\{S_1, S_2 \}$ representing the input sentences ($\{A, B \}$) and $S_{curr}$ as their overlap sentence ($O$). In total, we created $37,292$ TextOverlap samples (refer to Appendix Table \ref{tab:qualitative_overlap} for an example). 

\smallskip
\noindent \textbf{TextDifference:} 
TextDifference samples are stored as $((A,B),D)$, where the ordered pair $(A, B)$ are the input sentences and $D$ is the target ``difference'' sentence. Following this structure, we generated 3 separate synthesized samples:

\noindent 1) $((S_1, S_{prev} ), S_{curr} )$  where the output ``difference'' is $S_{curr}$ given the inputs $S_1$ and $S_{prev}$.\\
2) $((S_1, S_{curr}), S_{prev} )$ where $S_1$ and $S_{curr}$ are the input sentences and $S_{prev}$ is the target ``difference''.\\ 
3) $((S_1, S_2), S_{prev} )$ with $S_1$ and $S_2$ being the input sentences and $S_{prev}$ being the target ``difference''.

\smallskip
Similarly, we can create another set of three samples from the fused sentence $S_2$, i.e., $( (S_2, S_{curr} ), S_{next} )$, $( (S_2, S_{next} ), S_{curr} )$, $( (S_2, S_1 ), S_{next} )$ (Examples are shown in appendix Tables~\ref{tab:qualitative_difference1} and~\ref{tab:qualitative_difference2}). In constructing these difference samples, we need to ensure that there is little to no overlap between pairs $\{S_{prev}, S_{curr}\}$ and $\{S_{curr}, S_{next}\}$. 
To ensure this, we encoded the sentence pairs using various sentence encoder models, and only the pairs with a cosine similarity of less than 0.25 (determined empirically) were retained. 
In total, we generate $79,824$ samples.

\smallskip
\noindent \textbf{TextUnion:}
As previously stated, TextUnion can essentially be viewed as Text Fusion~\cite{marsi-krahmer-2005-explorations}. Thus, the fused sentences $S_1$ and $S_2$ we previously created can be used directly for TextUnion. More specifically, we get $(\{ S_{prev}, S_{curr} \}, S_1)$ and $(\{ S_{curr}, S_{next} \}, S_2)$ as TextUnion samples.  
In total, we get $74,582$ samples.  
Refer to Appendix Table \ref{tab:qualitative_union} for an example instance. Note that in generating the fusion sentences $S_1$ and $S_2$, we solely prompted the GPT-3.5 model without utilizing any embeddings.



\begin{figure}[!t]
    \centering
    \includegraphics[width=0.65\columnwidth, keepaspectratio=true]{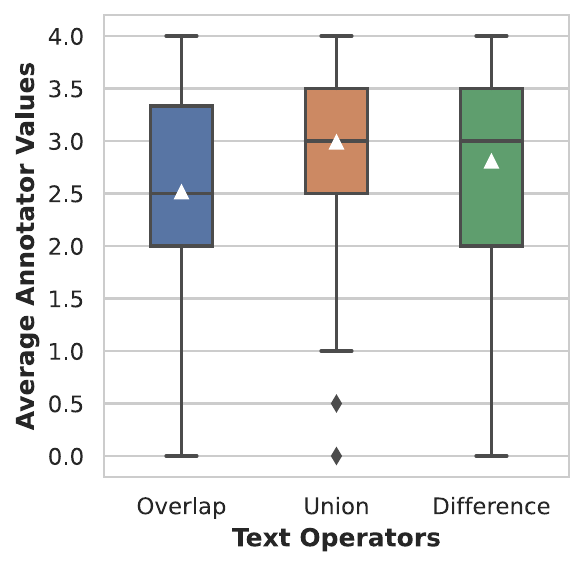}
    \vspace{-4mm}
    \caption{
    Box plot of human annotation scores for each operator. 
    Each score represents the average rating assigned by two or three annotators per sample. 
    The scores are based on human evaluations of $400$ synthetic samples, rated on a Likert scale from $0$ to $4$.
    The average annotation score for each operator is $>$ 2.5.
    }
    \vspace{-4mm}
    \label{fig:annotation_boxplot}
\end{figure}

\subsection{Dataset Quality and Human Evaluation}
\label{subsec:qualitative_analysis}

We engaged multiple human annotators, each with several years of NLP research experience, to assess the quality of the synthetic dataset. 
For interested readers, detailed annotation guidelines are provided in the Appendix~\ref{app:annotation_guidelines} and qualitative samples are shown in Tables \ref{tab:data_with_annotations_overlap}, \ref{tab:data_with_annotations_difference}, \ref{tab:data_with_annotations_union}.

For each operator, TextOverlap, TextDifference, and TextUnion, we randomly selected 400 samples. 
Each sample was evaluated by at least two annotators (with a maximum of three) using a Likert scale ranging from 0 to 4, where $0$ indicates ``Strongly Disagree'', $4$ indicates ``Strongly Agree'', and $2$ represents a neutral/satisfactory stance. 
In total, we collected $400 \times 3 \times 2.5 = 3000$ labels from human annotators, averaging $2.5$ annotators per sample. 
The average annotator score across operators was $2.77$, indicating that annotators generally found the data samples to be of ``satisfactory quality'' \footnote{Note that text generation is a challenging task, and an average score of $2.77$ (range [0,4]) reflects ``above satisfactory'' quality. For example, in the literature, the Rouge metric \citep{lin-2004-rouge} is commonly used to evaluate the quality of the generated samples, where $\sim 0.4$ score is considered ``good''.}.

As shown in Figure~\ref{fig:annotation_boxplot}, all operators received an average score exceeding 2.5, with TextUnion achieving an average rating of approximately 3.0 (reflecting the annotator's agreement on high-quality samples).
Notably, the median scores for both TextDifference and TextUnion are 3, indicating that annotators largely agreed on the correctness of the generated samples. 
The overall distribution of scores suggests consistent and favorable human evaluations, indicating a substantial level of confidence in the data quality.



\section{Experiment Settings and Results}
\label{sec:experiments}

In this study, we conduct a comparative analysis involving seven prominent classical sentence embedding models: 1) Universal Sentence Encoder (USE) \citep{cer-etal-2018-universal}, two Sentence-BERT (SBERT) \citep{reimers-gurevych-2019-sentence} variants 2) SBERT-L (\href{https://huggingface.co/sentence-transformers/all-mpnet-base-v2}{all-mpnet-base-v2}), currently their best \href{https://www.sbert.net/docs/sentence_transformer/pretrained_models.html}{performing variant}, 3) SBERT-mini (\href{https://huggingface.co/sentence-transformers/all-MiniLM-L6-v2}{all-MiniLM-L6-v2}), which strikes an optimal balance between performance and speed,  4) InferSent \cite{conneau-etal-2017-supervised}, 5) Language-Agnostic-SEntence Representation (LASER)~\cite{artetxe2019massively}, 6) SimCSE (\href{https://huggingface.co/princeton-nlp/unsup-simcse-bert-base-uncased}{unsup-simcse-bert-base-uncased}) ~\citep{gao-etal-2021-simcse}, and 7) RoBERTa (\href{https://huggingface.co/FacebookAI/roberta-base}{roberta-base}) \citep{liu2020roberta}, and nine Large Language Models (LLMs): 
1) GPT3-Ada2 \citep{brown2020language}, 
2) LLaMA2 (\href{https://huggingface.co/meta-llama/Llama-2-7b-hf}{Llama-2-7b-hf}) \citep{touvron2023llama}, 
3) LLaMA3 (\href{https://huggingface.co/meta-llama/Meta-Llama-3-8B}{Meta-Llama-3-8B}) \citep{llama3modelcard}, 
4) OLMo (\href{https://huggingface.co/allenai/OLMo-7B}{OLMo-7B}) \citep{groeneveld2024olmo}, 
5) OpenELM (\href{https://huggingface.co/apple/OpenELM-3B}{OpenELM-3B}) \citep{openelm}, 
6) Mistral (\href{https://huggingface.co/mistralai/Mistral-7B-v0.3}{Mistral-7B-v0.3}) ~\citep{mistral},
7) LLaMA3.2 (\href{https://huggingface.co/meta-llama/Llama-3.2-3B}{Llama-3.2-3B}) ~\citep{llama32modelcard}, 
8) Qwen (\href{https://huggingface.co/Qwen/Qwen2.5-7B}{Qwen2.5-7B}) ~\citep{qwen2.5}, and
9) Gemma (\href{https://huggingface.co/google/gemma-2-9b}{Gemma-2-9B}) ~\citep{gemma_2024}.
 While most of the LLMs are not specifically designed for sentence encoding, it is feasible to derive sentence embeddings from them.
 For comprehensive details, refer to Appendix \ref{app:sec:model_description}.

\subsection{Results}
\label{subsec:results}

We evaluated all models based on the six criteria described in section \ref{sec:hypothesis}. 
These criteria assess the extent to which sentence encoders capture set-like semantic compositions.
While we do not assert that a sentence encoder must meet all criteria to be effective across tasks, adherence to these criteria provides valuable insights into its ability to represent complex semantic transformations.

To compute the similarity between two sentences, we first compute the embeddings for each sentence using a sentence encoder.  Subsequently, we employ the standard cosine similarity as the primary metric. In addition to cosine similarity, the dot product between two embeddings is considered an alternative similarity measure, encompassing their respective norms. Furthermore, we calculate L1, L2, and \href{https://reference.wolfram.com/language/ref/NormalizedSquaredEuclideanDistance.html}{Normalized Euclidean Distance} (NED) between two embeddings as another measure of distance/similarity between the two embeddings. Our findings are detailed below:


\begin{table}[!t]
    \centering
    \ra{1.0}

    \begin{adjustbox}{width=0.9\columnwidth}
    \begin{tabular}{rcccc}
        \toprule
        \multirow{2}{*}{\textbf{Models}}
         & $\bf{C_{O_1}}$ = T & $\bf{C_{O_1}}$ = T & $\bf{C_{O_1}}$ = F & $\bf{C_{O_1}}$ = F \\
         & $\bf{C_{O_2}}$ = T & $\bf{C_{O_2}}$ = F & $\bf{C_{O_2}}$ = T & $\bf{C_{O_2}}$ = F \\
        \midrule
        SBERT-mini & $\bf{28.96}$ & $24.22$ & $25.24$ & $21.57$ \\
        LASER & $24.51$ & $25.47$ & $24.41$ & $25.61$ \\
        USE & $28.07$ & $24.72$ & $24.88$ & $22.33$ \\
        RoBERTa & $24.0$ & $26.11$ & $23.89$ & $26.0$ \\
        SBERT-L & $28.67$ & $24.47$ & $25.05$ & $21.81$ \\
        SimCSE & $26.91$ & $25.11$ & $24.69$ & $23.29$ \\
        InferSent & $24.66$ & $25.69$ & $24.26$ & $25.4$ \\

        \midrule
        GPT3 & $25.15$ & $25.8$ & $24.2$ & $24.85$ \\
        LLaMA2 & $24.57$ & $25.29$ & $24.55$ & $25.59$ \\
        Mistral & $24.77$ & $25.27$ & $24.64$ & $25.32$ \\
        LLaMA3 & $24.4$ & $25.13$ & $24.71$ & $25.76$ \\
        OLMo & $25.18$ & $25.01$ & $24.85$ & $24.96$ \\
        OpenELM & $23.93$ & $25.6$ & $24.19$ & $26.27$ \\
        LLaMA3.2 & $24.44$ & $25.53$ & $24.42$ & $25.6$ \\
        Qwen & $24.12$ & $25.85$ & $24.13$ & $25.91$ \\
        Gemma & $24.58$ & $24.98$ & $24.83$ & $25.6$ \\
        
        \bottomrule
    \end{tabular}
    \end{adjustbox}


        

    \vspace{-1mm}
    \caption{
    Percentage of samples adhering to various scenarios, ranging from instances where both conditions $C_{O_1}$ and $C_{O_2}$ (refer to section \ref{para:h2}) are satisfied to cases where neither condition is met.
    Notably, SBERT variants demonstrate superior efficacy in terms of the cosine metric, surpassing all six LLMs, with similar trends observed across the NED measure (refer Appendix \ref{tab:h1_analysis_appendix}).
    }
    \vspace{-3mm}
    \label{tab:h1_analysis}
\end{table}

\phantomsection
\label{para:h1_results}



\smallskip
\noindent \textbf{C1:}  
To evaluate criterion C1, we utilize TextOverlap samples ($(\{A, B\}, O)$) from our synthetic dataset. C1 expects that the average similarity between the pairs $(B, O)$ and $(A, O)$ would exceed that of the pair $(A, B)$, as expressed in terms of condition $C_{O_1}$ and  $C_{O_2}$ in section \ref{para:h1}. 
Table~\ref{tab:h1_analysis} presents the percentage of samples conforming to conditions $C_{O_1}$ and $C_{O_2}$ averaged across various values of $\epsilon_{O_1}$ and $\epsilon_{O_2}$ (see Appendix \ref{app:h1_results} for details). 
In particular, SBERT-mini and SBERT-L yield the highest percentage of samples that satisfy criterion C1. Furthermore, SBERT-mini conforms with criterion C1 more than SBERT-L despite its smaller size. Surprisingly, classical encoders, on average, satisfied C1 more often than LLM-based encoders.
This consistent trend is also observed for the NED measure (refer to Appendix Table \ref{tab:h1_analysis_appendix}).




\begin{table}[!t]
    \centering
    \ra{1.0}

    \begin{adjustbox}{width=0.90\columnwidth}
    
    \begin{tabular}{rcccc}
        \toprule
        \multirow{2}{*}{\textbf{Models}}
         & $\bf{C_{D_1}}$ = T & $\bf{C_{D_1}}$ = T & $\bf{C_{D_1}}$ = F & $\bf{C_{D_1}}$ = F \\
         & $\bf{C_{D_2}}$ = T & $\bf{C_{D_2}}$ = F & $\bf{C_{D_2}}$ = T & $\bf{C_{D_2}}$ = F \\
        \midrule
        SBERT-mini & $\bf{34.8}$ & $27.25$ & $22.33$ & $15.62$ \\
        LASER & $18.87$ & $26.58$ & $22.81$ & $31.74$ \\
        USE & $31.86$ & $27.38$ & $22.66$ & $18.1$ \\
        RoBERTa & $12.29$ & $24.53$ & $21.09$ & $ \bf{42.1}$ \\
        SBERT-L & $32.97$ & $27.36$ & $22.46$ & $17.21$ \\
        SimCSE & $26.66$ & $27.4$ & $23.01$ & $22.93$ \\
        InferSent & $16.68$ & $26.19$ & $22.3$ & $34.84$ \\
        \midrule
        GPT3 & $16.53$ & $26.31$ & $22.11$ & $35.04$ \\
        LLaMA2 & $18.31$ & $26.38$ & $22.58$ & $32.73$ \\
        Mistral & $18.06$ & $26.39$ & $22.5$ & $33.05$ \\
        LLaMA3 & $19.0$ & $26.51$ & $22.8$ & $31.69$ \\
        OLMo & $20.1$ & $26.82$ & $22.85$ & $30.23$ \\
        OpenELM & $15.43$ & $25.59$ & $21.98$ & $37.0$ \\
        LLaMA3.2 & $15.73$ & $25.9$ & $22.03$ & $36.34$ \\
        Qwen & $14.05$ & $25.29$ & $21.61$ & $39.06$ \\
        Gemma & $19.58$ & $26.69$ & $22.94$ & $30.79$ \\
        \bottomrule
    \end{tabular}
    \end{adjustbox}

    
    \vspace{-1mm}
    \caption{
    Percentage of samples meeting different scenarios, from both conditions $C_{D_1}$ and $C_{D_2}$ (refer to section \ref{para:h3}) being satisfied to neither being met.
    Interestingly, classical models exhibit significantly better alignment with criterion C3 compared to LLMs on average, with SBERT-mini demonstrating the highest percentage of samples adhering to C3.
    This trend persists across both cosine similarity and the NED measure (shown in Appendix Table \ref{tab:h3_analysis_appendix}). 
    }
    \vspace{-4mm}
    \label{tab:h3_analysis}
\end{table}

\phantomsection
\label{para:h2_results}


\smallskip
\noindent \textbf{C2:} Consider a TextOverlap sample $(\{A, B\}, O)$ with their embedidngs are represented by $E_A, E_B, E_O$, respectively. 
Additionally, we project $E_O$ onto the plane defined by $E_A$ and $E_B$, denoting the resulting projection as $E_{O_{proj}}$. 
Subsequently, we calculate the angles between the pairs $(E_A, E_B)$, $(E_A, E_{O_{proj}})$ and $(E_B, E_{O_{proj}})$. 
To facilitate a clearer interpretation, we normalize the angle between each pair such that the angle between
the pair $(E_A, E_B)$ is consistently set to 1.
Finally, we construct histograms depicting the distribution of angles between $(E_B, E_{O_{proj}})$. 

For the GPT3 model, the projection of the embedding of the overlap sentence $O$ lies in the ``middle'' of the embeddings of sentences $A$ and $B$ for $99.79 \%$ of the samples (Figure~\ref{fig:gpt_overlap}) and similar trends are observed for other embedding models, with results provided in the appendix (Figures \ref{fig:all_projection_results}). These results are consistent with criterion C2. 

\phantomsection
\label{para:h3_results}

\smallskip
\noindent \textbf{C3:}
To evaluate criterion C3, we utilize TextDifference samples $((A, B), D)$.
Here, we expect that the embedding of sentence $D$ will show higher similarity to the embedding of sentence $A$ compared to sentence $B$ (condition $C_{D_1}$). 
Similarly, we expect sentence $B$ to exhibit a greater similarity to $A$ than to sentence $D$ (condition $C_{D_2}$). 
To quantify this, we computed the average percentage of samples adhering to conditions $C_{D_1}$ and $C_{D_2}$ across various of $\epsilon_{D_1}$ and $\epsilon_{D_2}$, as detailed in Table \ref{tab:h3_analysis}.
Remarkably, classical models such as SBERT variants, USE, and SimCSE consistently outperform all LLMs, with SBERT-mini being the best performer.


\begin{figure*}[!t]
    \centering

    \begin{subfigure}[b]{0.3\textwidth}
         \centering
         \includegraphics[width=1.0\textwidth]{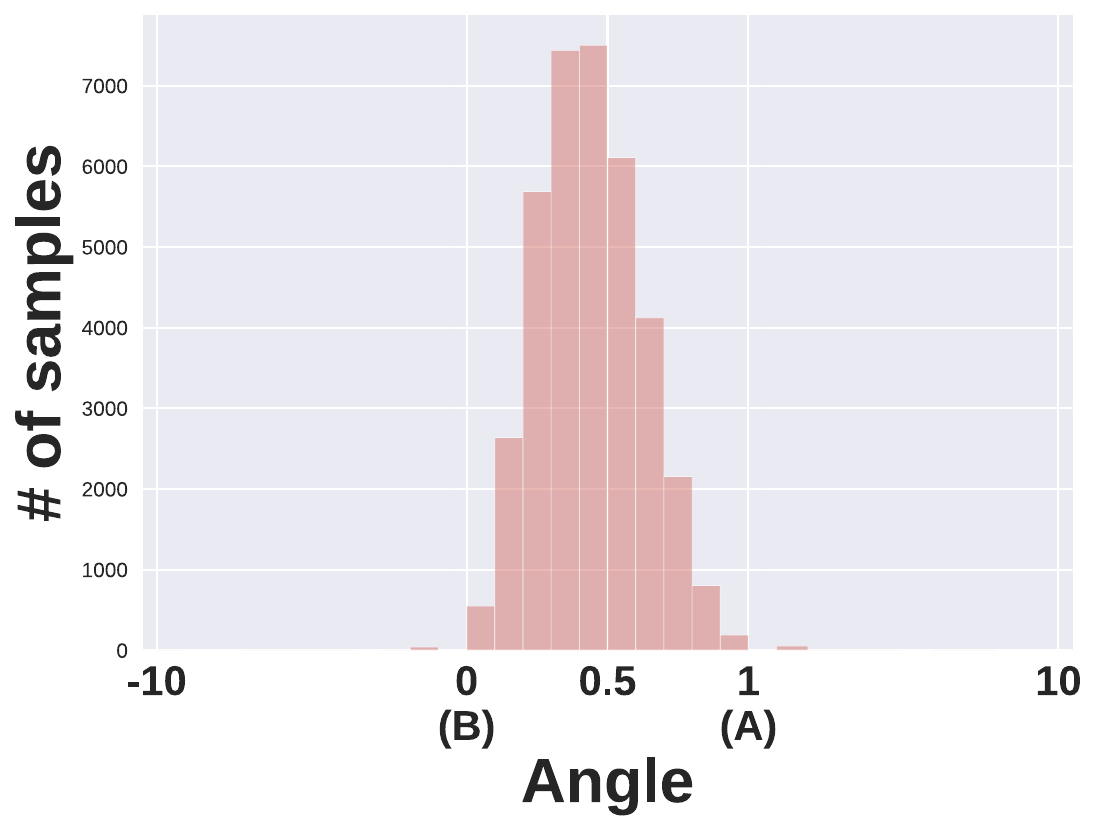}
         \caption{\textbf{TextOverlap:} The projection embedding mostly lie in the ``middle'' of the embeddings of the input sentences as described in criterion C2 \ref{para:h2_results}.}
         \label{fig:gpt_overlap}
     \end{subfigure}
    ~
     \begin{subfigure}[b]{0.3\textwidth}
         \centering
        \includegraphics[width=1.0\textwidth]{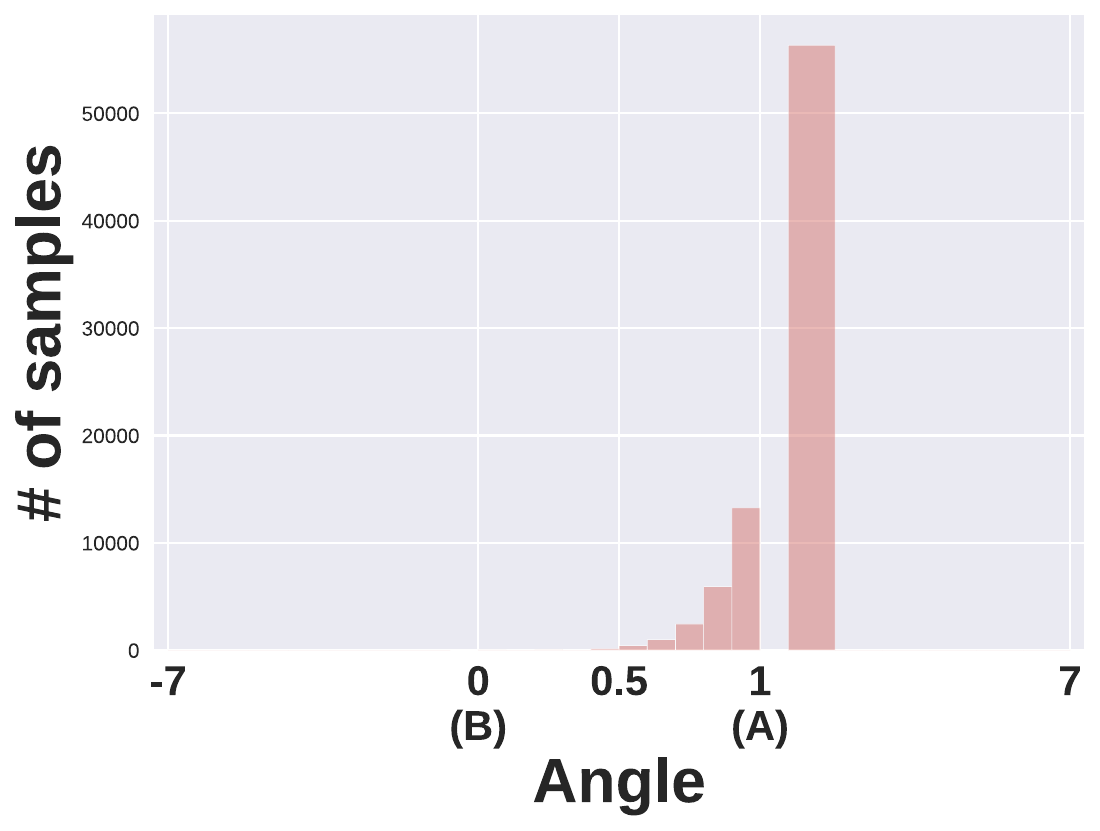}
         \caption{\textbf{TextDifference:} The projection embedding is mostly bounded by a small angle around the embedding of the input sentence $A$ (refer C5 \ref{para:h5_results}).
         }
         \label{fig:gpt_difference}
     \end{subfigure}
    ~
     \begin{subfigure}[b]{0.3\textwidth}
         \centering
         \includegraphics[width=\textwidth]{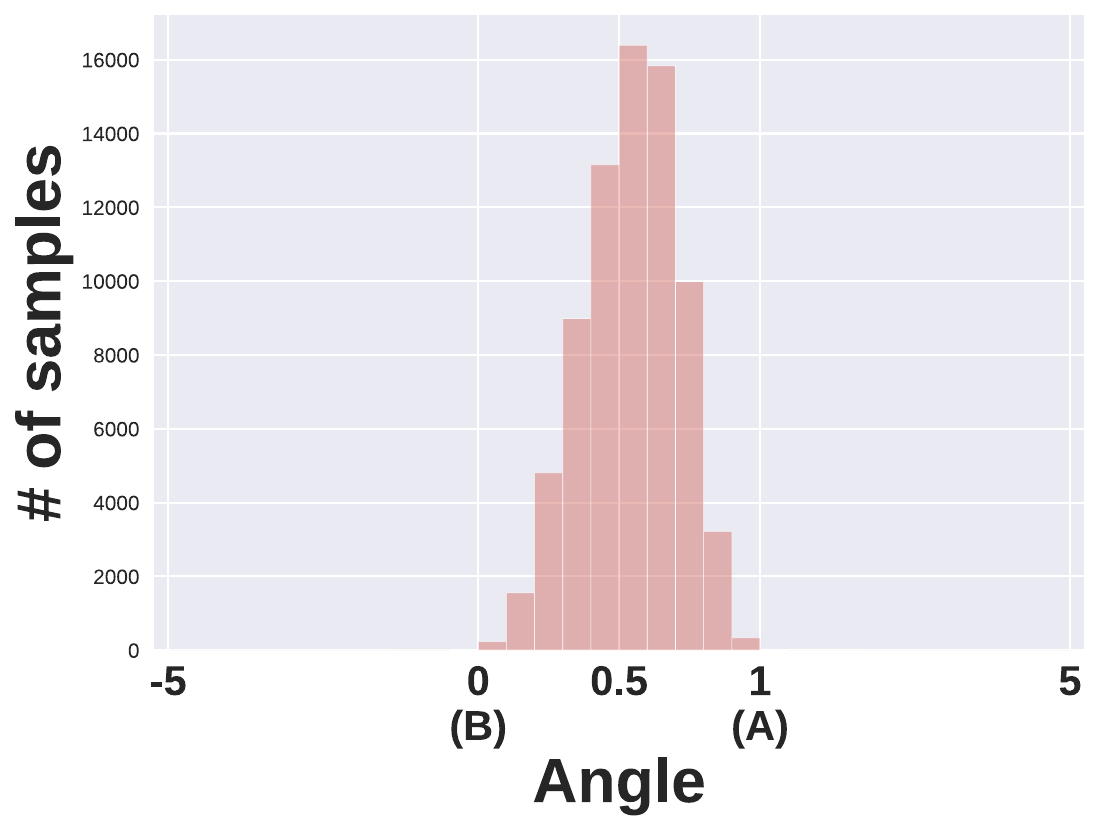}
         \caption{
         \textbf{TextUnion:} The projection embedding lies mostly in the ``middle'' of the input sentence embeddings (see \ref{para:h6_results}, \ref{fig:projection_expectation}c-middle). 
         }
         \label{fig:gpt_union}
     \end{subfigure}

    \vspace{-1mm}
    \caption{
    Histogram showing the angle between the target sentence  (projected onto the plane of embeddings from sentences  $A$ and $B$) and the embedding of sentence $B$ for GPT3.
    The target sentence embedding can be: 
    TextOverlap, TextDifference, TextUnion (Right).
    The results for other encoders are shown in Appendix Figure \ref{fig:all_projection_results}.
   We normalize this angle such that the angle between the embeddings of sentences $A$ and $B$ is consistently 1. 
    }
    \label{fig:gpt_projection}
    \vspace{-3.5mm}
\end{figure*}
\phantomsection
\label{para:h4_results}

\smallskip
\noindent \textbf{C4:} 
As outlined in section \ref{para:h4}, we postulate that the  algebraic difference of the embeddings of input sentences ($\Delta E_{A,B} = E_A - E_B$) should exhibit higher similarity to the embedding of the target difference sentence, $E_D$, 
compared to the embedding of sentence $B$ ($E_B$). 
To test this, we evaluate the percentage of samples where $(\Delta E_{A,B}, E_D)$ shows higher similarity (or lower distance) than $(\Delta E_{A,B}, E_{B})$, averaged across various $\epsilon_{D_3}$ values.
Once again, many classical models such as SBERT variants, USE, and SimCSE, demonstrate consistently better performance than all LLMs (see Table \ref{tab:h4_analysis}).
Moreover, the percentage of samples meeting criterion C4 (Table \ref{tab:h4_analysis}) consistently exceeds those meeting criterion C3 (Table \ref{tab:h3_analysis}) across models and evaluation metrics, suggesting that C4 may be relatively easier to satisfy compared to C3.

\begin{table}[!t]\small
    \centering
    \ra{1.0}

    \begin{adjustbox}{width=1.0\columnwidth}
    \begin{tabular}{rcc|rcc}
        \toprule
        \textbf{Classical} & \textbf{Cosine}  & \textbf{NED} & \textbf{LLMs} & \textbf{Cosine}  & \textbf{NED} \\
        \midrule
        SBERT-mini & $\bf{76.14}$ & $\bf{75.35}$ & GPT3 & $56.16$ & $50.5$ \\
        LASER & $59.95$ & $67.39$ & LLaMA2 & $59.02$ & $55.63$ \\
        USE & $74.01$ & $73.69$ & Mistral & $58.4$ & $54.74$ \\
        RoBERTa & $42.83$ & $36.75$ & LLaMA3 & $60.01$ & $57.05$ \\
        SBERT-L & $74.75$ & $73.87$ & OLMo & $62.21$ & $59.91$ \\
        SimCSE & $69.59$ & $67.69$ & OpenELM & $52.26$ & $46.53$ \\
        InferSent & $56.2$ & $64.27$ & LLaMA3.2 & $54.26$ & $49.61$ \\
        - & - & - & Qwen & $49.45$ & $43.22$ \\
        - & - & - & Gemma & $61.26$ & $59.01$ \\
        \bottomrule
    \end{tabular}
    \end{adjustbox}
    

    \vspace{-1mm}
    \caption{
    Analysis of the percentage of samples adhering to the conditions $C_{D_3}$ (refer to section \ref{para:h4}). 
    }
    \vspace{-3.5mm}
    \label{tab:h4_analysis}
\end{table}



\phantomsection
\label{para:h5_results}


\smallskip
\noindent \textbf{C5:} 
Similar to the second criterion (C2), we examine the projection of the embedding of the ``TextDifference'' with respect to the input sentences. 
Given the input sample, $((A, B), D)$, we compute their embeddings, denoted as $E_A$, $E_B$, and $E_D$, respectively, using the encoders.
Subsequently, we project the embedding $E_D$ onto the plane defined by the embeddings $E_A$ and $E_B$, and denote the projection vector as $E_{D_{proj}}$. 
Next, we calculate the angles between the pairs $(E_A, E_B)$, $(E_A, E_{D_{proj}})$ and $(E_B, E_{D_{proj}})$ and normalize the three angles such that the angle between the pair $(E_A, E_B)$ is consistently 1. Finally, we visualize the distribution of angles between $(E_B, E_{D_{proj}})$ using a histogram. 
As depicted in Figure \ref{fig:gpt_difference}, all samples are concentrated around the embedding of the sentence $A$ for the GPT3 model. 
Similar outcomes are observed for other encoders (details in Appendix, Figure \ref{fig:all_projection_results}).

\phantomsection
\label{para:h6_results}

\smallskip
\noindent \textbf{C6:} 
In this, we examine the projection of the embedding of ``TextUnion'' with regards to the input sentences $A$ and $B$, following a similar procedure as that for C2 and C5. 
The results for the GPT3 model are illustrated in Figure \ref{fig:gpt_union}, while those for the other encoders are provided in the Appendix Figures \ref{fig:all_projection_results}.
Surprisingly, across all the encoders, we consistently observe the anticipated scenario depicted in
figure \ref{fig:projection_expectation}c (middle) but not the other two.
Thus, we further conducted an analysis of the norms of the input embedding vectors by looking at their ratios. As depicted in Figure  \ref{fig:combined_norm_plots} in appendix, the histograms of these ratios are centered around 1 for all models, thus, confirming our expectations.

\section{Discussions and Final Words}
\label{sec:discussion}
In this paper, we conduct an exploratory but thorough analysis of various classical and LLM-based sentence encoders' compositional properties using a set-theory-inspired framework. To this end, we propose six intuitive evaluation criteria based on three set-like operators: \textit{TextOverlap}, \textit{TextDifference}, and \textit{TextUnion}. 
Our findings reveal that SBERT variants consistently align with the proposed compositionality criteria, while $1000\times$ larger LLMs like the GPT-3 and LLaMA variants don't align as much. Indeed, many classical
encoders demonstrate a higher alignment with the expected set-like compositional properties in comparison to the LLM-induced embeddings, an interesting finding given LLM's generally superior
performance on downstream benchmarks.
\textit{This result is perhaps due to the fact that classic encoders are exclusively trained to generate useful embeddings, as opposed to LLMs, which are decoder-only models and mainly trained for next-word prediction tasks.}
Finally and most importantly, this work provides a novel task-agnostic, intuitive framework for evaluating the set-like compositional properties of sentence encoders, offering a new evaluation scheme for analyzing/comparing them beyond traditional downstream task-based benchmarks:

\subsection{Implications and Future Applications}

\noindent \textbf{Quantifying ``Interpretability'':} As sentence embeddings represent a sentence as real-valued vector in a high-dimensional space, it is natural for humans to expect the common information in two sentences to be represented by a vector that ``lies somewhere in between'' the original sentence vectors. This mirrors how humans learn and interpret similarity in coordinate geometry. Therefore, such an alignment (if it indeed exists) can make sentence embeddings interpretable to humans as set theory is so intuitive to human minds. This way, our framework can help ``quantify'' the interpretability of a particular sentence encoder model and, hence, can enable quantitative comparison across different models in terms of their interpretability.

\noindent \textbf{Training with Joint Optimization of ``Interpretability'' and ``Accuracy'':} ``Interpretability'' and ``Accuracy" are two orthogonal axes of evaluation. Our results, in combination with the latest superior results of LLMs on popular benchmarks, basically suggests that we are yet to develop the ``ideal'' embedding that can optimize both ``Interpretability'' and ``Accuracy'' to a desired level, at least, when it comes to \textit{interpretability from a set-theoretic lens}. Therefore, one follow-up work could be to impose these set-theoretic properties as constraints while training and fine-tuning LLMs for benchmarking tasks. That may lead to an embedding model that optimizes for both set-theoretic ``Interpretability'' and ``Accuracy''.

\noindent \textbf{Choosing from Similar Performing Models:} In the case of multiple embedding models yielding similar accuracy, one can prefer a more ``interpretable model'' as a selection criterion.




\section*{Limitations}
\label{sec:limitations}

As previously noted, one major limitation of our study is its exclusive focus on the sentence-level operators in their current form. 
Another constraint pertains specifically to criterion C6. As detailed in section \ref{para:h6_results}, we solely encountered scenarios where the norms of the embeddings for both input sentences were comparable.
Therefore, further investigation is warranted to analyze other sentence encoders in cases where this is not the case.
Moreover, we mainly used linear distance/similarity measures such as cosine, dot product, euclidean distance, etc. in our study. However, these measures may not be suitable for all kinds of embeddings as the relationships are embedded in a non-linear and complex manner and their relationship can only be recovered with certain supervision and learnable transformation \citep{faruqui2016problems, wang-etal-2022-just}.
Furthermore, a key limitation of this study is that while the proposed criteria are grounded in set-theoretic operations, their ability to distinguish how different models capture set-like semantic compositions remains unexamined. Although these criteria provide valuable insights into set-like semantic compositions, their effectiveness in distinguishing how different models capture these operations has not been fully explored. As a result, their applicability as comprehensive evaluation tools is limited, and further research is needed to assess their generalizability across different models.

\bibliography{acl2023}
\bibliographystyle{acl_natbib}

\appendix

\section{Appendix}
\subsection{Sentence Fusion Prompt}
\label{app:sec:fusion_prompt}

Following the methodology outlined in Section \ref{sec:dataset}, we utilize the \textbf{gpt-3.5-turbo} model with a temperature setting of $0.5$ to generate the fusion sentences using the following prompt:

\begin{tcolorbox}[colback=blue!5!white,colframe=blue!75!black,title= \; \;   \; \;   \;   \;  \; \;   \; \footnotesize{Prompt Structure (in Python)}  ,fontupper=\scriptsize,width=\linewidth, left=0pt, right=0pt, bottom=0pt]
\begin{verbatim}

A = "Sentnece A"
B = "Sentnece B"
max_word = 0.5*(len(A.split()) + len(B.split()))
messages = [
    {
        "role": "system", 
        "content": "You are a paraphraser.", 
    },
    {
        "role": "user",
        "content": f"Fuse the following two sentences in "
                   f"{max_word} words: {A}\n{B}",
    },
]
\end{verbatim}
\end{tcolorbox}
\subsection{Projection Vector}
\label{app:sec:projection}

As discussed in criteria C2 (\ref{para:h2}), C5 (\ref{para:h5}), C6 (\ref{para:h6}), we compute the projection of a vector $\mathbf{v}$ on the plane of vectors $\mathbf{x}$ and $\mathbf{y}$. 
This projection vector is denoted as $\mathbf{v_{proj}}$. 

In the case of three-dimensional space, the projection vector can be trivially computed as  $\mathbf{v_{proj}} = \mathbf{v_{proj}} - \mathbf{v_{proj}} \cdot \mathbf{\hat{n}}$ where $\mathbf{\hat{n}}$ is the normal vector to the plane defined by the input vectors $\mathbf{x}$, $\mathbf{y}$, computed as $\mathbf{\hat{n}} = \mathbf{x} \times \mathbf{y} $.
However, this approach is not directly applicable in the $n$-dimensional space, where we have an $n-2$-dimensional normal subspace to the plane of inputs $\mathbf{x}$ and $\mathbf{y}$ instead of a single normal vector. 

To address this issue, we compute the basis vectors $\mathbf{\hat{b_1}}$ and $\mathbf{\hat{b_2}}$ of the plane defined by the input vectors $\mathbf{x}$ and $\mathbf{y}$ as follows:

\begin{equation}
\mathbf{\widehat{b_1}} = \frac{\mathbf{x}}{\norm{\mathbf{x}}}
\end{equation}

\begin{equation}
\mathbf{b_2} = \mathbf{y} - \mathbf{y} \cdot \mathbf{\widehat{b_1}}
\end{equation}

\begin{equation}
\mathbf{\widehat{b_2}} = \frac{\mathbf{b_2}}{\norm{\mathbf{b_2}}}
\end{equation}

The projection vector $\mathbf{v_{proj}}$ is then computed as the sum of its components along the basis vectors of the plane: 

\begin{equation}
\mathbf{v_{proj}} = \mathbf{v} \cdot \mathbf{\widehat{b_1}} + \mathbf{v} \cdot \mathbf{\widehat{b_2}}
\end{equation}

\subsection{Qualitative Samples}
Here we show some qualitative samples from our synthetic dataset described in Section \ref{sec:dataset}.

\begin{table*}[!htb]\footnotesize
\centering
\ra{1.0}

\begin{subtable}[h]{1.0\textwidth}
    \centering
    \adjustbox{max width=1.0\textwidth}{%
    \begin{tabular}{C{2.25cm}|C{2.25cm} | C{3cm}|C{3cm}|C{2.25cm}|C{2.25cm}}
        \toprule
         \multicolumn{2}{c|}{$\bm{S_{prev}}$} &  \multicolumn{2}{c|}{$\bm{S_{O}}$} & \multicolumn{2}{c}{$\bm{S_{next}}$} \\
         \midrule
         \multicolumn{2}{C{4.5cm}|}{``The leadership said last year's Congress, which was still under Republican control, had never passed a separate bill funding veterans programs.''} &  \multicolumn{2}{C{6cm}|}{``Congress also sent to the president legislation that would fund veterans care at the levels requested by the president through December 14, the leadership said.''} & \multicolumn{2}{C{4.5cm}}{``The current funding level "is still below the $\$$3.9 billion extra that we passed," said Nadeam Elshami, spokesman Pelosi.''} \\
         \midrule
         \multicolumn{3}{C{7.5cm}|}{$\bm{S_1}$} & \multicolumn{3}{C{7.5cm}}{$\bm{S_2}$}  \\
         \midrule
         \multicolumn{3}{C{7.5cm}|}{``According to the leadership, last year's Republican-controlled Congress never passed a separate bill funding veterans programs but sent a legislation funding veterans care to the president until December 14.''} & \multicolumn{3}{C{7.5cm}}{``Congress sent legislation to fund veterans care at requested levels through Dec. 14, despite current funding still being below the $\$$3.9B passed, said Pelosi's spokesman.''}  \\     
        \bottomrule
        \multicolumn{6}{c}{} \\
    \end{tabular}
    }
\end{subtable}
\caption{This table shows the 3 consecutive samples $S_{prev}$, $S_{curr}$ and $S_{next}$ from a narrative. 
Subsequently, employing GPT3.5~\citep{GPT3}, we combine information from pairs of sentences ($S_{prev}, S_{O}$) and ($S_O, S_{next}$), resulting in synthesized pairs denoted as $S_1$ and $S_2$. We use these sentences to create the synthetic samples for \textit{TextOverlap} (Table \ref{tab:qualitative_overlap}), \textit{TextUnion} (Table \ref{tab:qualitative_union}) and \textit{TextDifference} (Table \ref{tab:qualitative_difference1} and \ref{tab:qualitative_difference2}) as described in section \ref{sec:dataset}.
}
\label{tab:qualitative_input_samples}
\end{table*}

\begin{table*}[!htb]\footnotesize
\centering
\ra{1.0}

\begin{subtable}[h]{1.0\textwidth}
    \centering
    \adjustbox{max width=\textwidth}{%
    \begin{tabular}{C{2.25cm}|C{2.25cm} | C{3cm}|C{3cm}|C{2.25cm}|C{2.25cm}}
        \toprule
         
         \multicolumn{6}{c}{\textbf{TextOverlap}} \\
         \midrule
         \multicolumn{2}{c|}{$\bm{A}$ $\bm{(S_{1})}$} &  \multicolumn{2}{c|}{$\bm{B}$  $\bm{(S_{2})}$} & \multicolumn{2}{c}{$\bm{O}$ $\bm{(S_{curr})}$} \\
         \midrule
         \multicolumn{2}{C{4.5cm}|}{``According to the leadership, last year's Republican-controlled Congress never passed a separate bill funding veterans programs but sent a legislation funding veterans care to the president until December 14.''} &  \multicolumn{2}{C{6cm}|}{``Congress sent legislation to fund veterans care at requested levels through Dec. 14, despite current funding still being below the $\$$3.9B passed, said Pelosi's spokesman.''} & \multicolumn{2}{C{4.5cm}}{``Congress also sent to the president legislation that would fund veterans care at the levels requested by the president through December 14, the leadership said.''} \\
        \bottomrule
        \multicolumn{6}{c}{} \\
    \end{tabular}
    }
\end{subtable}
\caption{Synthetic sample for the \textit{TextOverlap} operator using the sentences ($S_1$, $S_2$ and $S_{curr}$) mentioned in table \ref{tab:qualitative_input_samples}.
Here $A$ and $B$ corresponds to the input sentences and $O$ corresponds to the target overlap sentence. 
Follow Section \ref{sec:dataset} for more details. 
}
\label{tab:qualitative_overlap}
\end{table*}

\begin{table*}[!htb]\footnotesize
\centering
\ra{1.0}

\begin{subtable}[h]{1.0\textwidth}
    \centering
    \adjustbox{max width=1.0\textwidth}{%
    \begin{tabular}{C{2.25cm}|C{2.25cm} | C{3cm}|C{3cm}|C{2.25cm}|C{2.25cm}}
        \toprule
         
         \multicolumn{6}{c}{\textbf{TextUnion}} \\
         \midrule
         \multicolumn{2}{c|}{$\bm{A}$  $\bm{(S_{prev})}$} &  \multicolumn{2}{c|}{$\bm{B}$ $\bm{(S_{curr})}$} & \multicolumn{2}{c}{$\bm{U}$ $\bm{(S_{1})}$} \\
         \midrule
         \multicolumn{2}{C{4.5cm}|}{``The leadership said last year's Congress, which was still under Republican control, had never passed a separate bill funding veterans programs.''} &  \multicolumn{2}{C{6cm}|}{``Congress also sent to the president legislation that would fund veterans care at the levels requested by the president through December 14, the leadership said.''} & \multicolumn{2}{C{4.5cm}}{``According to the leadership, last year's Republican-controlled Congress never passed a separate bill funding veterans programs but sent a legislation funding veterans care to the president until December 14.''} \\
         \midrule
         \multicolumn{2}{c|}{$\bm{A}$ 
         $\bm{(S_{curr})}$} &  \multicolumn{2}{c|}{$\bm{B}$ $\bm{(S_{next})}$} & \multicolumn{2}{c}{$\bm{U}$ $\bm{(S_{2})}$} \\
         \midrule
         \multicolumn{2}{C{4.5cm}|}{``Congress also sent to the president legislation that would fund veterans care at the levels requested by the president through December 14, the leadership said.''} &  \multicolumn{2}{C{6cm}|}{``The current funding level "is still below the $\$$3.9 billion extra that we passed," said Nadeam Elshami, spokesman Pelosi.''} & \multicolumn{2}{C{4.5cm}}{``Congress sent legislation to fund veterans care at requested levels through Dec. 14, despite current funding still being below the \$3.9B passed, said Pelosi's spokesman.''} \\
        \bottomrule
        \multicolumn{6}{c}{} \\
    \end{tabular}
    }
\end{subtable}

\caption{Synthetic sample for the \textit{TextUnion} operator using the sentences ($S_{prev}$, $S_{curr}$, $S_{next}$, $S_1$ and $S_2$) mentioned in Table \ref{tab:qualitative_input_samples}.
Here $A$ and $B$ corresponds to the input sentences and $U$ corresponds to the target ``union'' sentence. 
Follow section \ref{sec:dataset} for more details.}
\label{tab:qualitative_union}
\end{table*}

\begin{table*}[!htb]\footnotesize
\centering
\ra{1.0}

\begin{subtable}[h]{1.0\textwidth}
    \centering
    \adjustbox{max width=\textwidth}{%
    \begin{tabular}{C{2.25cm}|C{2.25cm} | C{3cm}|C{3cm}|C{2.25cm}|C{2.25cm}}
        \toprule
         
         \multicolumn{6}{c}{\textbf{TextDifference}} \\

         \midrule
         \multicolumn{2}{c|}{$\bm{A}$ $\bm{(S_{1})}$} &  \multicolumn{2}{c|}{$\bm{B}$ $\bm{(S_{prev})}$} & \multicolumn{2}{c}{$\bm{D}$ $\bm{(S_{curr})}$} \\
         \midrule
         \multicolumn{2}{C{4.5cm}|}{``According to the leadership, last year's Republican-controlled Congress never passed a separate bill funding veterans programs but sent a legislation funding veterans care to the president until December 14.''} &  \multicolumn{2}{C{6cm}|}{``The leadership said last year's Congress, which was still under Republican control, had never passed a separate bill funding veterans programs.''} & \multicolumn{2}{C{4.5cm}}{``Congress also sent to the president legislation that would fund veterans care at the levels requested by the president through December 14, the leadership said.''} \\
         \midrule
         \multicolumn{2}{c|}{$\bm{A}$  $\bm{(S_{1})}$} &  \multicolumn{2}{c|}{$\bm{B}$  $\bm{(S_{curr})}$} & \multicolumn{2}{c}{$\bm{D}$  $\bm{(S_{prev})}$} \\
         \midrule
         \multicolumn{2}{C{4.5cm}|}{``According to the leadership, last year's Republican-controlled Congress never passed a separate bill funding veterans programs but sent a legislation funding veterans care to the president until December 14.''} &  \multicolumn{2}{C{6cm}|}{``Congress also sent to the president legislation that would fund veterans care at the levels requested by the president through December 14, the leadership said.''} & \multicolumn{2}{C{4.5cm}}{``The leadership said last year's Congress, which was still under Republican control, had never passed a separate bill funding veterans programs.''} \\
         \midrule
         \multicolumn{2}{c|}{$\bm{A}$  $\bm{(S_{1})}$} &  \multicolumn{2}{c|}{$\bm{B}$  $\bm{(S_{2})}$} & \multicolumn{2}{c}{$\bm{D}$  $\bm{(S_{prev})}$} \\
         \midrule
         \multicolumn{2}{C{4.5cm}|}{``According to the leadership, last year's Republican-controlled Congress never passed a separate bill funding veterans programs but sent a legislation funding veterans care to the president until December 14.''} &  \multicolumn{2}{C{6cm}|}{``Congress sent legislation to fund veterans care at requested levels through Dec. 14, despite current funding still being below the \$3.9B passed, said Pelosi's spokesman.''} & \multicolumn{2}{C{4.5cm}}{``The leadership said last year's Congress, which was still under Republican control, had never passed a separate bill funding veterans programs.''} \\
        
        \bottomrule
    \end{tabular}
    }
\end{subtable}
\caption{Synthetic sample for the \textit{TextDifference} operator using the sentences ($S_{prev}$, $S_{curr}$, $S_1$ and $S_2$) mentioned in Table \ref{tab:qualitative_input_samples}.
Here $A$ and $B$ corresponds to the input sentences and $D$ corresponds to the target ``difference'' sentence. 
More samples for TextDifference are shown in Table \ref{tab:qualitative_difference2}. 
Follow Section \ref{sec:dataset} for more details.
}
\label{tab:qualitative_difference1}
\end{table*}

\begin{table*}[!htb]\footnotesize
\centering
\ra{1.0}

\begin{subtable}[h]{1.0\textwidth}
    \centering
    \adjustbox{max width=\textwidth}{%
    \begin{tabular}{C{2.25cm}|C{2.25cm} | C{3cm}|C{3cm}|C{2.25cm}|C{2.25cm}}
        \toprule
         
         \multicolumn{6}{c}{\textbf{TextDifference}} \\

         \midrule
         \multicolumn{2}{c|}{$\bm{A}$  $\bm{(S_{2})}$} &  \multicolumn{2}{c|}{$\bm{B}$  $\bm{(S_{curr})}$} & \multicolumn{2}{c}{$\bm{D}$  $\bm{(S_{next})}$} \\
         \midrule
         \multicolumn{2}{C{4.5cm}|}{``Congress sent legislation to fund veterans care at requested levels through Dec. 14, despite current funding still being below the \$3.9B passed, said Pelosi's spokesman.''} &  \multicolumn{2}{C{6cm}|}{``Congress also sent to the president legislation that would fund veterans care at the levels requested by the president through December 14, the leadership said.''} & \multicolumn{2}{C{4.5cm}}{``The current funding level "is still below the \$3.9 billion extra that we passed," said Nadeam Elshami, spokesman Pelosi.''} \\
         \midrule
         \multicolumn{2}{c|}{$\bm{A}$  $\bm{(S_{2})}$} &  \multicolumn{2}{c|}{$\bm{B}$  $\bm{(S_{next})}$} & \multicolumn{2}{c}{$\bm{D}$  $\bm{(S_{curr})}$} \\
         \midrule
         \multicolumn{2}{C{4.5cm}|}{``Congress sent legislation to fund veterans care at requested levels through Dec. 14, despite current funding still being below the \$3.9B passed, said Pelosi's spokesman.''} &  \multicolumn{2}{C{6cm}|}{``The current funding level "is still below the \$3.9 billion extra that we passed," said Nadeam Elshami, spokesman Pelosi.''} & \multicolumn{2}{C{4.5cm}}{``Congress also sent to the president legislation that would fund veterans care at the levels requested by the president through December 14, the leadership said.''} \\
         \midrule
         \multicolumn{2}{c|}{$\bm{A}$  $\bm{(S_{2})}$} &  \multicolumn{2}{c|}{$\bm{B}$  $\bm{(S_{1})}$} & \multicolumn{2}{c}{$\bm{D}$  $\bm{(S_{next})}$} \\
         \midrule
         \multicolumn{2}{C{4.5cm}|}{``Congress sent legislation to fund veterans care at requested levels through Dec. 14, despite current funding still being below the \$3.9B passed, said Pelosi's spokesman.''} &  \multicolumn{2}{C{6cm}|}{``According to the leadership, last year's Republican-controlled Congress never passed a separate bill funding veterans programs but sent a legislation funding veterans care to the president until December 14.''} & \multicolumn{2}{C{4.5cm}}{``The current funding level "is still below the \$3.9 billion extra that we passed," said Nadeam Elshami, spokesman Pelosi.''} \\
         
        \bottomrule
    \end{tabular}
    }
\end{subtable}
\caption{Synthetic sample for the \textit{TextDifference} operator using the sentences ($S_{curr}$, $S_{next}$, $S_1$ and $S_2$) mentioned in table \ref{tab:qualitative_input_samples}.
Here $A$ and $B$ corresponds to the input sentences and $D$ corresponds to the target ``difference'' sentence. 
More samples for TextDifference are shown in Table \ref{tab:qualitative_difference1}. 
Follow section \ref{sec:dataset} for more details.}
\label{tab:qualitative_difference2}
\end{table*}

\clearpage
\clearpage

\begin{table*}[!t]\footnotesize
\centering
\ra{1.0}

\begin{subtable}[h]{1.0\textwidth}
    \centering
    \adjustbox{max width=1.0\textwidth}{%
    \begin{tabular}{ C{2.0cm}|C{2.0cm} | C{2.0cm}|C{2.0cm}|C{2.0cm}|C{2.0cm}| C{1.0cm}}
        \toprule
         \multicolumn{2}{c|}{$\bm{A}$} &  \multicolumn{2}{c|}{$\bm{B}$} & \multicolumn{2}{c|}{$\bm{O}$} & Scores\\
         \midrule
         \multicolumn{2}{C{4cm}|}{``Iancu Munteau's father, who reassured police about relatives' blond hair and blue eyes, revealed his wife's sleeplessness and the sister's tears.''} &  \multicolumn{2}{C{4cm}|}{``The Dublin family's lawyer, Waheed Mudah, accused the police of unjustifiable action and issued a statement outside the Family Court.''} & \multicolumn{2}{C{4cm}|}{``He said his wife couldn't sleep all night and the boy's older sister cried much of the night.''} & $0, 0, 0$ \\

        \midrule
         \multicolumn{2}{C{4cm}|}{``Doretta Cocks from the Campaign for Weekly Waste Collections criticized the change to a three-weekly council collection, fearing the consequences of prolonged storage of rubbish in bins, especially in hot weather.''} &  \multicolumn{2}{C{4cm}|}{``With rubbish left in a bin for three weeks during hot weather, disposing of nappies would be a dreadful situation.''} & \multicolumn{2}{C{4cm}|}{``She said: 'I dread to imagine the consequences of rubbish remaining in a bin awaiting collection for three weeks especially during hot weather.''} & $1, 1$ \\
        \midrule
         \multicolumn{2}{C{4cm}|}{``According to him, the Greeks' picture was only logical between 1,800 to 1,700 BC before they established a civil society.''} &  \multicolumn{2}{C{4cm}|}{``The Greeks, who were previously nomadic, male-dominated, and violent tribes from the steppes, created a civil society upon arriving in the Mediterranean.''} & \multicolumn{2}{C{4cm}|}{``After that, the Greeks had arrived in the Mediterranean and started to create a civil society.''} & $2, 2$ \\
         \midrule
         \multicolumn{2}{C{4cm}|}{``Dr Daniel Sister's Youth pills, priced at £64 for a one-month supply, target wrinkles and free-radicals by combining amino acids and marine enzymes to boost volume and repair skin.''} &  \multicolumn{2}{C{4cm}|}{``Dr Sister's pills calm skin, but for quick relief, try Eminence Calm Skin VitaSkin Solution (£36.22) enriched with amino acids and marine enzymes to strengthen and repair.''} & \multicolumn{2}{C{4cm}|}{``Combining amino acids and marine enzymes, the supplement is said to boost volume while repairing and strengthening skin from within.''} & $3, 3$ \\

         \midrule
         \multicolumn{2}{C{4cm}|}{``Of those who pass away, 88\% die in a care home or hospital, while only 10.2\% die at home or in hospice care.''} &  \multicolumn{2}{C{4cm}|}{``Out of 35,867 people, 87\% women, only 0.2\% died in hospice care and a tenth died at home, on average at 101.''} & \multicolumn{2}{C{4cm}|}{``Only a tenth die in their own home and just 0.2 per cent die in hospice care.''} & $4, 4$ \\
                      
        \bottomrule
        \multicolumn{6}{c}{} \\
    \end{tabular}
    }
\end{subtable}
\caption{Overlap Samples and the Annotator Scores
}
\label{tab:data_with_annotations_overlap}
\end{table*}

\clearpage
\begin{table*}[!t]\footnotesize
\centering
\ra{1.0}

\begin{subtable}[h]{1.0\textwidth}
    \centering
    \adjustbox{max width=1.0\textwidth}{%
    \begin{tabular}{ C{2.0cm}|C{2.0cm} | C{2.0cm}|C{2.0cm}|C{2.0cm}|C{2.0cm}| C{1.0cm}}
        \toprule
         \multicolumn{2}{c|}{$\bm{A}$} &  \multicolumn{2}{c|}{$\bm{B}$} & \multicolumn{2}{c|}{$\bm{D}$} & Scores\\
         \midrule
         \multicolumn{2}{C{4cm}|}{``Ms Felstein was moved to Gold Coast University Hospital for microsurgery on her arm, and Mr Fuller saved her life.''} &  \multicolumn{2}{C{4cm}|}{``After sustaining multiple wounds to her head and upper body, the teenage victim underwent emergency surgery and later microsurgery on her arm at Gold Coast University Hospital.''} & \multicolumn{2}{C{4cm}|}{``Cameron Lindsay, the crime manager for the Richmond Local Area Command, told the Northern Star that Mr Fuller's actions saved Ms Felstein's life.''} & $0, 0, 0$ \\

        \midrule
         \multicolumn{2}{C{4cm}|}{``No response received to inquiries on the impact on your and your children's future.''} &  \multicolumn{2}{C{4cm}|}{``Mehdi Ramezani said they made him promise not to cause a commotion at the funeral, citing future consequences.''} & \multicolumn{2}{C{4cm}|}{``Calls to Iran's judiciary and security officials seeking comment were not returned.''} & $1, 1$ \\
        \midrule
         \multicolumn{2}{C{4cm}|}{``At 71 and with a history of health issues, former Boston mayor Menino vowed to fight disease.''} &  \multicolumn{2}{C{4cm}|}{``The scrappy politician has vowed to fight the disease with an unknown origin, seen in only 3-4\% of patients.''} & \multicolumn{2}{C{4cm}|}{``Menino, the longest-serving Boston mayor who retired from the office a year ago, is 71 and has had many health problems in recent years.''} & $2, 2$ \\
         \midrule
         \multicolumn{2}{C{4cm}|}{``On November 25, 2014, as travelers waited in lines at LaGuardia Airport, the Weather Underground predicted rain and snow on the East Coast for Thanksgiving Eve.''} &  \multicolumn{2}{C{4cm}|}{``Millions of travelers at LaGuardia Airport in New York on November 25, 2014, waited in lines to pass through security before a developing nor'easter.''} & \multicolumn{2}{C{4cm}|}{``The Weather Underground forecast for Wednesday shows the East Coast's chances for rain and snow the day before Thanksgiving .''} & $3, 3$ \\

         \midrule
         \multicolumn{2}{C{4cm}|}{``Over 60\% of France's 4,000 air traffic controllers voted for the protest against government aviation cuts, citing that the measures will lead to a cheaper, less efficient system and impede 'modernisation' between 2015 and 2019.''} &  \multicolumn{2}{C{4cm}|}{``More than 60 per cent of France's 4,000 air traffic controllers have voted in favour of the protest against government aviation cuts.''} & \multicolumn{2}{C{4cm}|}{``They said that the controversial measures, which will be implemented between 2015 and 2019, will lead to a cheaper, less efficient system, and threaten 'modernisation'.''} & $4, 4$ \\
                      
        \bottomrule
        \multicolumn{6}{c}{} \\
    \end{tabular}
    }
\end{subtable}
\caption{Difference Samples and the Annotator Scores
}
\label{tab:data_with_annotations_difference}
\end{table*}

\clearpage
\begin{table*}[!t]\footnotesize
\centering
\ra{1.0}

\begin{subtable}[h]{1.0\textwidth}
    \centering
    \adjustbox{max width=1.0\textwidth}{%
    \begin{tabular}{ C{2.0cm}|C{2.0cm} | C{2.0cm}|C{2.0cm}|C{2.0cm}|C{2.0cm}| C{1.0cm}}
        \toprule
         \multicolumn{2}{c|}{$\bm{A}$} &  \multicolumn{2}{c|}{$\bm{B}$} & \multicolumn{2}{c|}{$\bm{U}$} & Scores\\
         \midrule
         \multicolumn{2}{C{4cm}|}{``Germany have also qualified for Euro 2015 and are among the favourites to win it.''} &  \multicolumn{2}{C{4cm}|}{``The last time England’s Under 21s played at The Riverside Stadium was in February 2012, when they beat Belgium 4-0.''} & \multicolumn{2}{C{4cm}|}{``Germany, among the favorites for Euro 2015, last played at The Riverside Stadium in February 2012, defeating Belgium 4-0.''} & $0, 0, 0$ \\

        \midrule
         \multicolumn{2}{C{4cm}|}{``"The man who actually killed my husband was identified and imprisoned, but he is not sentenced to death," she said in August.''} &  \multicolumn{2}{C{4cm}|}{``The Iranian government's claims that she was convicted of murder are a lie, she told the Guardian newspaper through an intermediary.''} & \multicolumn{2}{C{4cm}|}{``In August, she stated that the man who killed her husband was jailed, but not sentenced to death, refuting the Iranian government's lie via a Guardian intermediary.''} & $1, 1$ \\
        \midrule
         \multicolumn{2}{C{4cm}|}{``The clip reveals that 'Loo-cy' is in fact a motorized toilet, and has a plow in front.''} &  \multicolumn{2}{C{4cm}|}{``On its platform, a toilet paper stand and some reading material are included.''} & \multicolumn{2}{C{4cm}|}{``The clip shows 'Loo-cy', a toilet with a plow, platform with paper stand and reading material.''} & $2, 1, 1$ \\
         \midrule
         \multicolumn{2}{C{4cm}|}{``Relaxing the hours will leave stores less packed and make for a more pleasant shopping experience for customers, say the retailers.''} &  \multicolumn{2}{C{4cm}|}{``The trading laws were successfully relaxed during the Olympic and Paralympic Games in London earlier this year.''} & \multicolumn{2}{C{4cm}|}{``Retailers claim that successfully relaxing the trading laws during the London Olympics led to fewer crowded stores and better shopping experiences.''} & $3, 3$ \\

         \midrule
         \multicolumn{2}{C{4cm}|}{``Previous research has shown that women who weigh more than 155 pounds are at a higher risk of regular oral contraception failure.''} &  \multicolumn{2}{C{4cm}|}{``These drugs take longer to reach normal concentration levels in the blood of obese women compared with normal weight women, according to one study.''} & \multicolumn{2}{C{4cm}|}{``In one study, it was found that oral contraception failure is more common in women weighing over 155 pounds due to the delayed normalization of drug concentration in their bloodstream.''} & $4, 4$ \\
                      
        \bottomrule
        \multicolumn{6}{c}{} \\
    \end{tabular}
    }
\end{subtable}
\caption{Union Samples and the Annotator Scores
}
\label{tab:data_with_annotations_union}
\end{table*}

\clearpage
\subsection{Models Setting}
\label{app:sec:model_description}

All experiments were carried out on a Linux server using NVIDIA Quadro RTX 5000 and NVIDIA RTX A4500 GPUs.
We used seven classical and six LLM-based models in our study, all of which are open-source except for GPT-3. 
Except for the InferSent and LASER models, all models are based on transformer-based architecture and can be further classified into encoder-only and decoder-only models.

To generate sentence embeddings for transformer-based models, we extracted contextualized token embeddings from the final layer corresponding to the input tokens and applied a mean pooling operation, following standard practice in the literature.

Here is a brief description of each model:
\begin{enumerate}
[
leftmargin=*,
itemsep=0ex,
partopsep=-1ex,
parsep=0ex,
]
    \item \textbf{USE} \citep{cer-etal-2018-universal}:  Universal Sentence Encoder (USE) is a transformer-based model that encodes the fixed-size 512-dimensional vector. The TF2.0 Saved Model (v4) was loaded from \citet{tfhub}. The model has been trained for text classification, sentence similarity, and clustering tasks. 
    
    \item \textbf{SBERT} \citep{reimers-gurevych-2019-sentence}: Sentence-BERT is a BERT \citep{devlin-etal-2019-bert} based model which produces semantically meaningful sentence embeddings. 
    These models have been trained on Wikipedia and Book corpus data and further fine-tuned on the NLI dataset. 
    In this work, we used \textbf{SBERT-L} (\href{https://huggingface.co/sentence-transformers/all-mpnet-base-v2}{all-mpnet-base-v2}), currently their best \href{https://www.sbert.net/docs/sentence_transformer/pretrained_models.html}{performing model}, and  \textbf{SBERT-mini} (\href{https://huggingface.co/sentence-transformers/all-MiniLM-L6-v2}{all-MiniLM-L6-v2}), which strikes an optimal balance between performance and speed, variants. 
    Furthermore, we used \textit{SentenceTransformer} library to load the pre-trained model. 
    
    \item \textbf{InferSent} \cite{conneau-etal-2017-supervised}: The model produces sentence embeddings having semantic representations of English sentences. In this work, our model used pre-trained GloVe word embeddings \cite{pennington-etal-2014-glove} with 840B tokens, 2.2M vocabulary, 300-dimensional vector, and, InferSent version 1 encoder. We have also set the batch size to 64, the word embedding dimension size to 300d, and the LSTM encoder size to 2048 with max-pooling layers enabled. Additionally, the model has been trained on the NLI dataset to classify into three categories: entailment, contradiction, and neutral.
    
    \item \textbf{LASER} \citep{artetxe2019massively}: Language-Agnostic-SEntence Representation (LASER) is a model built to perform multilingual sentence embedding tasks and trained in 93 different languages. The model used five BI-LSTM layers in the encoder with max-pooling on the last layer to produce embeddings of a sentence. In this study, we used a pre-trained LASER model with its default settings to produce a sentence embedding for a given English sentence. 
    
    

    \item \textbf{RoBERTa} \citep{liu2020roberta}: Similar to BERT \citep{devlin-etal-2019-bert} model, it was pre-trained with the Masked language modelling (MLM) objective but removed the next-sentence pretraining objective. It encodes the text into a 768-dimensional vector. 

    \item \textbf{SimCSE} ~\cite{gao-etal-2021-simcse}:SimCSE, which stands for Simple Contrastive Learning of Sentence Embedding, is an encoder only model designed to improve sentence embeddings through a contrastive learning framework. By leveraging contrastive loss, SimCSE aims to enhance the semantic representation of sentences, facilitating better performance in various natural language processing tasks.

    \item\textbf{GPT3-Ada}: We used GPT3~\cite{GPT3} text-embedding-ada-002 model which is trained for text search, text similarity, and code search. We generate embedding using OpenAI API~\cite{openai-emb}. 
    
    \item\textbf{LlaMa2}: The LlaMa2~\cite{touvron2023llama} model is a collection of pre-trained and fine-tuned large language models (LLMs) ranging in scale from 7 billion to 70 billion parameters. In this work, we used 7B parameter for encoding text. We utilized the HuggigFace framework with LlaMa2 weights and generated the encodings. 
    To generate the embedding vector, the decoder processes input tokens (embeddings) to generate corresponding output embeddings, and we computed the mean of these output embeddings to serve as the sentence embedding, following standard practice.

    \item \textbf{LLaMA3}: 
    The LLaMA-3 ~\citep{llama3modelcard} is a decoder-only model, which is a pre-trained and instructed fine-tuned language model released in 8B and 70B sizes. In this work, we used the 8B pre-trained model, and the HuggingFace framework was utilized to load the model. We perform inference testing on our proposed criteria using this model. To encode a sentence, we used a similar approach as mentioned in the LLaMA-2 model. The model generates a 4096-dimensional embedding vector.

    \item \textbf{LLaMA3.2}: 
    The LLaMA-3.2 ~\citep{llama32modelcard} is an auto-regressive language model that uses an optimized transformer architecture and is released in 1B and 3B variants for the text-only models. In this work, we used the BB pre-trained model, and the HuggingFace framework was utilized to load the model. To encode a sentence, we used a similar approach as mentioned in the LLaMA-2 model. The model generates a 3072-dimensional embedding vector.

    \item\textbf{Mistral}: Mistral~\cite{mistral} is an open source model multilingual model available in various sizes. In this work, we used Mistral-7B-v0.3 model which we loaded using HuggingFace Framework. The model produced embeddings of size 4096. 

    \item\textbf{OLMo}: Open Language Model by Allen Institute of AI (AI2) is an open-source model. OLMo released different sizes and we use 7B model. We used Huggingface frame work to load the model. We use the same process as above to generate the embedding. It produces a 4096-dimensional embedding. 

    \item\textbf{OpenELM}: Open Efficient Language Model was release by Apple in various sizes. In this work, we use OpenELM-3B model which is their biggest model. We use a similar setup as other decoder-only models. The output embedding size is 3072 dimensions.

    \item\textbf{Qwen}: Qwen  is the large language model and large multimodal model series of the Qwen Team, Alibaba Group. In this work, we use their latest Qwen2.5 model with 7B parameters. It's architecture follows that of LLaMA model with some improvements ~\cite{bai2023qwen}. To generate embeddings, we use a setup similar to that of other decoder-only models. The output embedding size is 3584 dimensions.

    \item\textbf{Gemma}: Gemma is a family of open-source models. In this work, we use Gemma-2 models with 9B parameters and use a similar setup as that of other decoder-only models to generate embeddings. The output embedding size is 3584 dimensions.

\end{enumerate}

\subsection{Extended Results}
\label{app:sec:extended_results}

This section presents the results from additional experiments that we conducted.

\phantomsection
\label{app:h1_results}

\noindent \textbf{C1:}  
As mentioned in the main text, to evaluate criterion C1, we utilize TextOverlap samples ($(\{A, B\}, O)$) from our synthetic dataset. C1 expects that the average similarity between the pairs $(B, O)$ and $(A, O)$ would exceed that of the pair $(A, B)$, 
as indicated by the differ- 470
ences: $Cos(A, O) - Cos(A, B) > 0$ and $Cos(B, O)  Cos(A, B) > 0$, assuming the minimum margin value to be $0$ in condition $C_{O_1}$ and  $C_{O_2}$ (ref section \ref{para:h1}). The results are shown in the Table \ref{tab:h1_analysis_average_appendix}. The average differences across the dataset are more or less close to 0.

\smallskip
\noindent \textbf{Epsilon Sampling}:
To compute the percentage of samples that satisfy conditions $C_{O_1}$ and $C_{O_2}$, as presented in the Table \ref{tab:h1_analysis}, we sample multiple values of $\epsilon_{O_1}$ and $\epsilon_{O_2}$ to ensure robustness, rather than relying solely on a single value of $0$.
In theory, the values of $\epsilon_{O_1}$ and $\epsilon_{O_2}$ can range from $-1$ to $1$. 
However, due to the need to sample over two $\epsilon$ values (essentially sampling over a 2D grid), the computation becomes prohibitively expensive and sparse.
To address this, we compute the minimum and maximum values of $Sim(A, O) - Sim(A, B)$ across all samples and models, which defines the range for $\epsilon_{O_1}$. 
A similar procedure is followed to determine the range for $\epsilon_{O_2}$. 
Subsequently, we sample $132$ values for each $\epsilon$, compute the percentage of samples that satisfy conditions $C_{O_1}$ and $C_{O_2}$ for each combination of $\epsilon_{O_1}$ and $\epsilon_{O_2}$, resulting in a total of $17424$ combinations. The final results are obtained by averaging across all combinations.

\phantomsection
\label{app:h3_results}

\noindent \textbf{C3:}
Akin to the experiment in C1 \ref{app:h1_results}, we also analyzed whether the similarities of pairs $(A, D)$ and $(A, B)$ exceed that of $(B, D)$. 
Our analysis across all similarity and distance measures confirms that all models consistently meet these conditions, showing alignment with our hypotheses. 
Detailed results are presented in the Appendix Table \ref{tab:h3_analysis_average_appendix}.

\smallskip
\noindent \textbf{Epsilon Sampling}:
We follow a similar sampling procedure as that of C1 (sec \ref{app:h1_results}), to compute the percentage of samples that follow conditions $C_{D_1}$ and $C_{D_2}$.


\begin{table*}[!t]\notsotiny
    \centering
    \ra{1.0}

    \begin{subtable}[h]{1.0\textwidth}
    \centering
    \begin{tabular}{rcc}
        \toprule
        \multirow{2}{*}{\textbf{Models}}
         & Cos($A$, $O$) $-$  & Cos($B$, $O$) $-$ \\
         & Cos($A$, $B$) & Cos($A$, $B$) \\
         
        \midrule
        SBERT-mini & $0.0478 \pm 0.127$ & $0.1109 \pm 0.1471 $ \\
        LASER & $-0.0019 \pm 0.0601$ & $0.0171 \pm 0.0679 $ \\
        USE & $0.0417 \pm 0.1168$ & $0.0886 \pm 0.1336 $ \\
        RoBERTa  & $-0.0005 \pm 0.0053$ & $0.0011 \pm 0.0058 $ \\
        SBERT-L & $0.0471 \pm 0.1237$ & $0.1023 \pm 0.1423 $ \\
        SimCSE & $0.0298 \pm 0.0899$ & $0.0646 \pm 0.1014$ \\
        InferSent & $0.0037 \pm 0.047$ & $0.0174 \pm 0.0523$ \\
        \midrule
        GPT3 & $0.0125 \pm 0.0291$ & $0.0253 \pm 0.0333 $ \\
        LLaMA2 & $-0.0036 \pm 0.0715$ & $0.0207 \pm 0.0783 $ \\
        Mistral & $-0.0007 \pm 0.0637$ & $0.026 \pm 0.0716$ \\
        LLaMA3 & $-0.0088 \pm 0.0718$ & $0.0204 \pm 0.0817$ \\
        OLMo & $0.0015 \pm 0.0787$ & $0.0367 \pm 0.0904$ \\
        OpenELM & $-0.0086 \pm 0.0715$ & $0.0036 \pm 0.0744$ \\
        LLaMA3.2 & $-0.0017 \pm 0.043$ & $0.0166 \pm 0.0488$ \\
        Qwen & $-0.002 \pm 0.0297$ & $0.0064 \pm 0.0322$ \\
        Gemma & $-0.0081 \pm 0.0774$ & $0.0258 \pm 0.09$ \\
        \bottomrule
    \end{tabular}
    \caption{Cos Product scores}
    \label{subtab:h1_average_cos}
    \end{subtable}%

    \hfill
    \newline
    
    \begin{subtable}[h]{1.0\textwidth}
    \centering
    \begin{tabular}{rcc}
        \toprule
        \multirow{2}{*}{\textbf{Model}}
         & Dot($A$, $O$) $-$  & Dot($B$, $O$) $-$ \\
         & Dot($A$, $B$) & Dot($A$, $B$) \\
         
        \midrule
        SBERT-mini & $0.0478 \pm 0.127$ & $0.1109 \pm 0.1471 $ \\
        LASER & $-0.0019 \pm 0.0601$ & $0.0171 \pm 0.0679 $ \\
        USE & $0.0417 \pm 0.1168$ & $0.0886 \pm 0.1336 $ \\
        RoBERTa & $0.817 \pm 4.1781$ & $1.011 \pm 4.4399$ \\
        SBERT-L & $0.0471 \pm 0.1237$ & $0.1023 \pm 0.1423$ \\
        SimCSE & $5.5297 \pm 13.9868$ & $10.3788 \pm 15.8607$ \\
        InferSent & $-0.9162 \pm 2.2271$ & $-0.7196 \pm 2.4119$ \\
        \midrule
        GPT3 & $0.0125 \pm 0.0291$ & $0.0253 \pm 0.0333 $ \\
        LLaMA2 & $-0.0036 \pm 0.0715$ & $0.0207 \pm 0.0783 $ \\
        Mistral & $1075.1118 \pm 5278.5236$ & $2359.0408 \pm 5660.1179$ \\
        LLaMA3 & $-44.9503 \pm 656.7071$ & $170.436 \pm 729.8401$ \\
        OLMo & $30.6913 \pm 119.101$ & $71.7974 \pm 130.4169$ \\
        OpenELM & $0.0957 \pm 6.0255$ & $0.6717 \pm 6.7635$ \\
        LLaMA3.2 & $41.42 \pm 234.7219$ & $106.5151 \pm 250.9682$ \\
        Qwen & $665.3865 \pm 5229.5611$ & $1388.2295 \pm 5567.7608$ \\
        Gemma & $-71.7269 \pm 953.6203$ & $265.6794 \pm 1086.0704$ \\
        
        \bottomrule
    \end{tabular}
    \caption{Dot Product scores}
    \label{subtab:h1_average_dot}
    \end{subtable}%
    \hfill
    \newline
    
    \begin{subtable}[h]{1.0\textwidth}
    \centering
    \begin{tabular}{rcc}
        \toprule
        \multirow{2}{*}{\textbf{Model}}
          & NED($A$, $O$) $-$  & NED($B$, $O$) $-$ \\
         & NED($A$, $B$) & NED($A$, $B$) \\
        \midrule
        SBERT-mini & $-0.0239 \pm 0.0635$ & $-0.0554 \pm 0.0736$ \\
        LASER & $-0.006 \pm 0.0499$ & $-0.023 \pm 0.0573$ \\
        USE & $-0.0208 \pm 0.0584$ & $-0.0443 \pm 0.0668$ \\
        RoBERTa & $0.0003 \pm 0.0027$ & $-0.0006 \pm 0.003$ \\
        SBERT-L & $-0.0235 \pm 0.0618$ & $-0.0512 \pm 0.0711$ \\
        SimCSE & $-0.0149 \pm 0.0451$ & $-0.0324 \pm 0.0508$ \\
        InferSent & $-0.009 \pm 0.0438$ & $-0.0235 \pm 0.0494$ \\
        \midrule
        GPT3 & $-0.0063 \pm 0.0145$ & $-0.0127 \pm 0.0167$ \\
        LLaMA2 & $0.0018 \pm 0.0365$ & $-0.0106 \pm 0.04$ \\
        Mistral & $0.0004 \pm 0.0331$ & $-0.0133 \pm 0.0372$ \\
        LLaMA3 & $0.0044 \pm 0.0361$ & $-0.0102 \pm 0.0411$ \\
        OLMo & $-0.0006 \pm 0.0398$ & $-0.0183 \pm 0.0457$ \\
        OpenELM & $0.0046 \pm 0.0371$ & $-0.0019 \pm 0.0386$ \\
        LLaMA3.2 & $0.0008 \pm 0.0219$ & $-0.0084 \pm 0.0249$ \\
        Qwen & $0.0009 \pm 0.0157$ & $-0.0035 \pm 0.017$ \\
        Gemma & $0.004 \pm 0.0389$ & $-0.013 \pm 0.0453$ \\
        
        \bottomrule
    \end{tabular}
    \caption{Normalized Euclidean Distance (NED) Scores}
    \label{subtab:h1_average_ned}
    \end{subtable}

    \caption{Average similarity/distance scores (mean $\pm$ std) for pairs $(A, B)$, $(A, O)$ and $(B, O)$ in terms of their pairwise differences, denoted as $M(A, O) - M(A, B)$ and $M(B, O) - M(A, B)$ where $M$ represents the similarity/distance measure.
    Positive differences are desired for similarity measures, while negative differences indicate favorable outcomes for distance measures. 
    Cases where the criteria are not met are highlighted in \textcolor{red}{red}.}
    \label{tab:h1_analysis_average_appendix}
\end{table*}

\begin{table*}[!t]\notsotiny\ContinuedFloat
    \centering
    \ra{1.0}
    
    \begin{subtable}[h]{1.0\textwidth}
    \centering
    \begin{tabular}{rcc}
        \toprule
        \multirow{2}{*}{\textbf{Model}}
         & L1($A$, $O$) $-$  & L1($B$, $O$) $-$ \\
         & L1($A$, $B$) & L1($A$, $B$) \\
        \midrule
        SBERT-mini & $-0.9016 \pm 2.3522$ & $-2.1694 \pm 2.8285$ \\
        LASER & $-0.4751 \pm 1.5642$ & $-1.0651 \pm 1.8168$ \\
        USE & $-0.8574 \pm 2.4252$ & $-1.8763 \pm 2.828$ \\
        RoBERTa & $0.1887 \pm 6.622$ & $-2.1853 \pm 7.6077$ \\
        SBERT-L & $-1.2212 \pm 3.2268$ & $-2.805 \pm 3.8712$ \\
        SimCSE & $-10.1851 \pm 33.0862$ & $-23.9622 \pm 38.1013$ \\
        InferSent & $-6.5953 \pm 19.4294$ & $-12.9134 \pm 21.4627$ \\
        \midrule
        GPT3 & $-0.8877 \pm 2.0592$ & $-1.875 \pm 2.4403$ \\
        LLaMA2 & $25.9401 \pm 274.3383$ & $-89.4733 \pm 320.8035$ \\
        Mistral & $73.1599 \pm 829.0824$ & $-318.9665 \pm 992.263$ \\
        LLaMA3 & $53.258 \pm 465.509$ & $-152.8833 \pm 553.7105$ \\
        OLMo & $17.4305 \pm 196.0634$ & $-71.2001 \pm 232.6856$ \\
        OpenELM & $1.6981 \pm 15.6927$ & $-4.8918 \pm 18.2963$ \\
        LLaMA3.2 & $17.3877 \pm 207.3297$ & $-76.5787 \pm 246.3243$ \\
        Qwen & $55.0646 \pm 527.7385$ & $-165.5535 \pm 622.2416$ \\
        Gemma & $43.0519 \pm 515.6764$ & $-196.8893 \pm 621.6083$ \\
        
        \bottomrule
    \end{tabular}
    \caption{L1 scores}
    \label{subtab:h1_average_l1}
    \end{subtable}%
    \hfill
    \newline
    
    \begin{subtable}[h]{1.0\textwidth}
    \centering
    \begin{tabular}{rcc}
        \toprule
        \multirow{2}{*}{\textbf{Model}}
         & L2($A$, $O$) $-$  & L2($B$, $O$) $-$ \\
         & L2($A$, $B$) & L2($A$, $B$) \\
        \midrule
        SBERT-mini & $-0.0578 \pm 0.1502$ & $-0.1393 \pm 0.1812$ \\
        LASER & $-0.0165 \pm 0.0644$ & $-0.0385 \pm 0.0738$ \\
        USE & $-0.0477 \pm 0.1278$ & $-0.1022 \pm 0.1497$ \\
        RoBERTa & $0.0312 \pm 0.3748$ & $-0.0923 \pm 0.4229$ \\
        SBERT-L & $-0.0576 \pm 0.1492$ & $-0.1315 \pm 0.1793$ \\
        SimCSE & $-0.4599 \pm 1.4941$ & $-1.0838 \pm 1.7214$ \\
        InferSent & $-0.1111 \pm 0.3558$ & $-0.2327 \pm 0.399$ \\
        \midrule
        GPT3 & $-0.0285 \pm 0.0659$ & $-0.0601 \pm 0.0782$ \\
        LLaMA2 & $0.6894 \pm 6.4284$ & $-1.6552 \pm 7.3423$ \\
        Mistral & $2.0148 \pm 22.171$ & $-7.418 \pm 25.9386$ \\
        LLaMA3 & $1.2035 \pm 9.6925$ & $-2.9451 \pm 11.4574$ \\
        OLMo & $0.3442 \pm 3.8367$ & $-1.3922 \pm 4.5553$ \\
        OpenELM & $0.1244 \pm 0.9982$ & $-0.1038 \pm 1.0808$ \\
        LLaMA3.2 & $0.4167 \pm 4.8033$ & $-1.7425 \pm 5.6961$ \\
        Qwen & $1.8233 \pm 21.3757$ & $-4.8419 \pm 24.0756$ \\
        Gemma & $1.1069 \pm 11.3373$ & $-4.0356 \pm 13.5977$ \\
        
        \bottomrule
    \end{tabular}
    \caption{L2 scores}
    \label{subtab:h1_average_l2}
    \end{subtable}

    \caption*{Table \ref{tab:h1_analysis_average_appendix}: Average similarity/distance scores (mean $\pm$ std) for pairs $(A, B)$, $(A, O)$ and $(B, O)$ in terms of their pairwise differences, denoted as $M(A, O) - M(A, B)$ and $M(B, O) - M(A, B)$ where $M$ represents the similarity/distance measure.
    Positive differences are desired for similarity measures, while negative differences indicate favorable outcomes for distance measures. 
    Cases where the criteria are not met are highlighted in \textcolor{red}{red}.}
\end{table*}

\begin{table*}[!t]\notsotiny
    \centering
    \ra{1.0}

        
    \begin{subtable}[h]{0.5\linewidth} 
    \centering
    \begin{tabular}{rcccc}
        \toprule
        \multirow{2}{*}{\textbf{Model}}
         & $\bf{C_{O_1}}$ = T & $\bf{C_{O_1}}$ = T & $\bf{C_{O_1}}$ = F & $\bf{C_{O_1}}$ = F \\
         & $\bf{C_{O_2}}$ = T & $\bf{C_{O_2}}$ = F & $\bf{C_{O_2}}$ = T & $\bf{C_{O_2}}$ = F \\
        \midrule
        SBERT-mini & $\bf{28.83}$ & $24.23$ & $25.26$ & $21.68$ \\
        LASER & $25.73$ & $25.01$ & $24.7$ & $24.56$ \\
        USE & $27.94$ & $24.73$ & $24.89$ & $22.45$ \\
        RoBERTa & $23.73$ & $26.13$ & $23.86$ & $26.28$ \\
        SBERT-L & $28.54$ & $24.47$ & $25.06$ & $21.93$ \\
        SimCSE & $26.79$ & $25.12$ & $24.69$ & $23.4$ \\
        InferSent & $25.89$ & $25.24$ & $24.58$ & $24.28$ \\
        \midrule
        GPT3 & $25.02$ & $25.85$ & $24.14$ & $24.98$ \\
        LLaMA2 & $24.45$ & $25.29$ & $24.55$ & $25.71$ \\
        Mistral & $24.66$ & $25.26$ & $24.65$ & $25.44$ \\
        LLaMA3 & $24.27$ & $25.13$ & $24.7$ & $25.9$ \\
        OLMo & $25.04$ & $25.0$ & $24.85$ & $25.1$ \\
        OpenELM & $23.78$ & $25.58$ & $24.19$ & $26.45$ \\
        LLaMA3.2 & $24.3$ & $25.54$ & $24.41$ & $25.75$ \\
        Qwen & $23.96$ & $25.86$ & $24.1$ & $26.07$ \\
        Gemma & $24.47$ & $24.99$ & $24.82$ & $25.72$ \\
        
        \bottomrule
    \end{tabular}
    \caption{Normalized Euclidean Distance (NED) scores}
    \end{subtable}

    \caption{
    Percentage of samples adhering to various scenarios, ranging from instances where both conditions $C_{O_1}$ and $C_{O_2}$ (refer to section \ref{para:h2}) are satisfied to cases where neither condition is met.
    Notably, SBERT variants demonstrates highest efficacy even outperforming all the 6 LLMs.
    The results for cosine similarity are provided in Table \ref{tab:h1_analysis} in the main text.
    }
    \label{tab:h1_analysis_appendix}
\end{table*}

\begin{table*}[!t]\notsotiny
    \centering
    \ra{1.0}

    \begin{subtable}[h]{1.0\columnwidth}
    \centering
    \begin{tabular}{rccc}
        \toprule
        \multirow{2}{*}{\textbf{Model}}
         & \multirow{2}{*}{Cos($B$, $D$)} & Cos($A$, $D$) $-$  & Cos($A$, $B$) $-$ \\
         & & Cos($B$, $D$) & Cos($B$, $D$) \\
         
        \midrule
        SBERT-mini & $0.1759$ & $0.4292$ & $0.4111$ \\
        LASER & $0.5787$ & $0.1535$ & $0.1547$ \\
        USE & $0.1718$ & $0.3825$ & $0.3678$ \\

        RoBERTa & $0.9706$ & $0.0137$ & $0.014$ \\
        SBERT-L & $0.2243$ & $0.4007$ & $0.3829$ \\
        SimCSE & $0.4202$ & $0.2966$ & $0.2872$ \\
        InferSent & $0.6954$ & $0.1105$ & $0.11$ \\

        \midrule
        
        GPT3 & $0.7871$ & $0.1093$ & $0.105$ \\
        LLaMA2 & $0.6618$ & $0.141$ & $0.1416$ \\
        Mistral & $0.6808$ & $0.1371$ & $0.1361$ \\
        LLaMA3 & $0.633$ & $0.1547$ & $0.1567$ \\
        OLMo & $0.5668$ & $0.1781$ & $0.1758$ \\
        OpenELM & $0.8193$ & $0.0807$ & $0.0835$ \\
        LLaMA3.2 & $0.7833$ & $0.0902$ & $0.09$ \\
        Qwen & $0.8811$ & $0.0536$ & $0.0545$ \\
        Gemma & $0.5798$ & $0.1672$ & $0.1686$ \\
        
        \bottomrule
    \end{tabular}
    \caption{Cosine Similarity scores}
    \label{subtab:h3_average_cos}
    \end{subtable}

    \hfill
    \newline
    
    \begin{subtable}[h]{0.5\linewidth} 
    \centering
    \begin{tabular}{rccc}
        \toprule
        \multirow{2}{*}{\textbf{Model}}
         & \multirow{2}{*}{Dot($B$, $D$)} & Dot($A$, $D$) $-$  & Dot($A$, $B$) $-$ \\
         & & Dot($B$, $D$) & Dot($B$, $D$) \\
         
        \midrule
        SBERT-mini & $0.1759$ & $0.4292$ & $0.4111$ \\
        LASER & $0.2968$ & $0.1011$ & $0.1126$ \\
        USE & $0.1718$ & $0.3825$ & $0.3678$ \\
        RoBERTa & $144.8261$ & $1.2535$ & $1.0657$ \\
        SBERT-L & $0.2243$ & $0.4007$ & $0.3829$ \\
        SimCSE & $59.7491$ & $41.1437$ & $39.3416$ \\
        InferSent & $13.3254$ & $3.0418$ & $3.4375$ \\
        \midrule
        GPT3 & $0.7871$ & $0.1093$ & $0.105$ \\
        LLaMA2 & $2131.2321$ & $377.7673$ & $353.4466$ \\
        Mistral & $30052.8194$ & $4907.6797$ & $4415.9909$ \\
        LLaMA3 & $4502.058$ & $1052.4489$ & $1055.2276$ \\
        OLMo & $641.5517$ & $169.5942$ & $155.5329$ \\
        OpenELM & $39.0733$ & $3.5559$ & $3.4091$ \\
        LLaMA3.2 & $2492.9407$ & $235.8688$ & $216.1651$ \\
        Qwen & $60057.2013$ & $2933.3561$ & $2681.7417$ \\
        Gemma & $5490.4508$ & $1537.6927$ & $1540.2926$ \\
        
        \bottomrule
    \end{tabular}
    \caption{Dot Product scores}
    \label{subtab:h3_average_dot}
    \end{subtable}%
    \begin{subtable}[h]{0.5\linewidth} 
    \centering
    \begin{tabular}{rccc}
        \toprule
        \multirow{2}{*}{\textbf{Model}}
         & \multirow{2}{*}{NED($B$, $D$)} & NED($A$, $D$) $-$  & NED($A$, $B$) $-$ \\
         & & NED($B$, $D$) & NED($B$, $D$) \\
        \midrule
        SBERT-mini & $0.412$ & $-0.2146$ & $-0.2056$ \\
        LASER & $0.3587$ & $-0.1256$ & $-0.1233$ \\
        USE & $0.4141$ & $-0.1913$ & $-0.1839$ \\
        RoBERTa & $0.0151$ & $-0.007$ & $-0.0072$ \\
        SBERT-L & $0.3879$ & $-0.2004$ & $-0.1915$ \\
        SimCSE & $0.2908$ & $-0.1484$ & $-0.1437$ \\
        InferSent & $0.2978$ & $-0.1046$ & $-0.1015$ \\
        \midrule
        GPT3 & $0.1065$ & $-0.0547$ & $-0.0525$ \\
        LLaMA2 & $0.1724$ & $-0.0714$ & $-0.0718$ \\
        Mistral & $0.164$ & $-0.07$ & $-0.0696$ \\
        LLaMA3 & $0.1846$ & $-0.0775$ & $-0.0785$ \\
        OLMo & $0.2189$ & $-0.0894$ & $-0.0884$ \\
        OpenELM & $0.0957$ & $-0.0423$ & $-0.0438$ \\
        LLaMA3.2 & $0.1101$ & $-0.0458$ & $-0.0457$ \\
        Qwen & $0.0625$ & $-0.0282$ & $-0.0287$ \\
        Gemma & $0.2112$ & $-0.0837$ & $-0.0844$ \\
        
        \bottomrule
    \end{tabular}
    \caption{Normalized Euclidean Distance scores}
    \label{subtab:h3_average_ned}
    \end{subtable}

    \hfill
    \newline
    
    \begin{subtable}[h]{0.5\linewidth} 
    \centering
    \begin{tabular}{rccc}
        \toprule
        \multirow{2}{*}{\textbf{Model}}
         & \multirow{2}{*}{L1($B$, $D$)} & L1($A$, $D$) $-$  & L1($A$, $B$) $-$ \\
         & & L1($B$, $D$) & L1($B$, $D$) \\
        \midrule
        SBERT-mini & $19.9154$ & $-6.5012$ & $-6.1517$ \\
        LASER & $13.6975$ & $-2.9907$ & $-2.7936$ \\
        USE & $23.3426$ & $-6.9085$ & $-6.5969$ \\
        RoBERTa & $57.8847$ & $-16.2788$ & $-16.4528$ \\
        SBERT-L & $26.7974$ & $-8.5946$ & $-8.1259$ \\
        SimCSE & $283.5249$ & $-91.0625$ & $-87.9439$ \\
        InferSent & $155.2643$ & $-35.4458$ & $-32.8676$ \\
        \midrule
        GPT3 & $20.2788$ & $-6.3683$ & $-6.0546$ \\
        LLaMA2 & $2203.3889$ & $-570.3142$ & $-576.4169$ \\
        Mistral & $7122.5638$ & $-1761.0884$ & $-1767.1291$ \\
        LLaMA3 & $3535.5936$ & $-898.9593$ & $-907.2771$ \\
        OLMo & $1597.9687$ & $-408.2086$ & $-411.5927$ \\
        OpenELM & $115.7073$ & $-30.6653$ & $-31.0341$ \\
        LLaMA3.2 & $1608.3736$ & $-403.8263$ & $-405.1005$ \\
        Qwen & $4035.4402$ & $-1022.0625$ & $-1034.0838$ \\
        Gemma & $4132.0292$ & $-994.5374$ & $-996.2598$ \\
        
        \bottomrule
    \end{tabular}
    \caption{L1 scores}
    \label{subtab:h3_average_l1}
    \end{subtable}%
    \begin{subtable}[h]{0.5\linewidth} 
    \centering
    \begin{tabular}{rccc}
        \toprule
        \multirow{2}{*}{\textbf{Model}}
         & \multirow{2}{*}{L2($B$, $D$)} & L2($A$, $D$) $-$  & L2($A$, $B$) $-$ \\
         & & L2($B$, $D$) & L2($B$, $D$) \\
        \midrule
        SBERT-mini & $1.2811$ & $-0.418$ & $-0.3957$ \\
        LASER & $0.6556$ & $-0.1239$ & $-0.1174$ \\
        USE & $1.2834$ & $-0.3572$ & $-0.34$ \\
        RoBERTa & $2.9712$ & $-0.8398$ & $-0.8569$ \\
        SBERT-L & $1.2417$ & $-0.3997$ & $-0.3775$ \\
        SimCSE & $12.836$ & $-4.1229$ & $-3.9818$ \\
        InferSent & $3.4389$ & $-0.6627$ & $-0.6208$ \\
        \midrule
        GPT3 & $0.6504$ & $-0.2044$ & $-0.1943$ \\
        LLaMA2 & $46.5523$ & $-11.9966$ & $-12.1941$ \\
        Mistral & $167.2457$ & $-43.9942$ & $-44.3437$ \\
        LLaMA3 & $71.9837$ & $-18.359$ & $-18.5808$ \\
        OLMo & $31.3103$ & $-7.9917$ & $-8.0589$ \\
        OpenELM & $4.1316$ & $-1.1716$ & $-1.2114$ \\
        LLaMA3.2 & $36.9676$ & $-9.3002$ & $-9.3365$ \\
        Qwen & $127.6435$ & $-34.9708$ & $-35.6346$ \\
        Gemma & $89.1021$ & $-21.4315$ & $-21.5531$ \\
        
        \bottomrule
    \end{tabular}
    \caption{L2 scores}
    \label{subtab:h3_average_l2}
    \end{subtable}
    
    \caption{
    Average similarity/distance scores for pairs $(B, D)$, $(A, D)$ and $(A, B)$ in terms of their pairwise differences, denoted as $M(A, D) - M(B, D)$ and $M(A, B) - (M(B, D))$ where $M$ represents the similarity/distance measure.
    Positive differences are desired for similarity measures, while negative differences indicate favorable outcomes for distance measures. 
    All the models follow our expectations across all the metrics. Refer to \ref{para:h3_results} for details. 
    }
    \label{tab:h3_analysis_average_appendix}
\end{table*}
\begin{table*}[!t]\notsotiny
    \centering
    \ra{1.0}

    \begin{subtable}[t]{0.5\textwidth}
    \centering
    \begin{tabular}{rcccc}
        \toprule
        \multirow{2}{*}{\textbf{Model}}
         & $\bf{C_{D_1}}$ = T & $\bf{C_{D_1}}$ = T & $\bf{C_{D_1}}$ = F & $\bf{C_{D_1}}$ = F \\
         & $\bf{C_{D_2}}$ = T & $\bf{C_{D_2}}$ = F & $\bf{C_{D_2}}$ = T & $\bf{C_{D_2}}$ = F \\
        \midrule
        SBERT-mini & $34.75$ & $27.25$ & $22.37$ & $15.63$ \\
        LASER & $23.85$ & $27.42$ & $23.14$ & $25.59$ \\
        USE & $31.78$ & $27.41$ & $22.68$ & $18.14$ \\
        RoBERTa & $95.23$ & $2.02$ & $2.49$ & $0.25$ \\
        SBERT-L & $98.22$ & $0.82$ & $0.88$ & $0.07$ \\
        SimCSE & $98.28$ & $0.78$ & $0.87$ & $0.07$ \\
        InferSent & $94.3$ & $2.66$ & $2.76$ & $0.27$ \\
        \midrule
        GPT3 & $16.34$ & $26.44$ & $21.91$ & $35.31$ \\
        LLaMA2 & $18.22$ & $26.52$ & $22.44$ & $32.83$ \\
        Mistral & $91.13$ & $3.87$ & $4.33$ & $0.67$ \\
        LLaMA3 & $91.26$ & $3.67$ & $4.28$ & $0.79$ \\
        OLMo & $92.6$ & $3.27$ & $3.58$ & $0.55$ \\
        OpenELM & $77.69$ & $9.29$ & $9.82$ & $3.21$ \\
        LLaMA3.2 & $15.59$ & $26.04$ & $21.84$ & $36.52$ \\
        Qwen & $13.99$ & $25.46$ & $21.4$ & $39.15$ \\
        Gemma & $19.41$ & $26.81$ & $22.8$ & $30.98$ \\
        \bottomrule
    \end{tabular}
    \caption{Normalized Euclidean Distance scores}
    \end{subtable}

    \caption{
    Analysis of the percentage of samples adhering to various scenarios, ranging from instances where both conditions $C_{D_1}$ and $C_{D_2}$ (refer to section \ref{para:h3}) are satisfied to cases where both conditions are not met.
    Remarkably, classical models exhibit significantly better interpretability according to criterion C3 compared to LLMs on average, with SBERT-mini demonstrating the highest percentage of samples adhering to C3.
    The results for cosine similarity are provided in Table \ref{tab:h3_analysis} in the main text.
    }
    \label{tab:h3_analysis_appendix}
\end{table*}



\clearpage



\begin{figure*}[!t]
    \centering
    
    \begin{subfigure}[b]{1.0\textwidth}
    \centering
    \begin{subfigure}[b]{0.32\textwidth}
         \centering
         \begin{tikzpicture}
            \node[anchor=north] at (current bounding box.north) {\textbf{TextOverlap}};
        \end{tikzpicture}
         \includegraphics[width=1.0\textwidth]{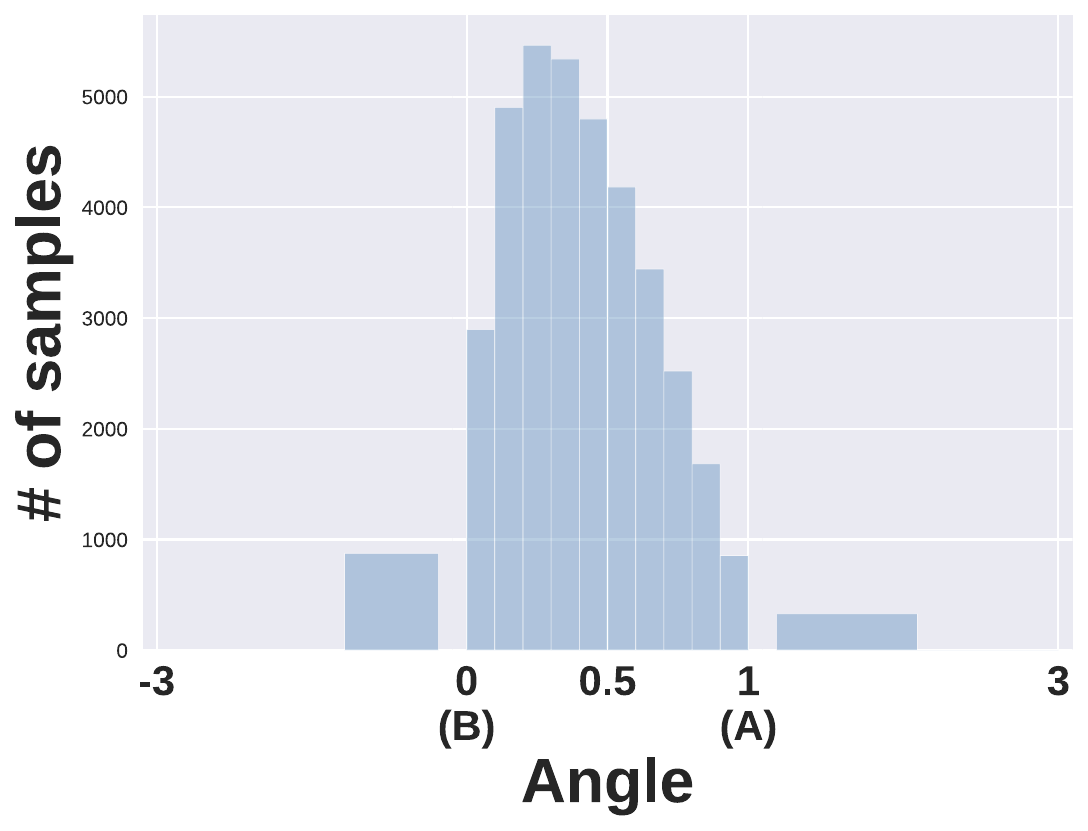}
         \label{fig:sbert_overlap}
     \end{subfigure}
    ~
     \begin{subfigure}[b]{0.32\textwidth}
         \centering
         \begin{tikzpicture}
            \node[anchor=north] at (current bounding box.north) {\textbf{TextDifference}};
        \end{tikzpicture}
        \includegraphics[width=1.0\textwidth]{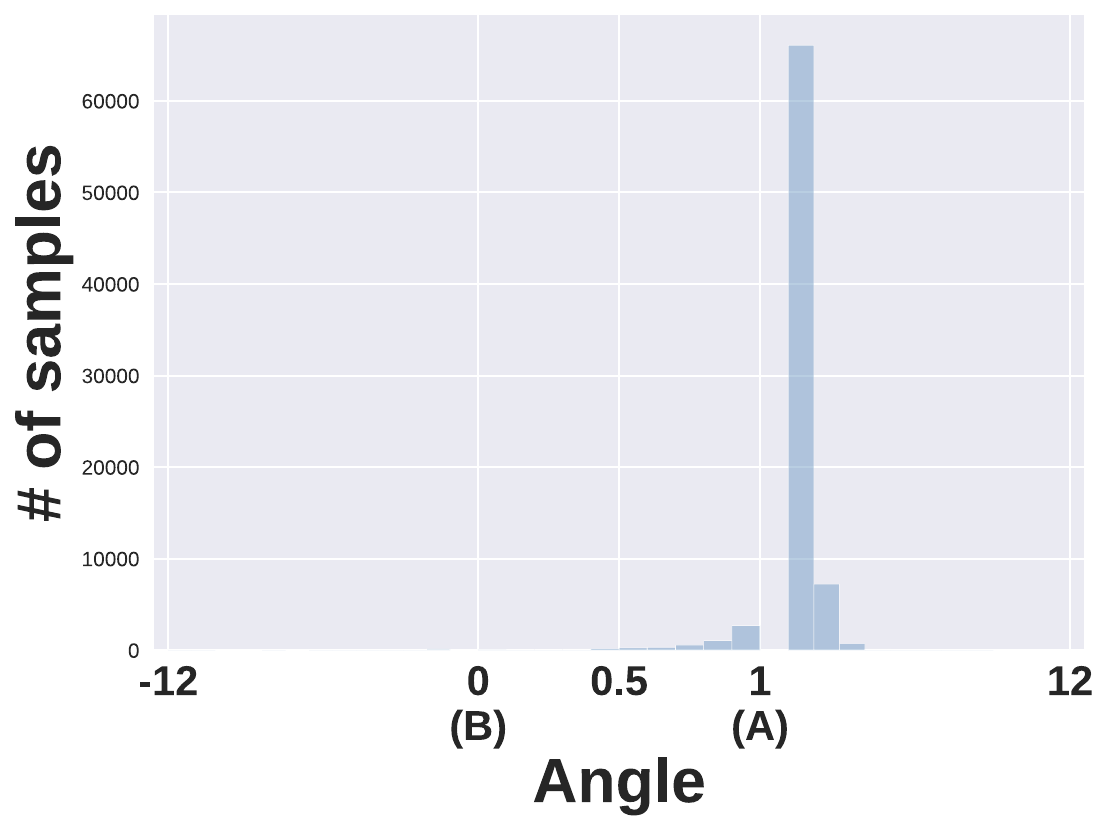}
         \label{fig:sbert_difference}
     \end{subfigure}
    ~
     \begin{subfigure}[b]{0.32\textwidth}
         \centering
         \begin{tikzpicture}
            \node[anchor=north] at (current bounding box.north) {\textbf{TextUnion}};
        \end{tikzpicture}
         \includegraphics[width=1.0\textwidth]{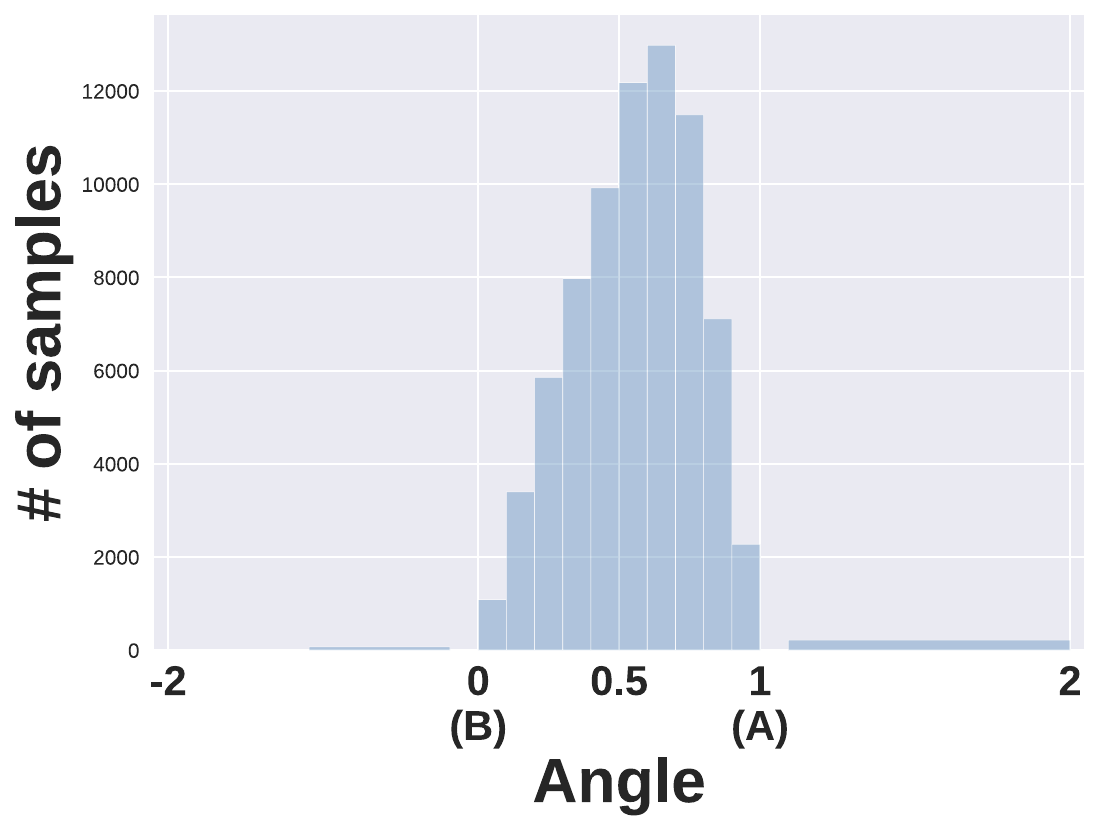}
         \label{fig:sbert_union}
     \end{subfigure}

    \vspace{-4mm}
     \subcaption{\textbf{SBERT-mini}}
    \label{fig:sbert_projection}
    \end{subfigure}
     \hfill

    \begin{subfigure}[b]{1.0\textwidth}
    \centering
    \begin{subfigure}[b]{0.32\textwidth}
         \centering
         \includegraphics[width=1.0\textwidth]{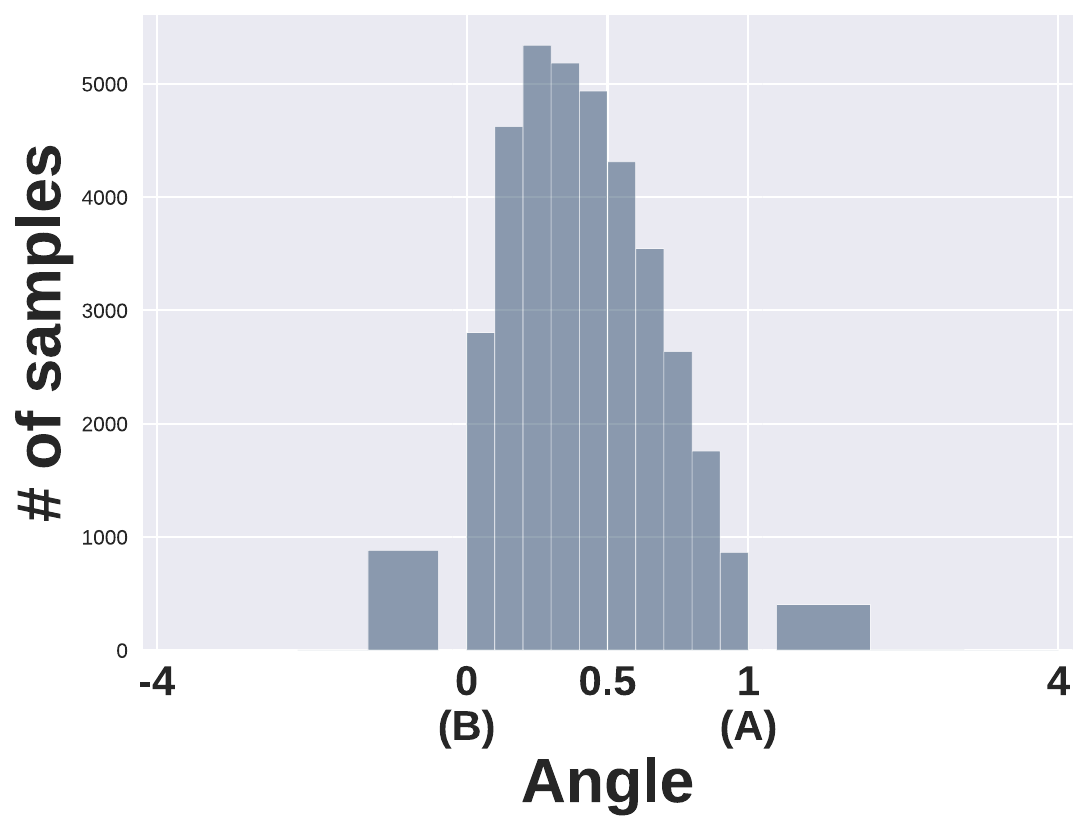}
         \label{fig:mpnet_overlap}
     \end{subfigure}
    ~
     \begin{subfigure}[b]{0.32\textwidth}
         \centering
        \includegraphics[width=1.0\textwidth]{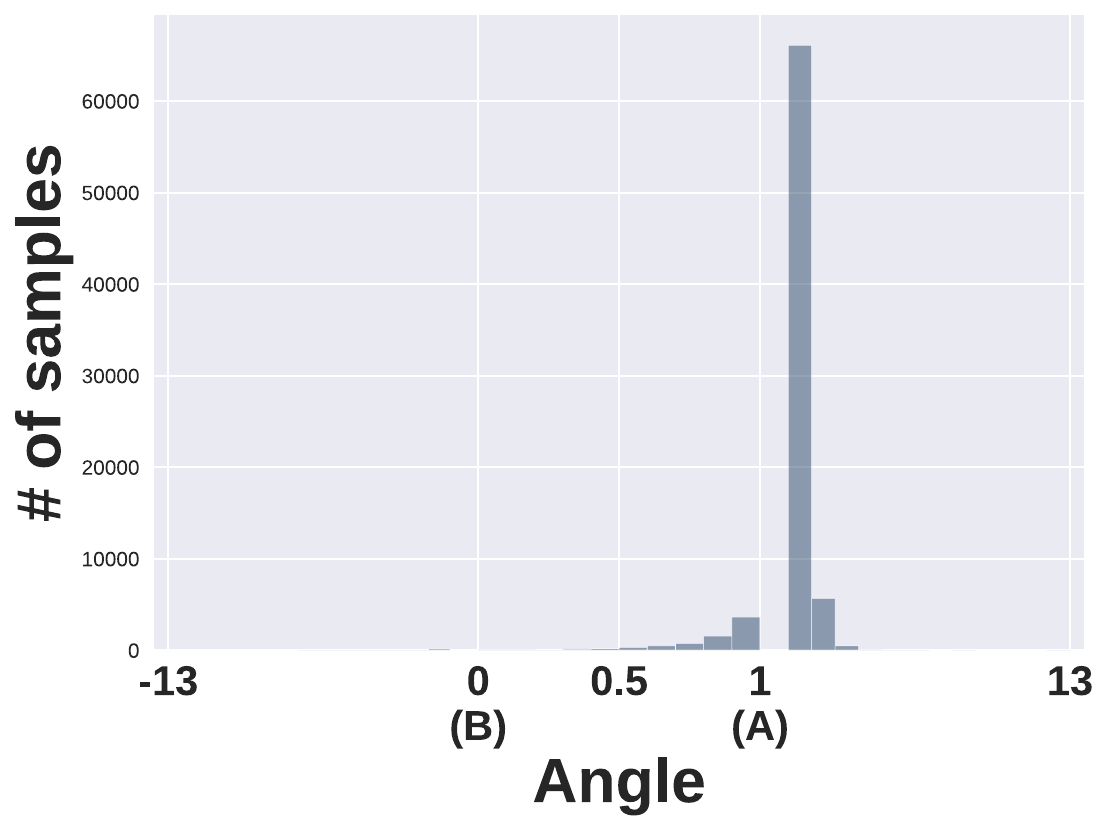}
         \label{fig:mpnet_difference}
     \end{subfigure}
    ~
     \begin{subfigure}[b]{0.32\textwidth}
         \centering
         \includegraphics[width=\textwidth]{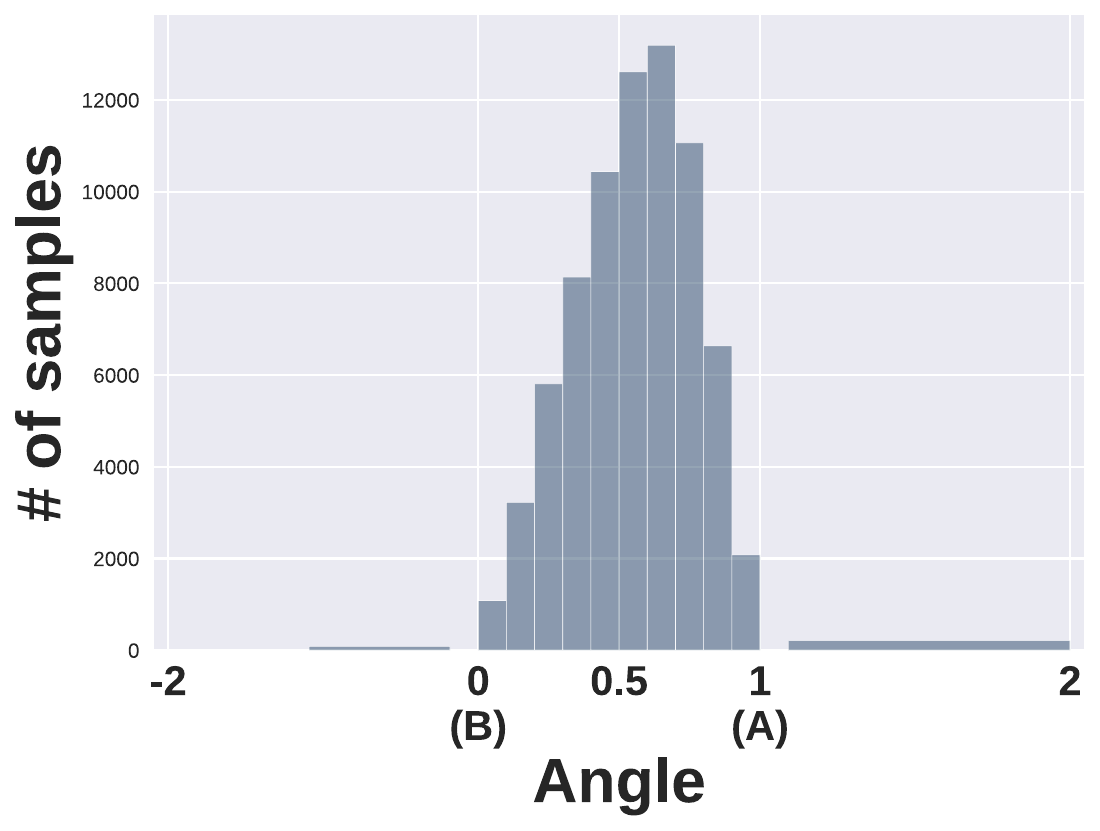}
         \label{fig:mpnet_union}
     \end{subfigure}
     \vspace{-4mm}
     \subcaption{\textbf{SBERT-L}}
    \label{fig:mpnet_projection}
    \end{subfigure}
     \hfill

     \begin{subfigure}[b]{1.0\textwidth}
    \centering
    \begin{subfigure}[b]{0.32\textwidth}
         \centering
         \includegraphics[width=1.0\textwidth]{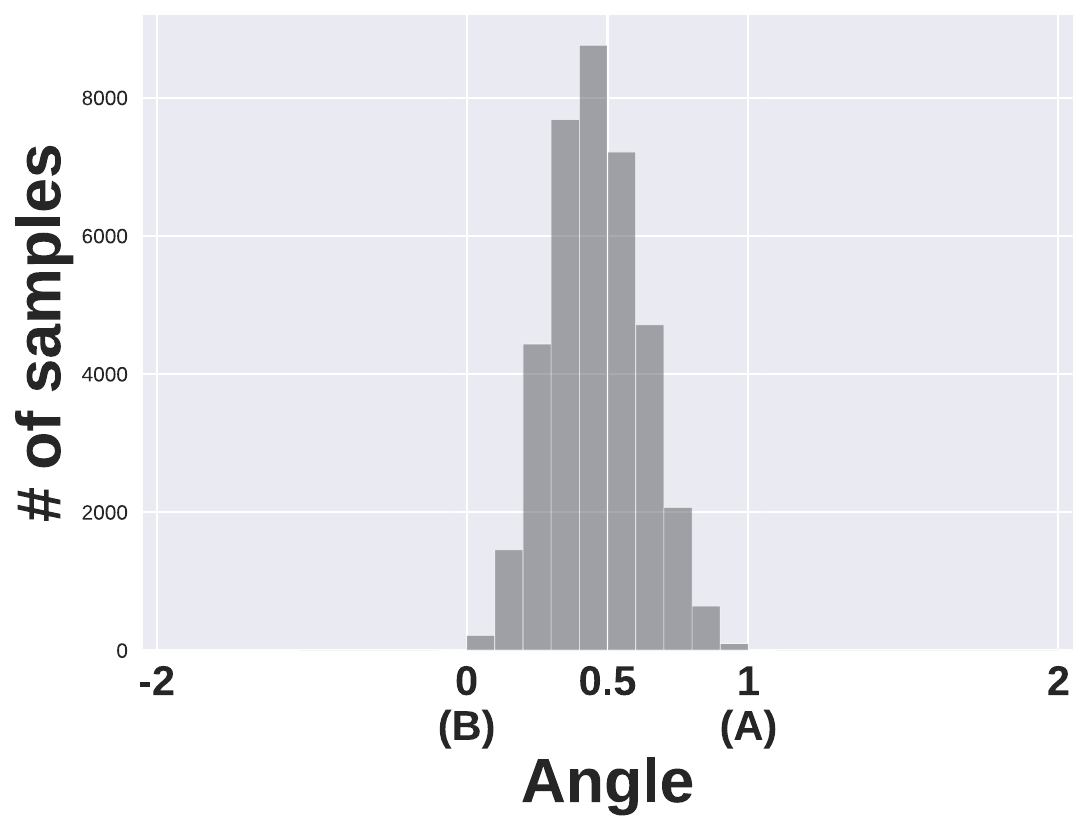}
         \label{fig:infersent_overlap}
     \end{subfigure}
    ~
     \begin{subfigure}[b]{0.32\textwidth}
         \centering
        \includegraphics[width=1.0\textwidth]{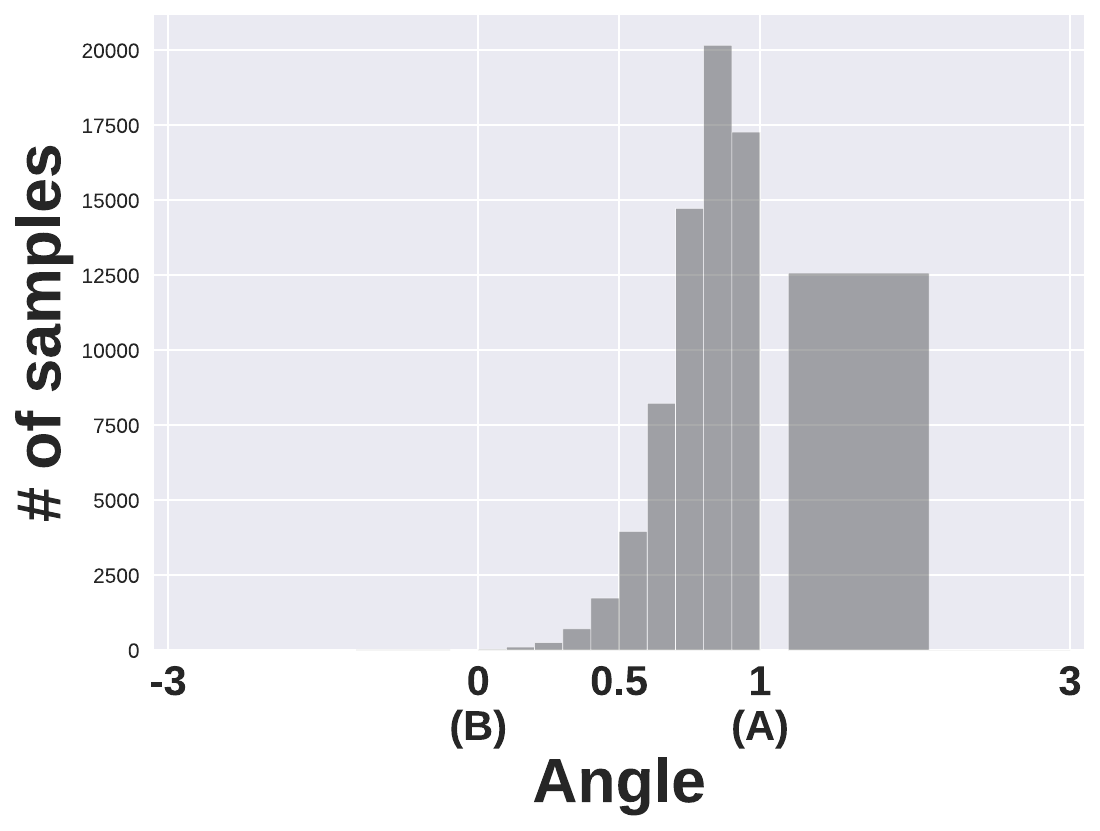}
         \label{fig:infersent_difference}
     \end{subfigure}
    ~
     \begin{subfigure}[b]{0.32\textwidth}
         \centering
         \includegraphics[width=\textwidth]{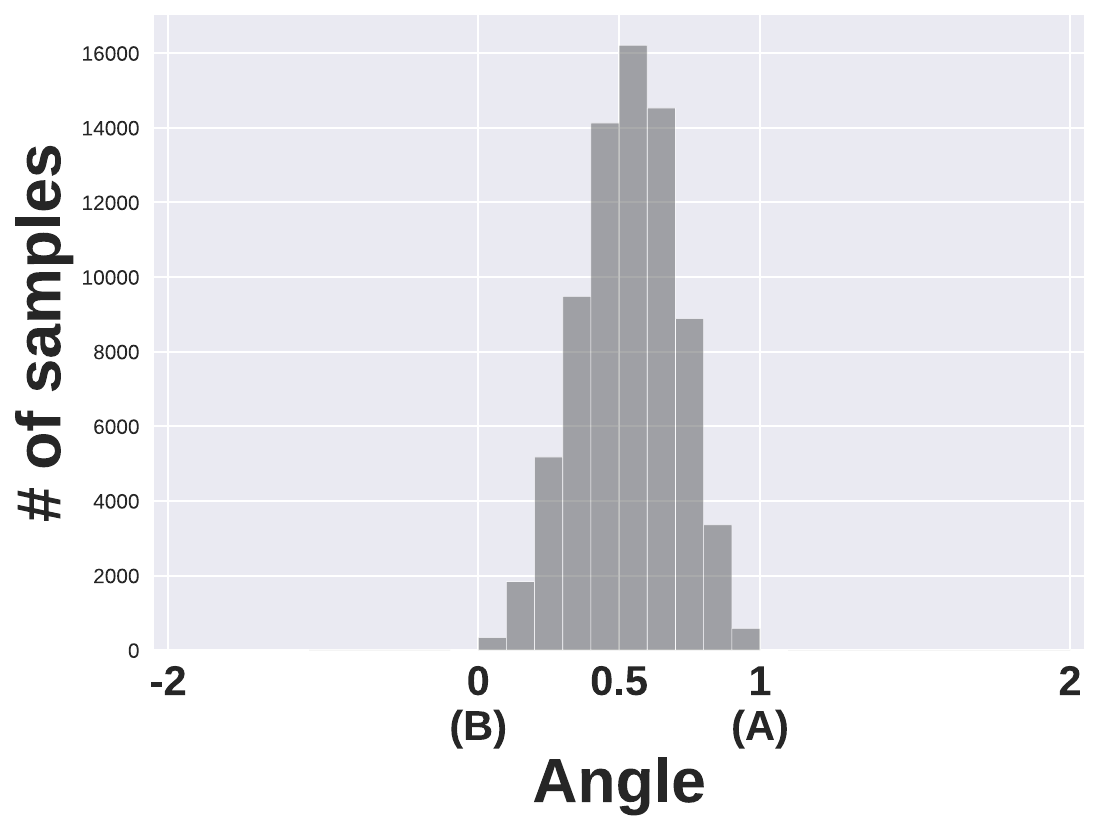}
         \label{fig:infersent_union}
     \end{subfigure}
     \vspace{-4mm}
     \subcaption{\textbf{InferSent}}
    \label{fig:infersent_projection}
    \end{subfigure}
     \hfill

    \begin{subfigure}[b]{1.0\textwidth}
    \centering
    \begin{subfigure}[b]{0.32\textwidth}
         \centering
         \includegraphics[width=1.0\textwidth]{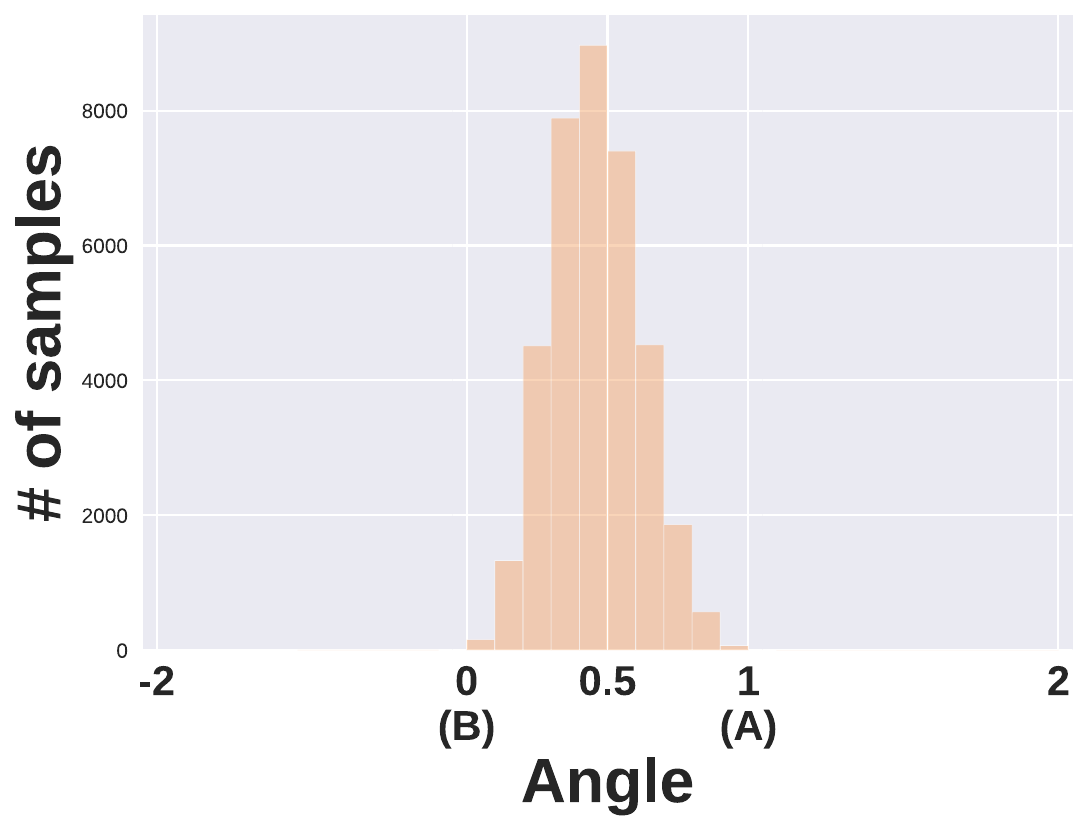}
         \label{fig:laser_overlap}
     \end{subfigure}
    ~
     \begin{subfigure}[b]{0.32\textwidth}
         \centering
        \includegraphics[width=1.0\textwidth]{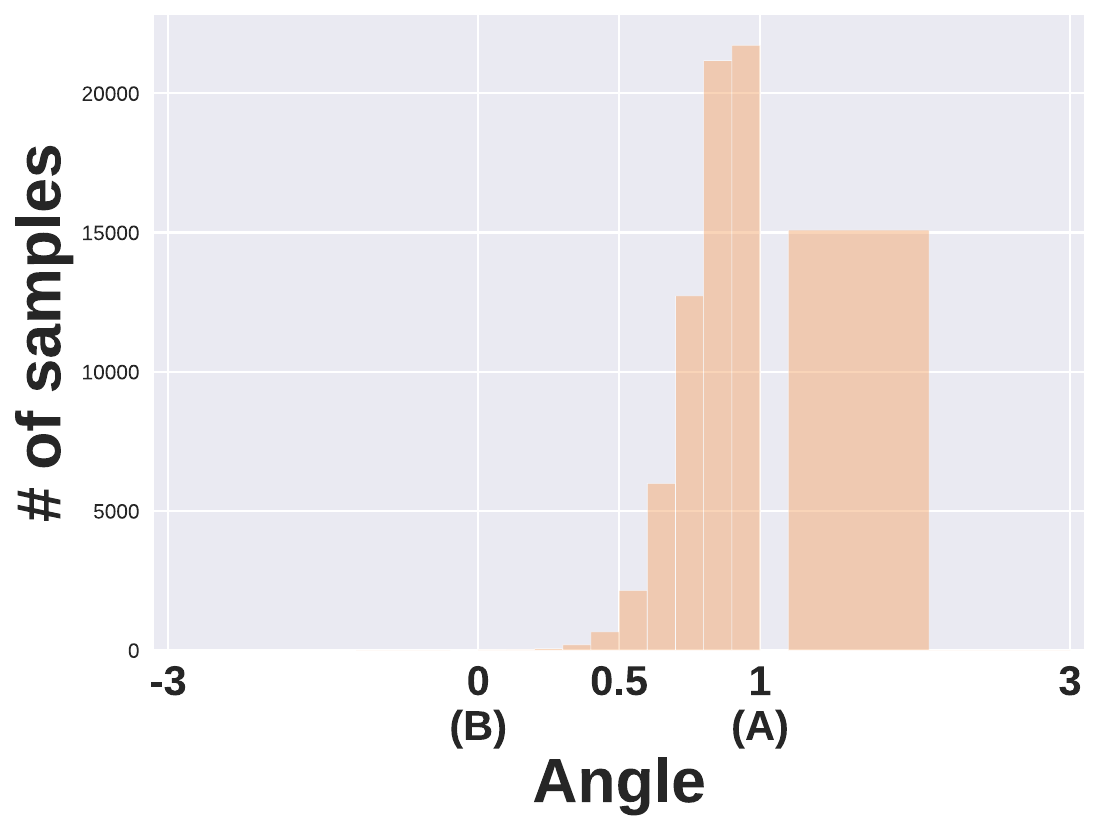}
         \label{fig:laser_difference}
     \end{subfigure}
    ~
     \begin{subfigure}[b]{0.32\textwidth}
         \centering
         \includegraphics[width=\textwidth]{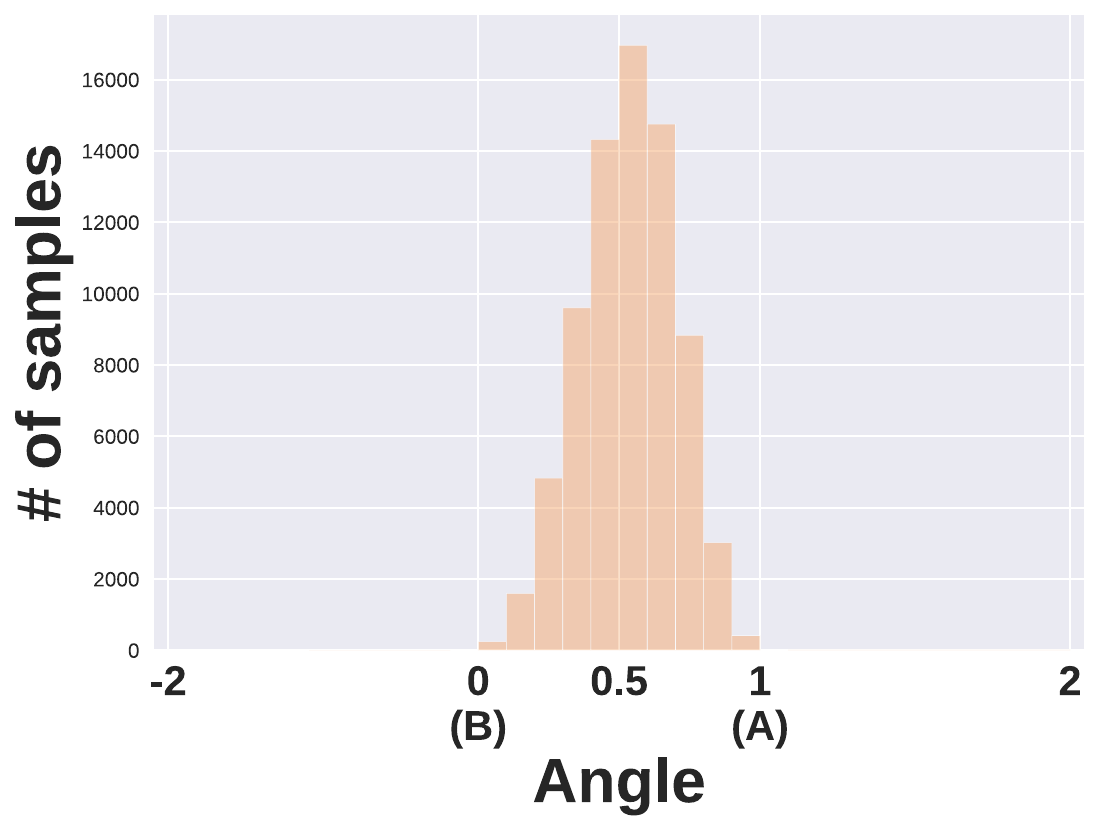}
         \label{fig:laser_union}
     \end{subfigure}
     \vspace{-4mm}
     \subcaption{\textbf{LASER}}
    \label{fig:laser_projection}
    \end{subfigure}
     
    \caption{
    Histogram of the angle between the projection of the target sentence embedding (onto the plane of the embeddings of the input sentences $A$ and $B$) and the embedding of sentence $B$. The target sentence embedding can be:\\
        \textbf{TextOverlap (Left)}:
        The projection embedding mostly lie in the ``middle'' of the embeddings of the input sentences as described in criterion C2 \ref{para:h2_results} and follows  our expectation shown in figure \ref{fig:projection_expectation}b \\
        \textbf{TextDifference (Middle)}: The projection embedding is mostly bounded by a small angle around the embedding of the input sentence $A$ (refer C5\ref{para:h5_results}).
        This follows our expectation shown in figure \ref{fig:projection_expectation}a \\
        \textbf{TextUnion (Right)}: The projection embedding mostly lie in ``middle'' of the input sentence embeddings (refer \ref{para:h6_results}). This follows  our expectation shown in figure \ref{fig:projection_expectation}c-middle.\\
        We normalize this angle such that the angle between the embeddings of sentences $A$ and $B$ is consistently 1 (refer \ref{para:h2_results} for details). 
    }
    \label{fig:all_projection_results}
\end{figure*}

\begin{figure*}[!t]\ContinuedFloat
    \centering

    \begin{subfigure}[b]{1.0\textwidth}
    \centering
    \begin{subfigure}[b]{0.32\textwidth}
         \centering
         \begin{tikzpicture}
            \node[anchor=north] at (current bounding box.north) {\textbf{TextOverlap}};
        \end{tikzpicture}
         \includegraphics[width=1.0\textwidth]{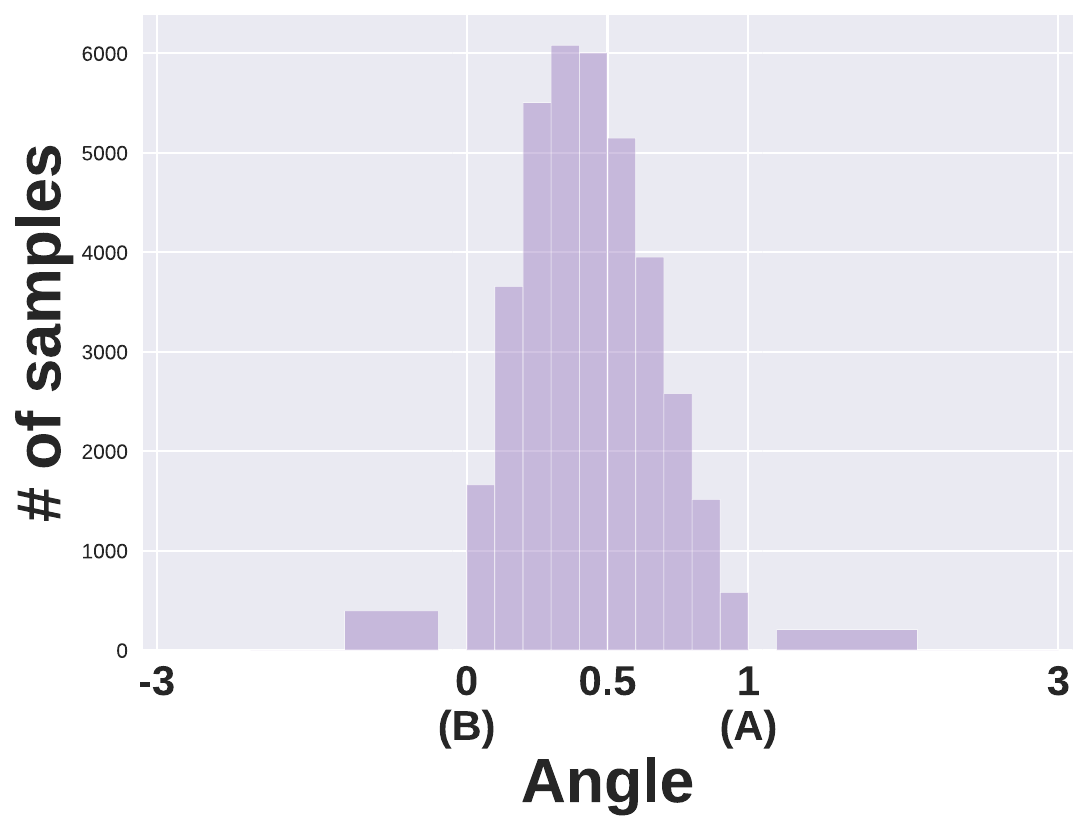}
         \label{fig:use_overlap}
     \end{subfigure}
    ~
     \begin{subfigure}[b]{0.32\textwidth}
         \centering
         \begin{tikzpicture}
            \node[anchor=north] at (current bounding box.north) {\textbf{TextDifference}};
        \end{tikzpicture}
        \includegraphics[width=1.0\textwidth]{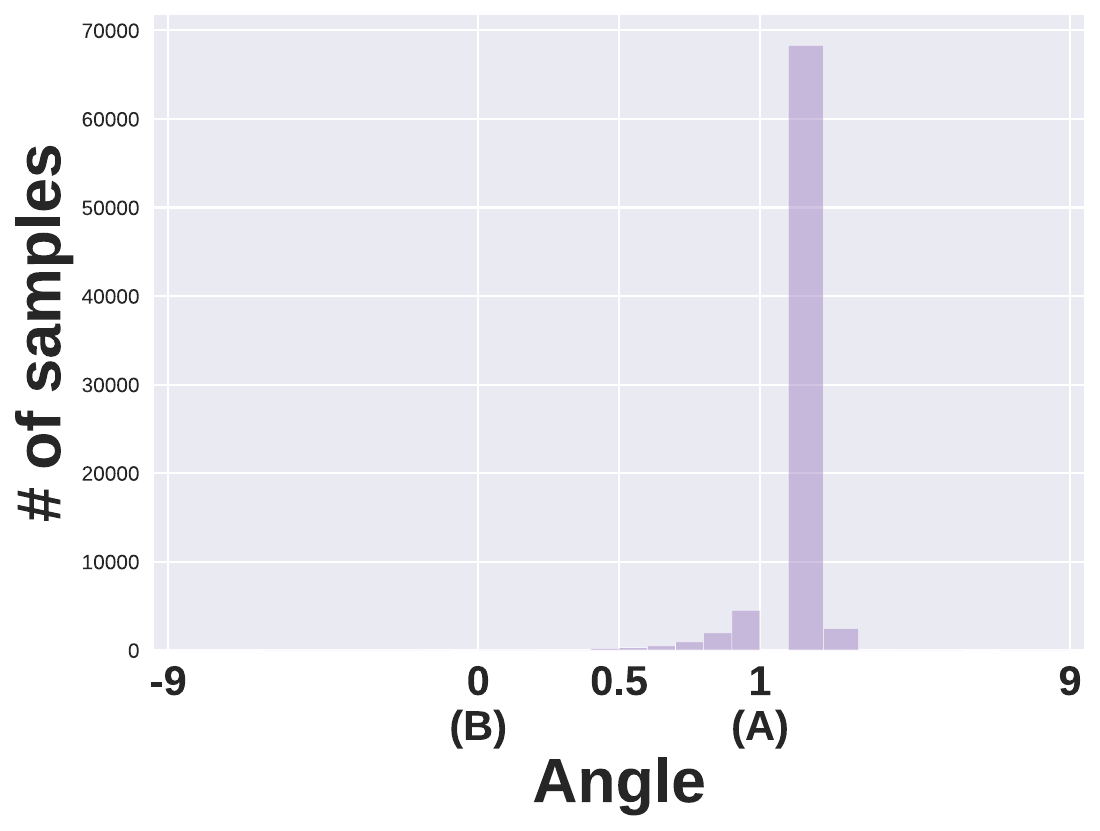}
         \label{fig:use_difference}
     \end{subfigure}
    ~
     \begin{subfigure}[b]{0.32\textwidth}
         \centering
         \begin{tikzpicture}
            \node[anchor=north] at (current bounding box.north) {\textbf{TextUnion}};
        \end{tikzpicture}
         \includegraphics[width=\textwidth]{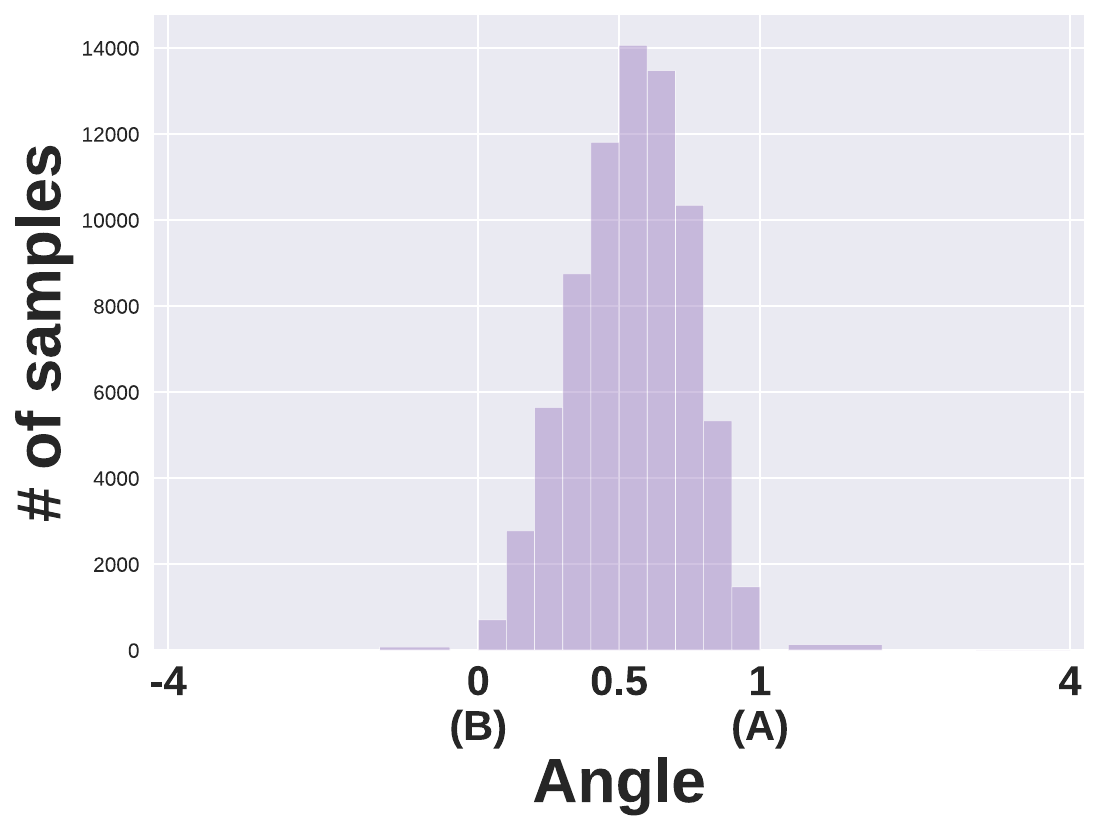}
         \label{fig:use_union}
     \end{subfigure}

     \vspace{-4mm}
     \subcaption{\textbf{USE}}
    \label{fig:use_projection}
    
     \end{subfigure}
     \hfill

     \begin{subfigure}[b]{1.0\textwidth}
    \centering
    \begin{subfigure}[b]{0.32\textwidth}
         \centering
         \includegraphics[width=1.0\textwidth]{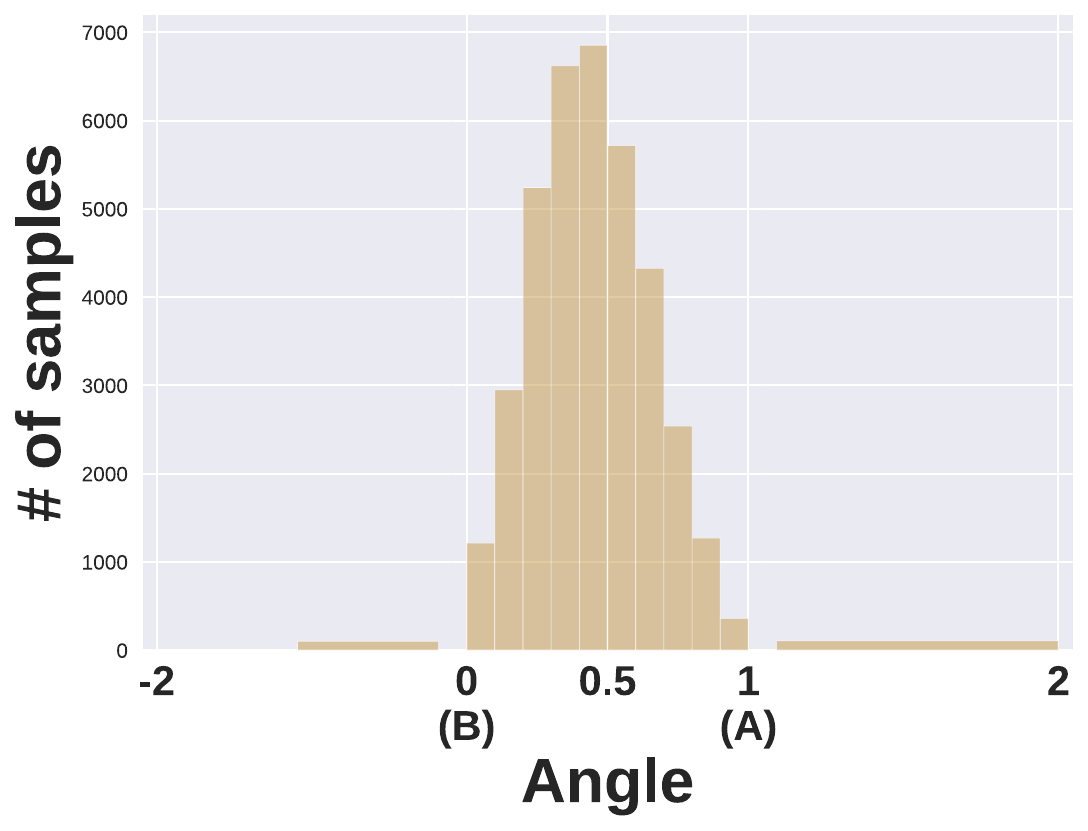}
         \label{fig:roberta_overlap}
     \end{subfigure}
    ~
     \begin{subfigure}[b]{0.32\textwidth}
         \centering
        \includegraphics[width=1.0\textwidth]{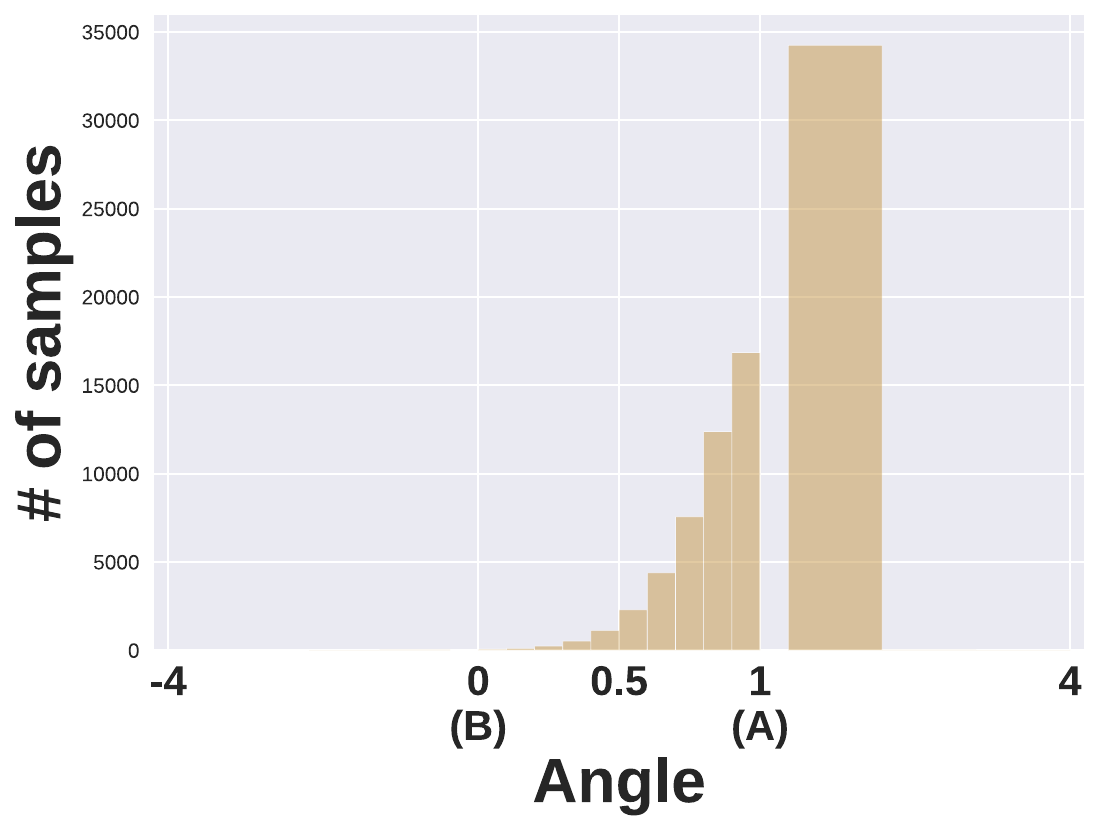}
         \label{fig:roberta_difference}
     \end{subfigure}
    ~
     \begin{subfigure}[b]{0.32\textwidth}
         \centering
         \includegraphics[width=1.0\textwidth]{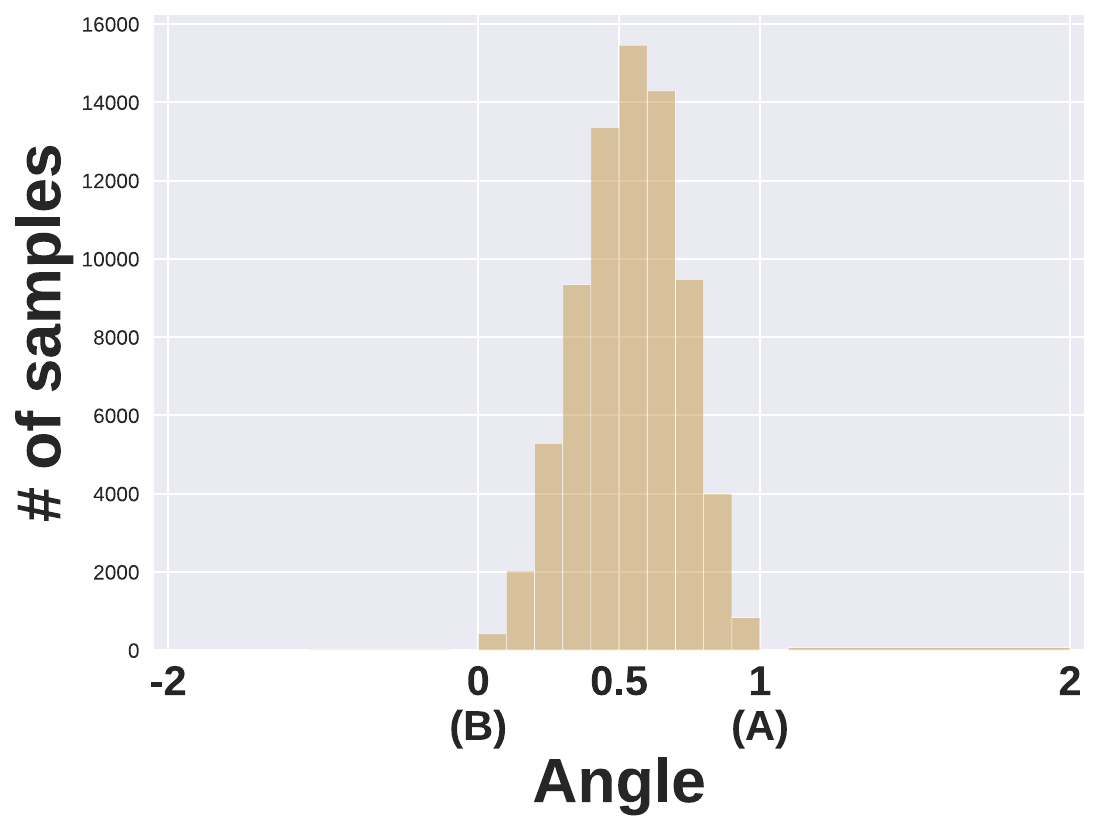}
         \label{fig:roberta_union}
     \end{subfigure}

     \vspace{-4mm}
     \subcaption{\textbf{RoBERTa}}
    \label{fig:roberta_projection}
     \end{subfigure}
    \hfill
    
    \begin{subfigure}[b]{1.0\textwidth}
    \centering
    \begin{subfigure}[b]{0.32\textwidth}
         \centering
         \includegraphics[width=1.0\textwidth]{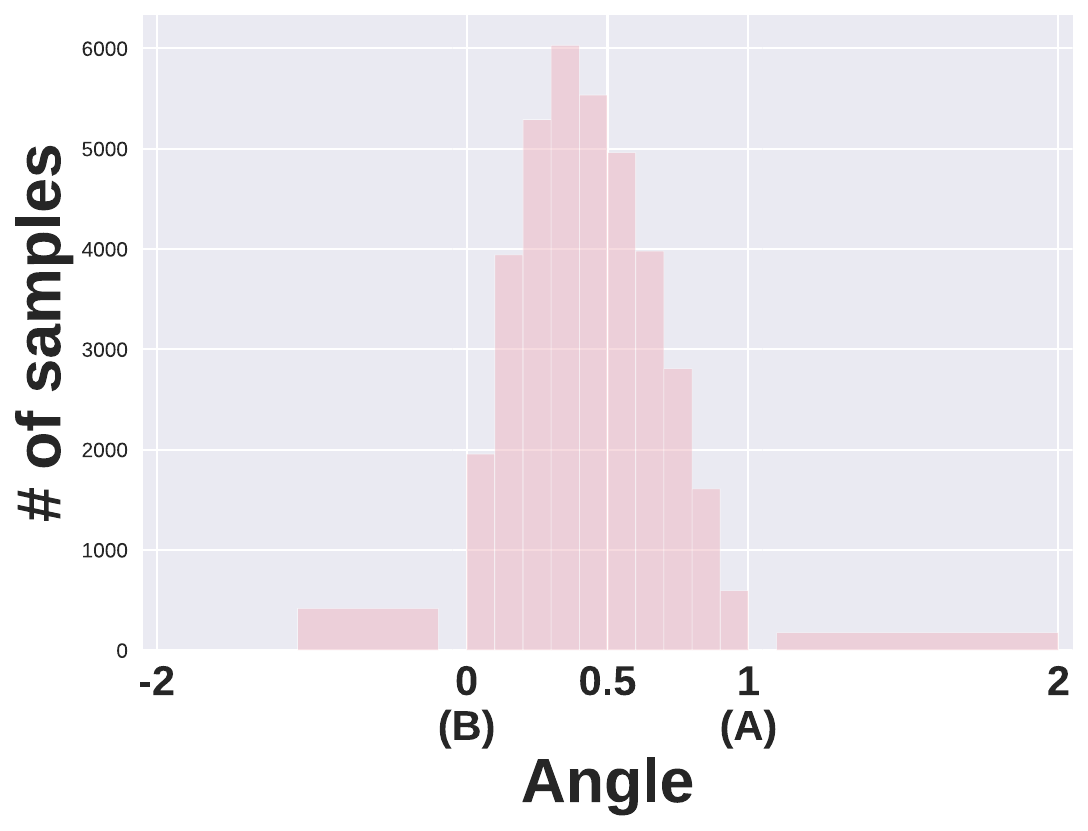}
         \label{fig:simcse_overlap}
     \end{subfigure}
    ~
     \begin{subfigure}[b]{0.32\textwidth}
         \centering
        \includegraphics[width=1.0\textwidth]{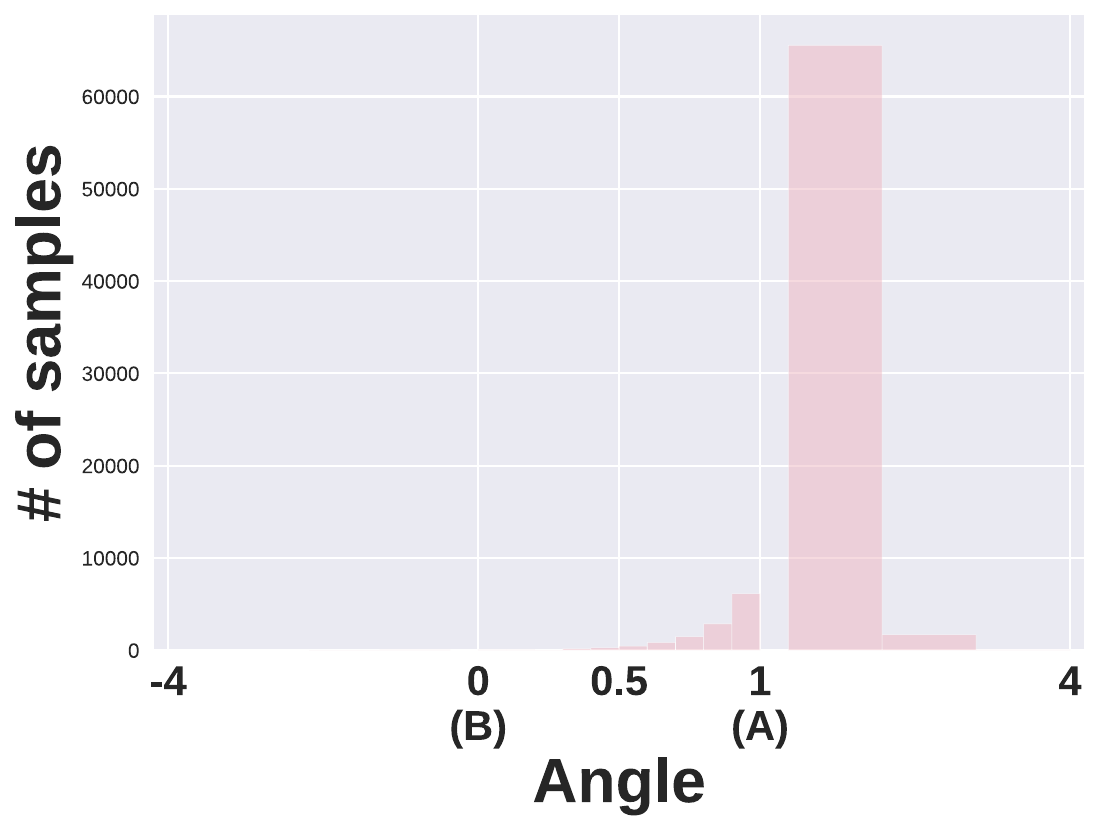}
         \label{fig:simcse_difference}
     \end{subfigure}
    ~
     \begin{subfigure}[b]{0.32\textwidth}
         \centering
         \includegraphics[width=1.0\textwidth]{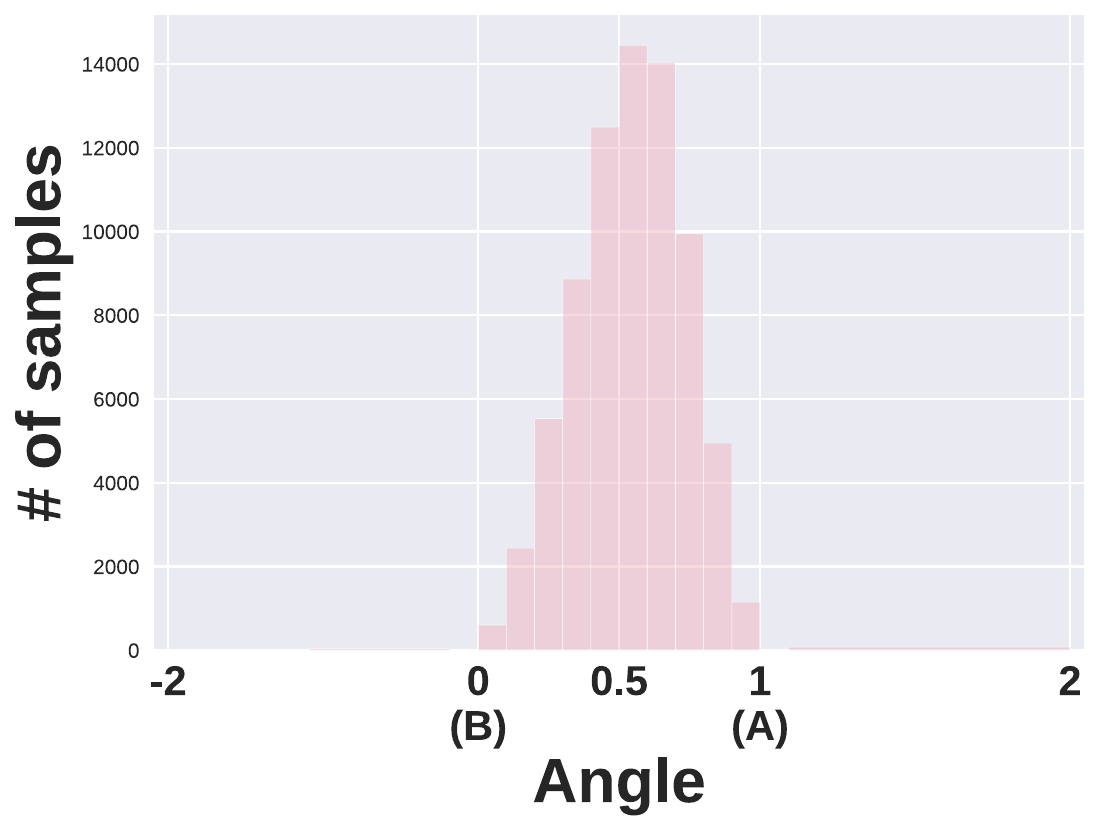}
         \label{fig:simcse_union}
     \end{subfigure}
     \vspace{-4mm}
     \subcaption{\textbf{SimCSE}}
    \label{fig:simcse_projection}
     \end{subfigure}
     \hfill

    \begin{subfigure}[b]{1.0\textwidth}     
    \centering
    \begin{subfigure}[b]{0.32\textwidth}
         \centering
         \includegraphics[width=1.0\textwidth]{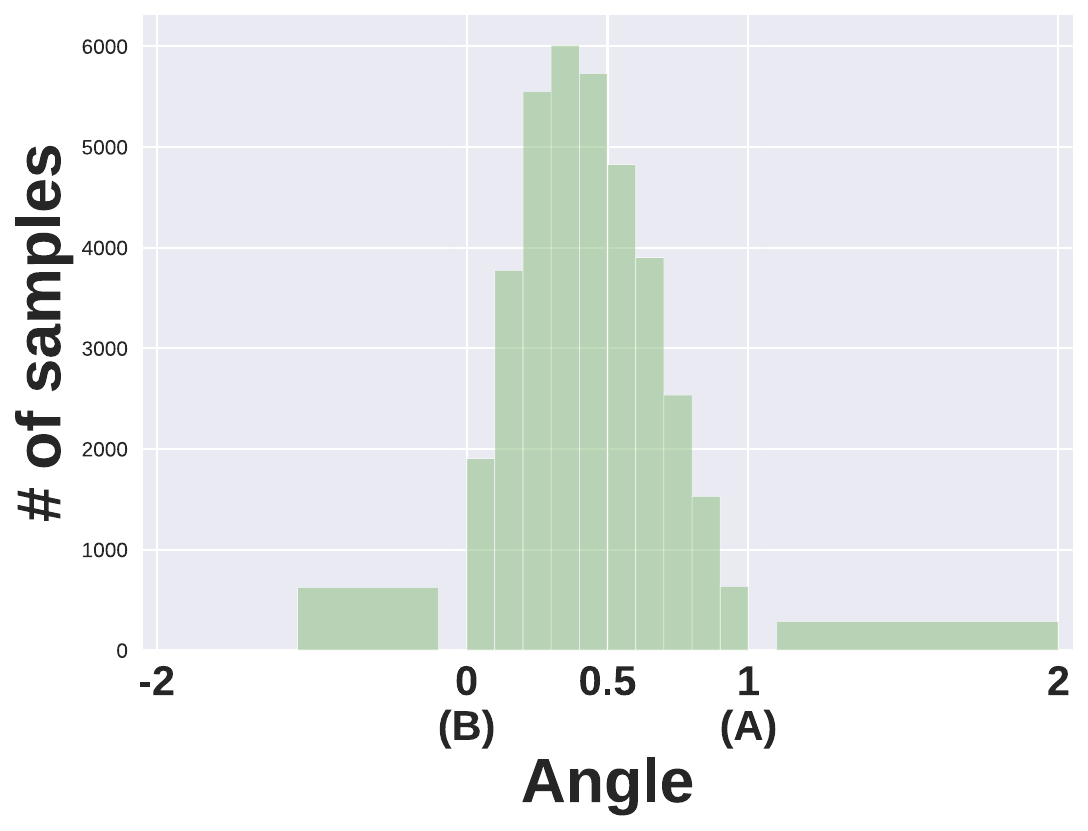}
         \label{fig:llama_overlap}
     \end{subfigure}
    ~
     \begin{subfigure}[b]{0.32\textwidth}
         \centering
        \includegraphics[width=1.0\textwidth]{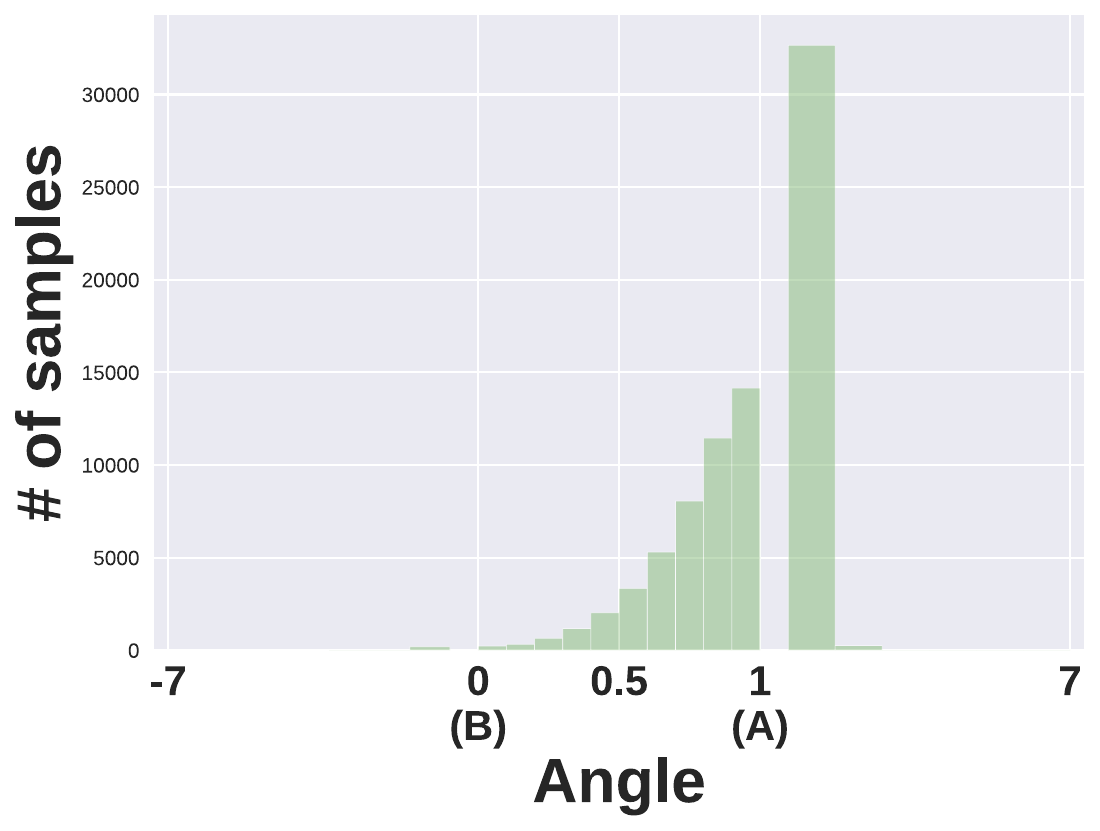}
         \label{fig:llama_difference}
     \end{subfigure}
    ~
     \begin{subfigure}[b]{0.32\textwidth}
         \centering
         \includegraphics[width=\textwidth]{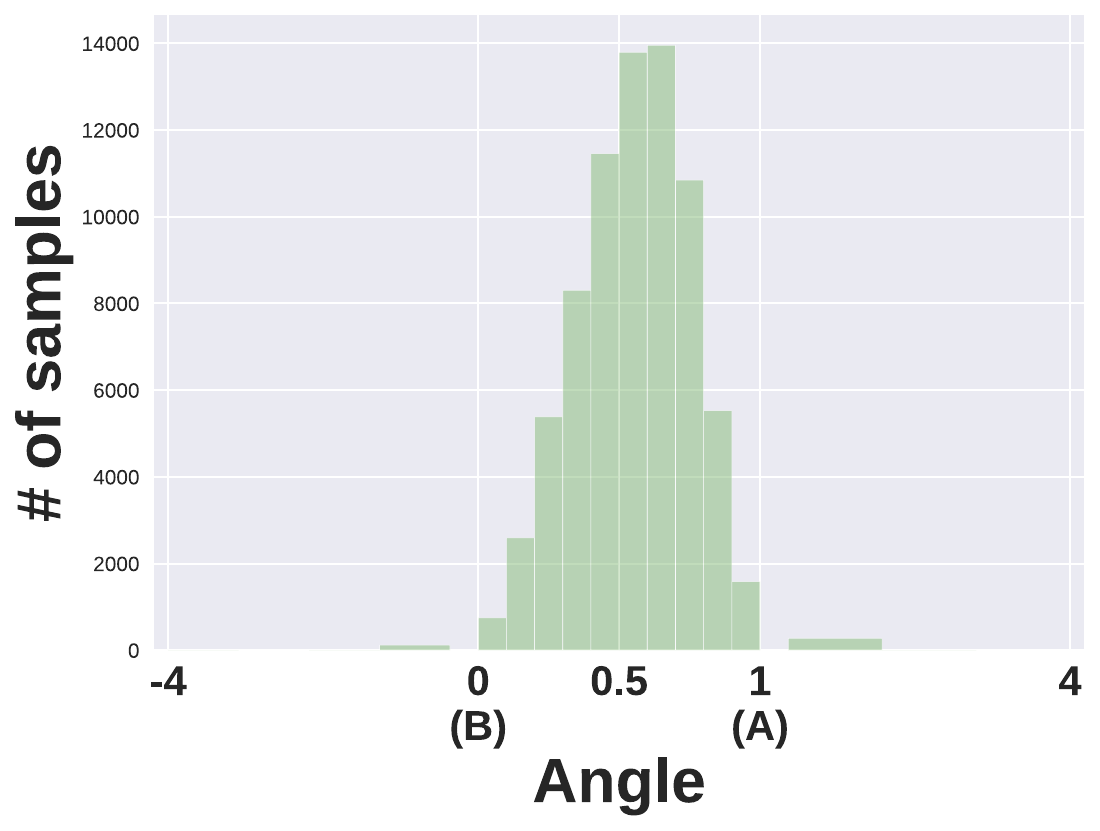}
         \label{fig:llama_union}
     \end{subfigure}
     \vspace{-4mm}
     \subcaption{\textbf{LLaMA2}}
    \label{fig:llama2_projection}
    \end{subfigure}
     
    \caption*{Figure~\ref{fig:all_projection_results}: Histogram of the angle between the projection of the target sentence embedding (onto the plane of the embeddings of the input sentences $A$ and $B$) and the embedding of sentence $B$. The target sentence embedding can be:\\
        \textbf{TextOverlap (Left)}:
        The projection embedding mostly lie in the ``middle'' of the embeddings of the input sentences as described in criterion C2 \ref{para:h2_results} and follows  our expectation shown in figure \ref{fig:projection_expectation}b \\
        \textbf{TextDifference (Middle)}: The projection embedding is mostly bounded by a small angle around the embedding of the input sentence $A$ (refer C5\ref{para:h5_results}).
        This follows our expectation shown in figure \ref{fig:projection_expectation}a \\
        \textbf{TextUnion (Right)}: The projection embedding mostly lie in ``middle'' of the input sentence embeddings (refer \ref{para:h6_results}). This follows  our expectation shown in figure \ref{fig:projection_expectation}c-middle.\\
        We normalize this angle such that the angle between the embeddings of sentences $A$ and $B$ is consistently 1 (refer \ref{para:h2_results} for details).}
\end{figure*}

\begin{figure*}[!t]\ContinuedFloat
    \centering

     \begin{subfigure}[b]{1.0\textwidth}     \centering
    \begin{subfigure}[b]{0.32\textwidth}
         \centering
         \begin{tikzpicture}
            \node[anchor=north] at (current bounding box.north) {\textbf{TextOverlap}};
        \end{tikzpicture}
         \includegraphics[width=1.0\textwidth]{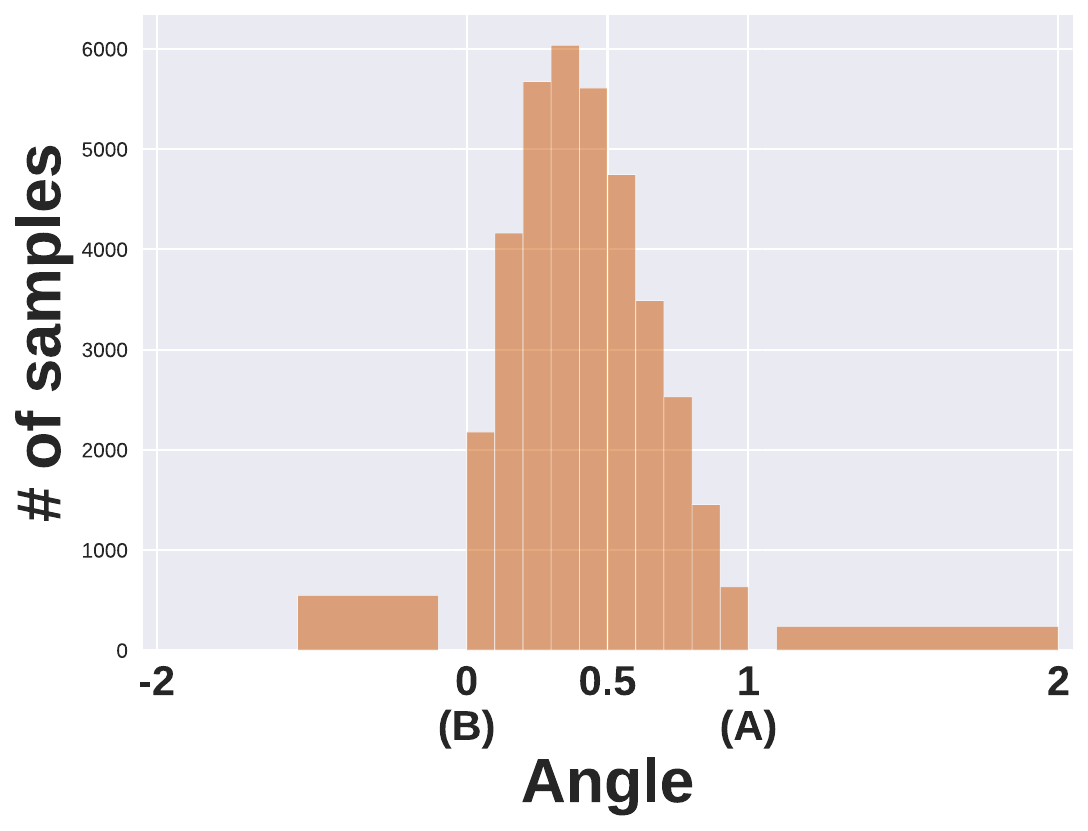}
         \label{fig:llama3_overlap}
     \end{subfigure}
    ~
     \begin{subfigure}[b]{0.32\textwidth}
         \centering
         \begin{tikzpicture}
            \node[anchor=north] at (current bounding box.north) {\textbf{TextDifference}};
        \end{tikzpicture}
        \includegraphics[width=1.0\textwidth]{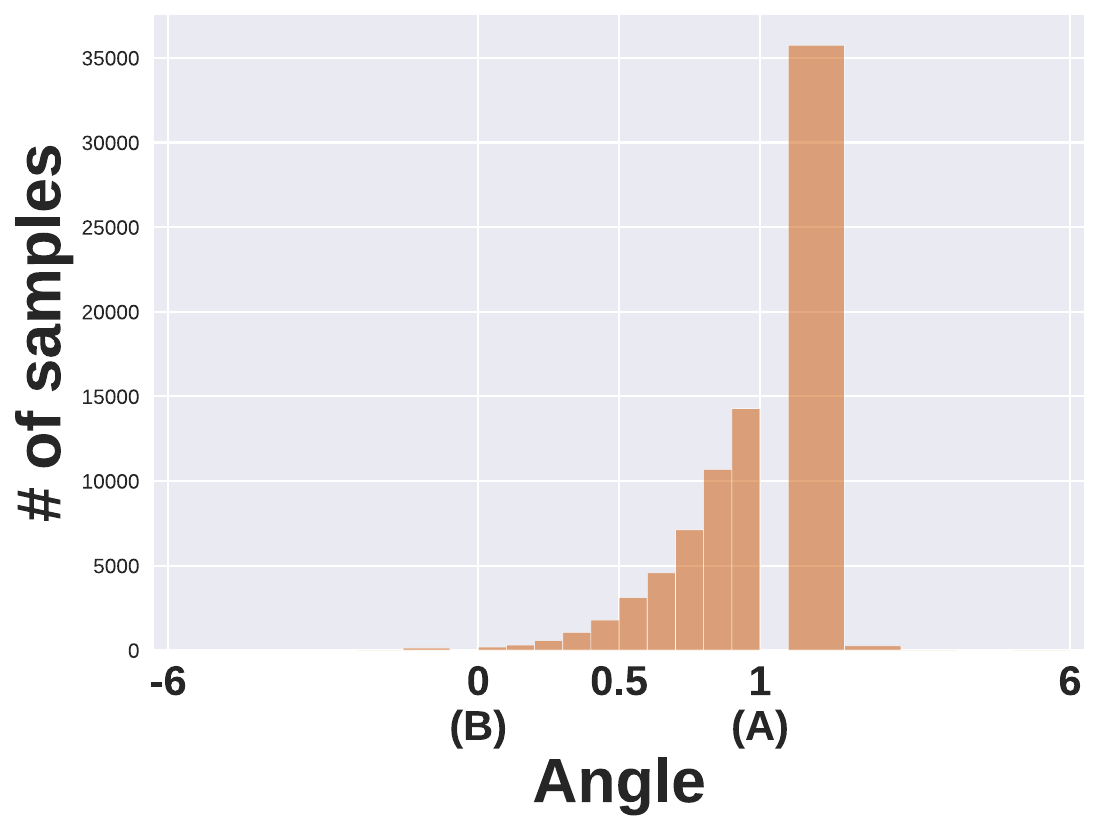}
         \label{fig:llama3_difference}
     \end{subfigure}
    ~
     \begin{subfigure}[b]{0.32\textwidth}
         \centering
         \begin{tikzpicture}
            \node[anchor=north] at (current bounding box.north) {\textbf{TextUnion}};
        \end{tikzpicture}
         \includegraphics[width=\textwidth]{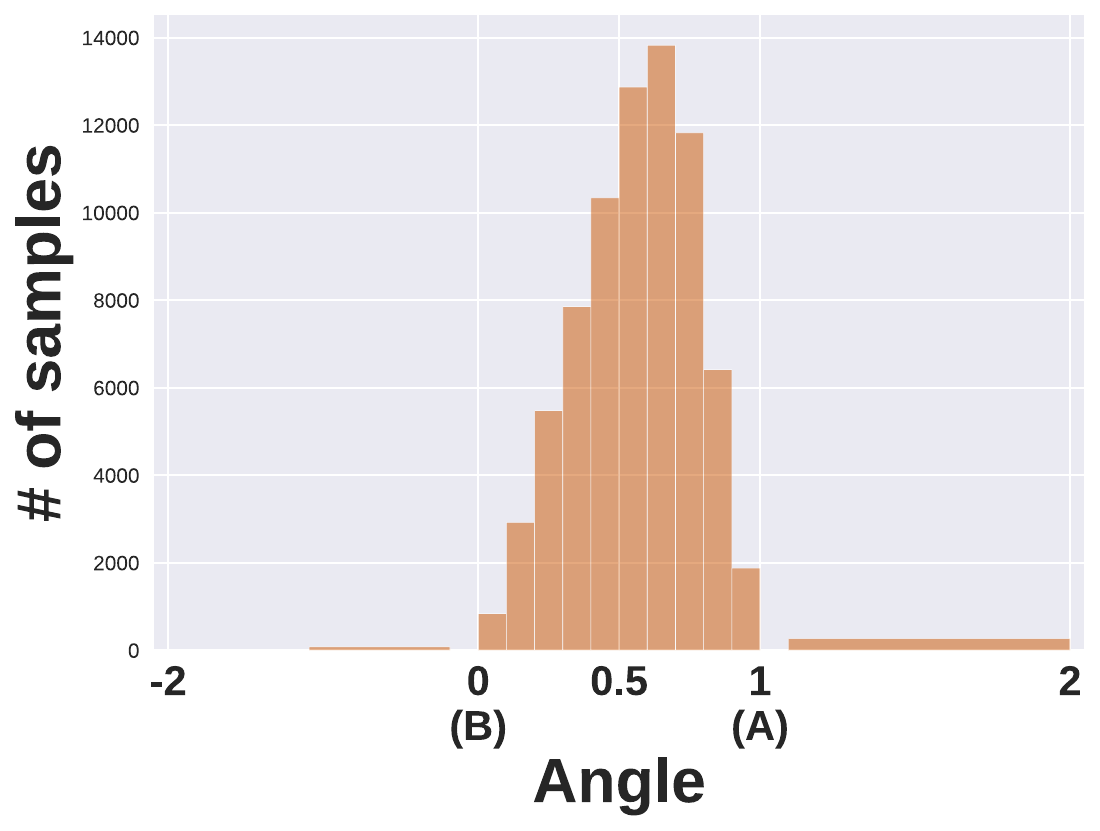}
         \label{fig:llama3_union}
     \end{subfigure}
     \vspace{-4mm}
     \subcaption{\textbf{LLaMA3}}
    \label{fig:llama3_projection}
     \end{subfigure}
     \hfill
     
    \begin{subfigure}[b]{1.0\textwidth}     \centering
    \begin{subfigure}[b]{0.32\textwidth}
         \centering
         \includegraphics[width=1.0\textwidth]{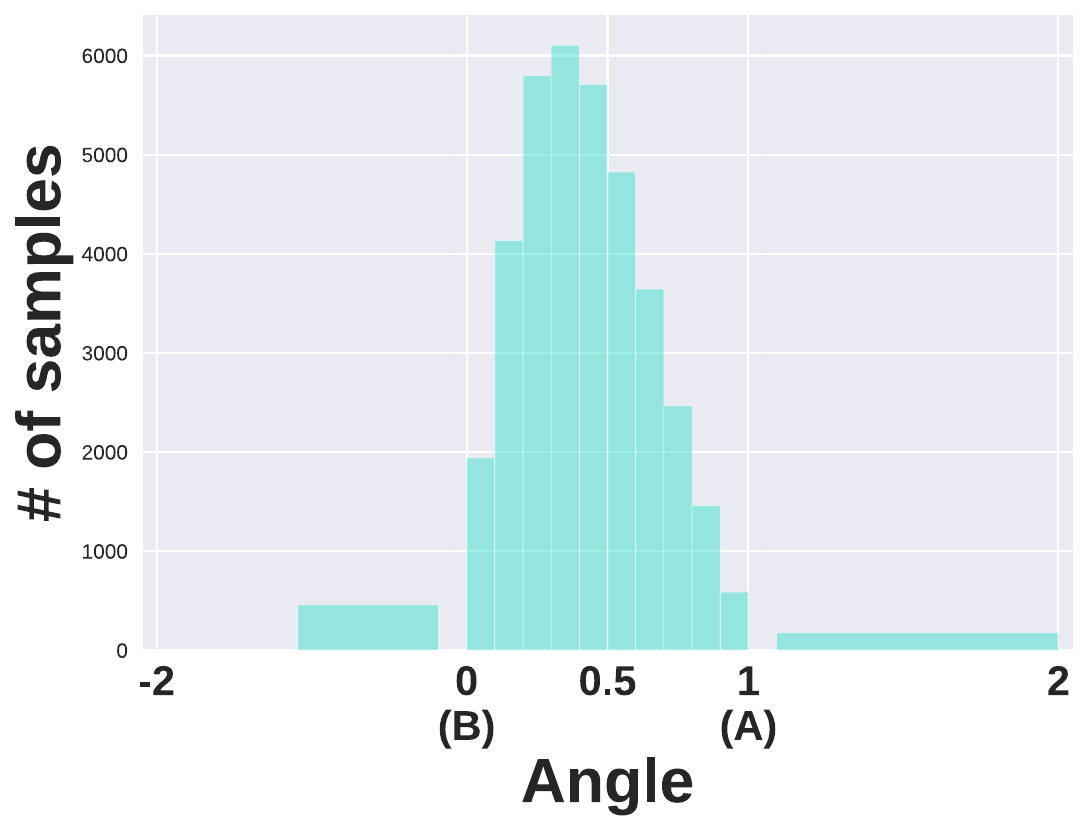}
         \label{fig:mistral_overlap}
     \end{subfigure}
    ~
     \begin{subfigure}[b]{0.32\textwidth}
         \centering
        \includegraphics[width=1.0\textwidth]{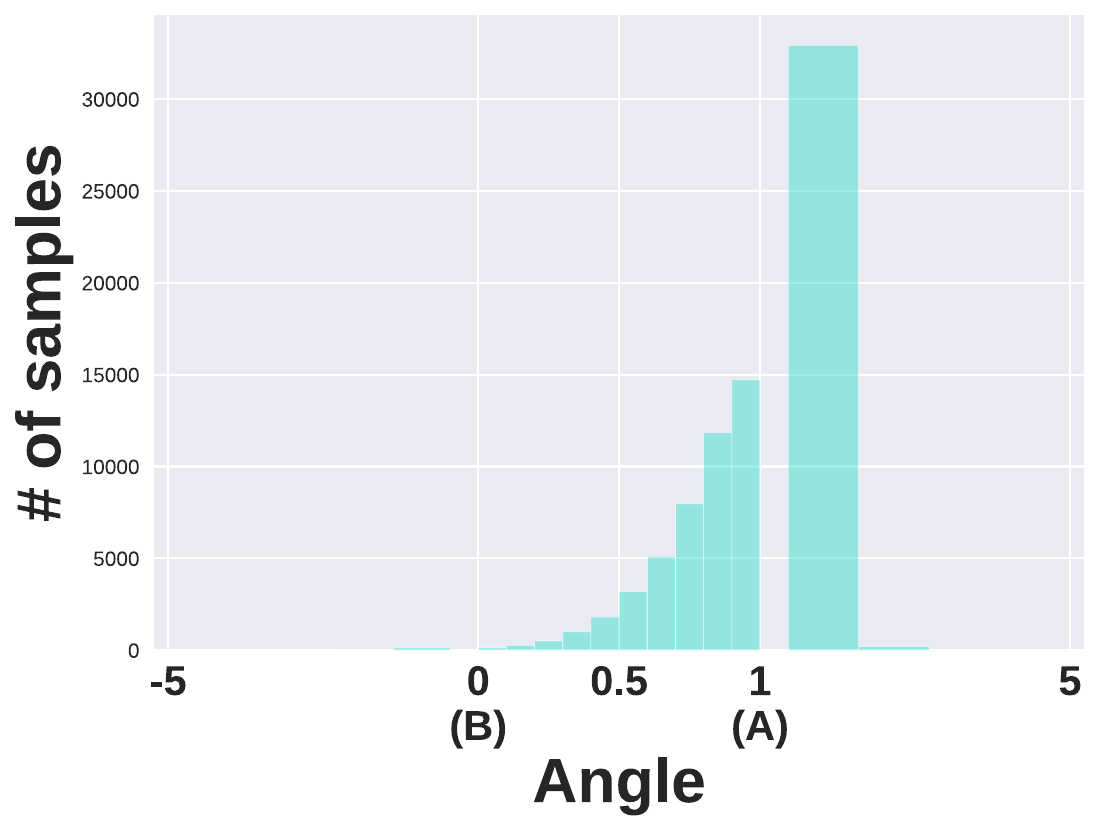}
         \label{fig:mistral_difference}
     \end{subfigure}
    ~
     \begin{subfigure}[b]{0.32\textwidth}
         \centering
         \includegraphics[width=\textwidth]{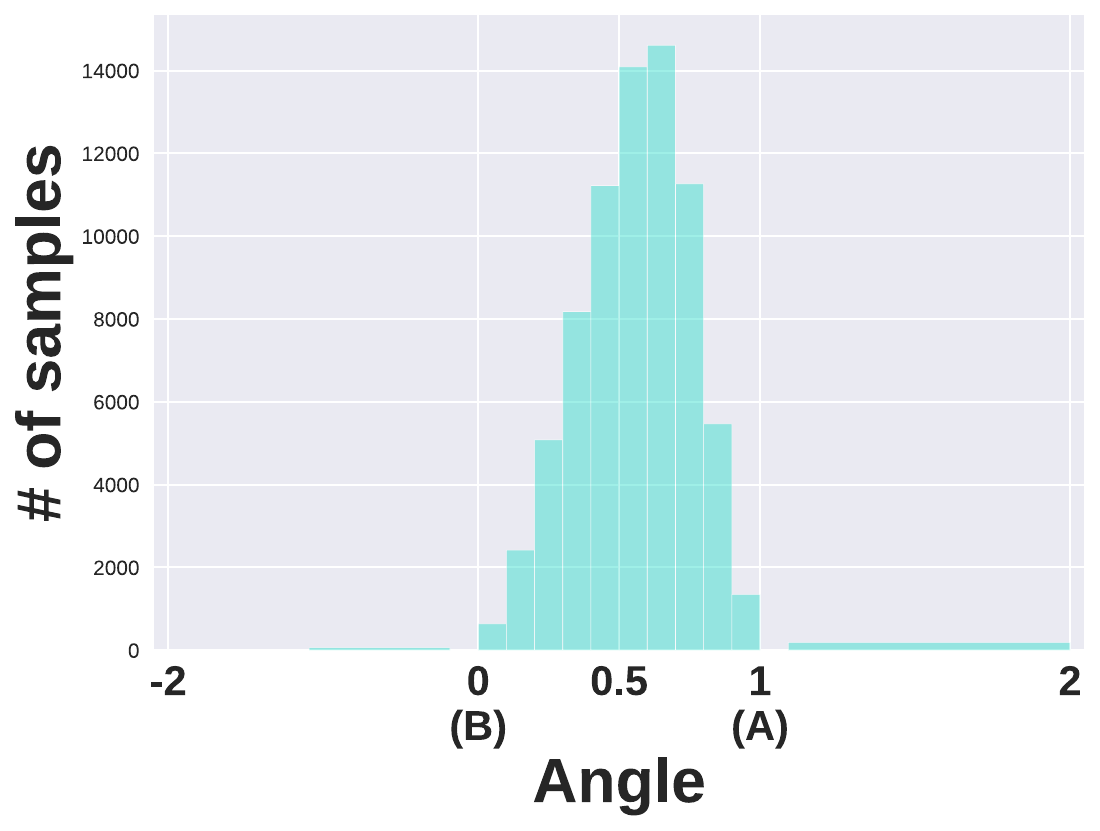}
         \label{fig:mistral_union}
     \end{subfigure}
     \vspace{-4mm}
     \subcaption{\textbf{Mistral}}
    \label{fig:mistral_projection}
     \end{subfigure}
     \hfill

    \begin{subfigure}[b]{1.0\textwidth}     \centering
    \begin{subfigure}[b]{0.32\textwidth}
         \centering
         \includegraphics[width=1.0\textwidth]{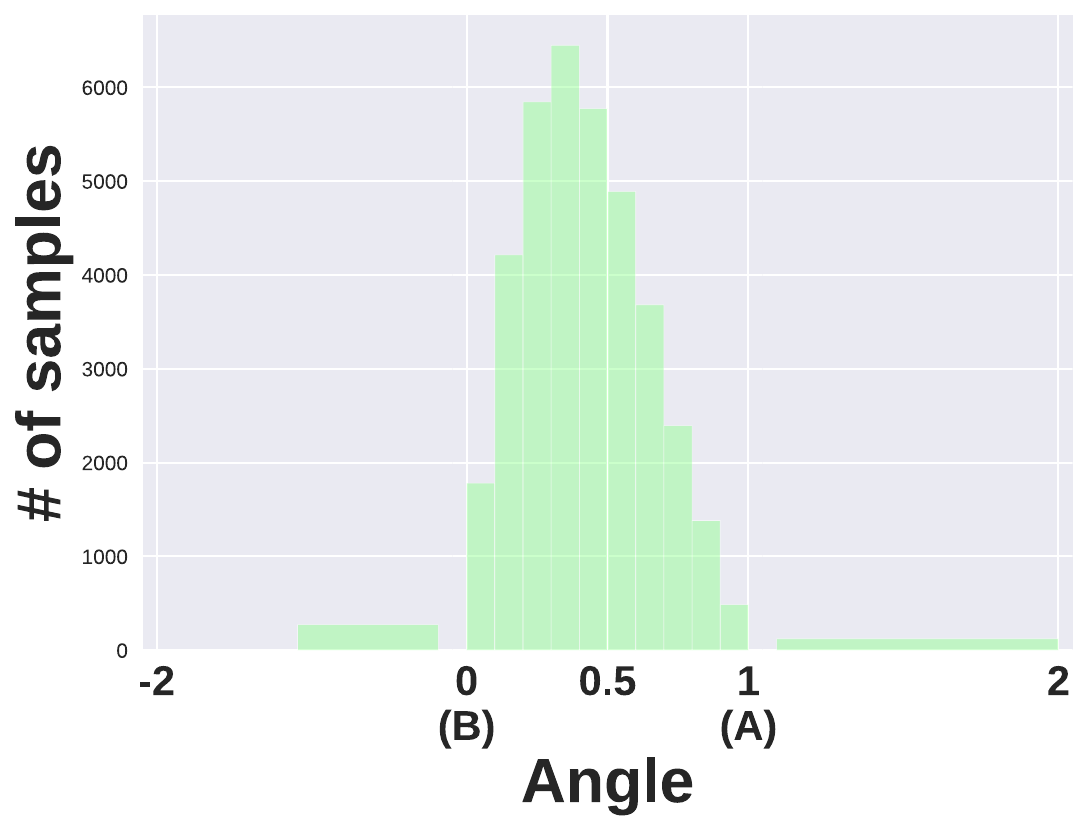}
         \label{fig:olmo_overlap}
     \end{subfigure}
    ~
     \begin{subfigure}[b]{0.32\textwidth}
         \centering
        \includegraphics[width=1.0\textwidth]{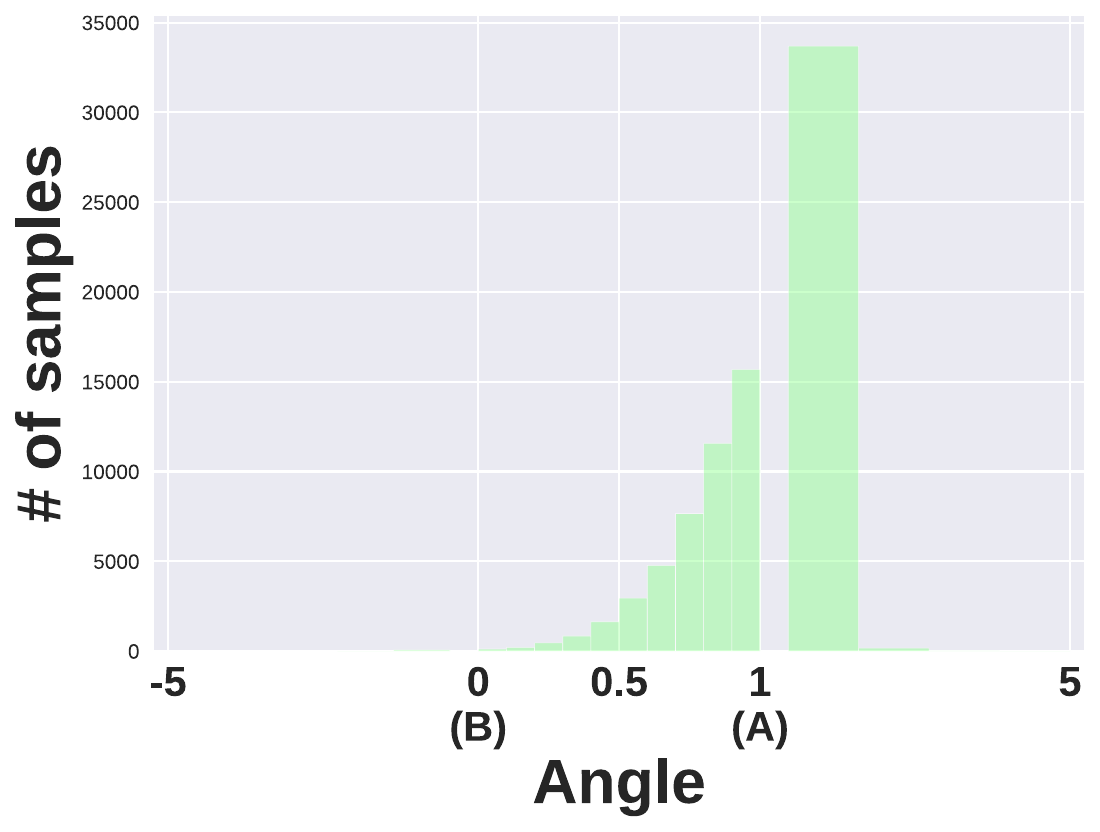}
         \label{fig:olmo_difference}
     \end{subfigure}
    ~
     \begin{subfigure}[b]{0.32\textwidth}
         \centering
         \includegraphics[width=\textwidth]{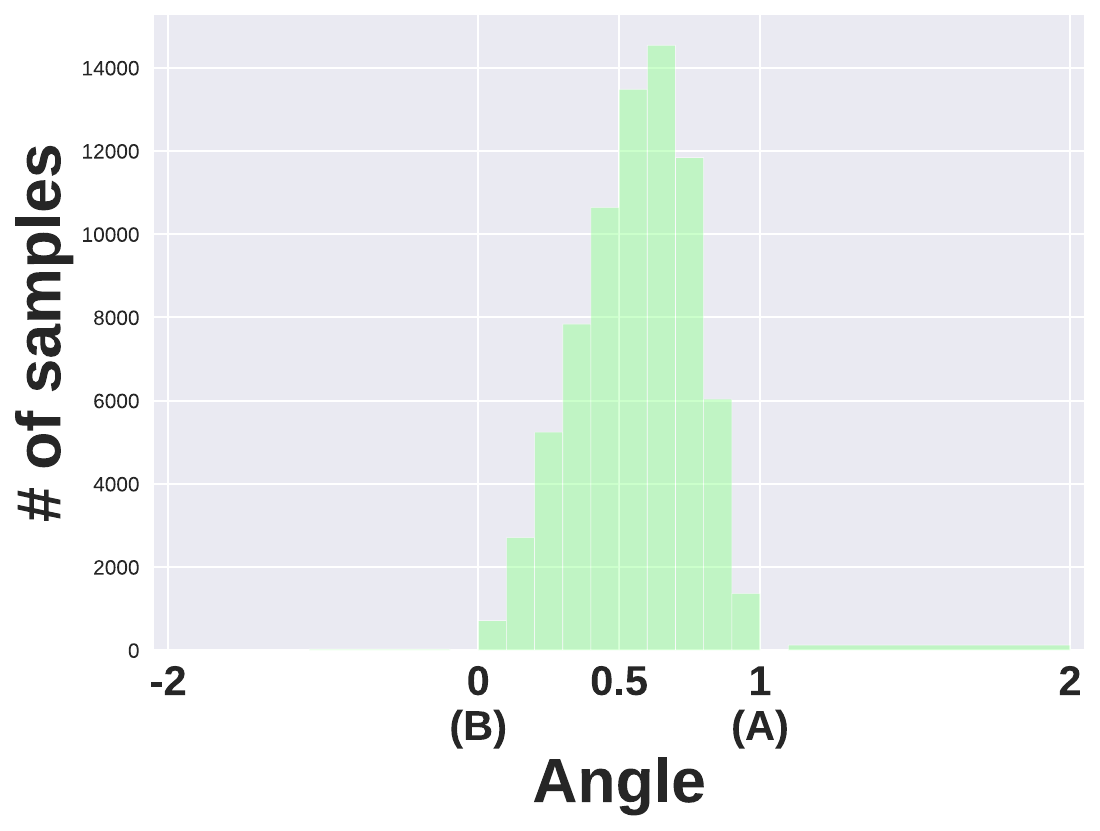}
         \label{fig:olmo_union}
     \end{subfigure}
     \vspace{-4mm}
     \subcaption{\textbf{OLMo}}
    \label{fig:olmo_projection}
    \end{subfigure}
     \hfill

     \begin{subfigure}[b]{1.0\textwidth}     \centering
    \begin{subfigure}[b]{0.32\textwidth}
         \centering
         \includegraphics[width=1.0\textwidth]{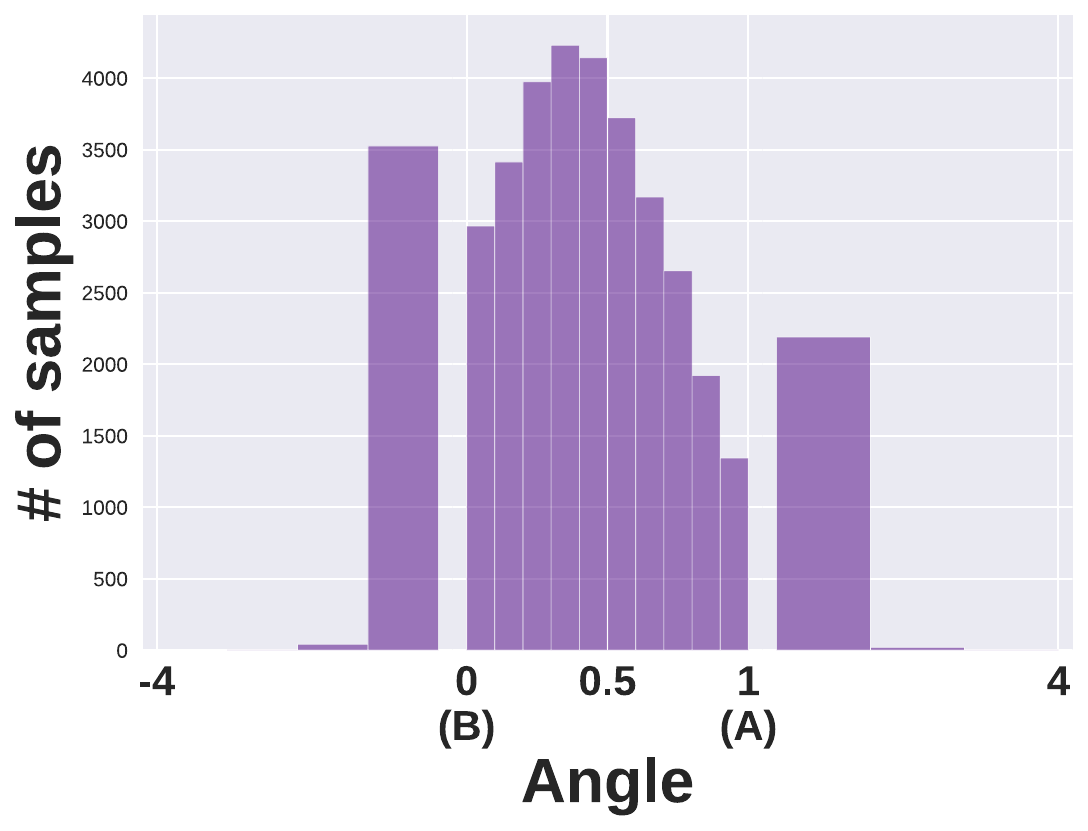}
         \label{fig:openelm_overlap}
     \end{subfigure}
    ~
     \begin{subfigure}[b]{0.32\textwidth}
         \centering
        \includegraphics[width=1.0\textwidth]{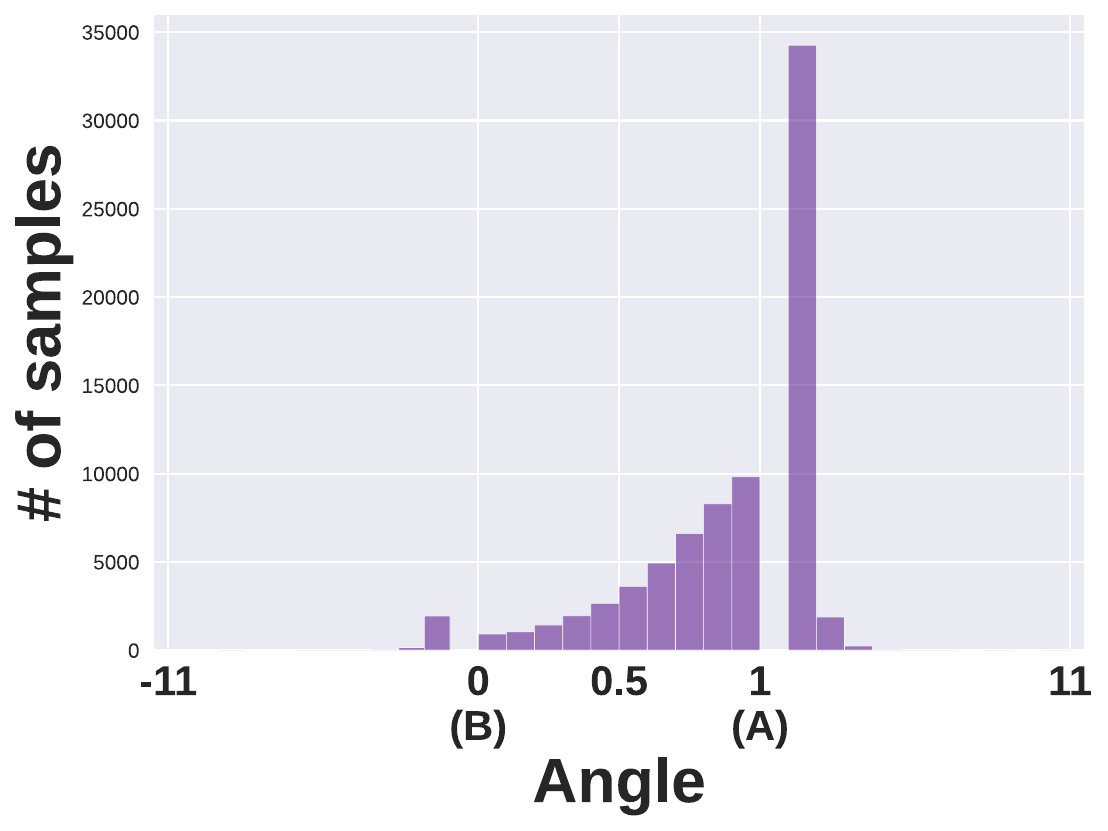}
         \label{fig:openelm_difference}
     \end{subfigure}
    ~
     \begin{subfigure}[b]{0.32\textwidth}
         \centering
         \includegraphics[width=\textwidth]{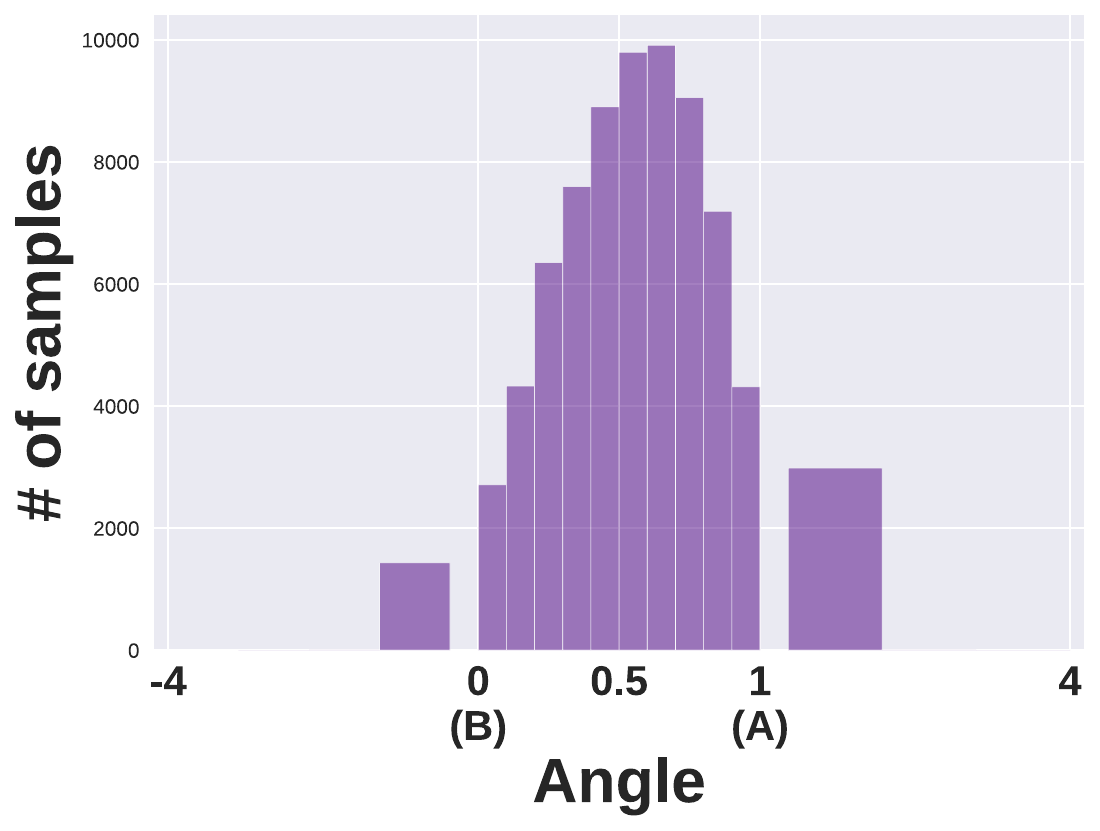}
         \label{fig:openelm_union}
     \end{subfigure}
     \vspace{-4mm}
     \subcaption{\textbf{OpenELM}}
    \label{fig:openelm_projection}
     \end{subfigure}

    \caption*{Figure~\ref{fig:all_projection_results}: Histogram of the angle between the projection of the target sentence embedding (onto the plane of the embeddings of the input sentences $A$ and $B$) and the embedding of sentence $B$. The target sentence embedding can be:\\
        \textbf{TextOverlap (Left)}:
        The projection embedding mostly lie in the ``middle'' of the embeddings of the input sentences as described in criterion C2 \ref{para:h2_results} and follows  our expectation shown in figure \ref{fig:projection_expectation}b \\
        \textbf{TextDifference (Middle)}: The projection embedding is mostly bounded by a small angle around the embedding of the input sentence $A$ (refer C5\ref{para:h5_results}).
        This follows our expectation shown in figure \ref{fig:projection_expectation}a \\
        \textbf{TextUnion (Right)}: The projection embedding mostly lie in ``middle'' of the input sentence embeddings (refer \ref{para:h6_results}). This follows  our expectation shown in figure \ref{fig:projection_expectation}c-middle.\\
        We normalize this angle such that the angle between the embeddings of sentences $A$ and $B$ is consistently 1 (refer \ref{para:h2_results} for details).}
\end{figure*}

\begin{figure*}[!t]\ContinuedFloat
    \centering

     \begin{subfigure}[b]{1.0\textwidth}     \centering
    \begin{subfigure}[b]{0.32\textwidth}
         \centering
         \begin{tikzpicture}
            \node[anchor=north] at (current bounding box.north) {\textbf{TextOverlap}};
        \end{tikzpicture}
         \includegraphics[width=1.0\textwidth]{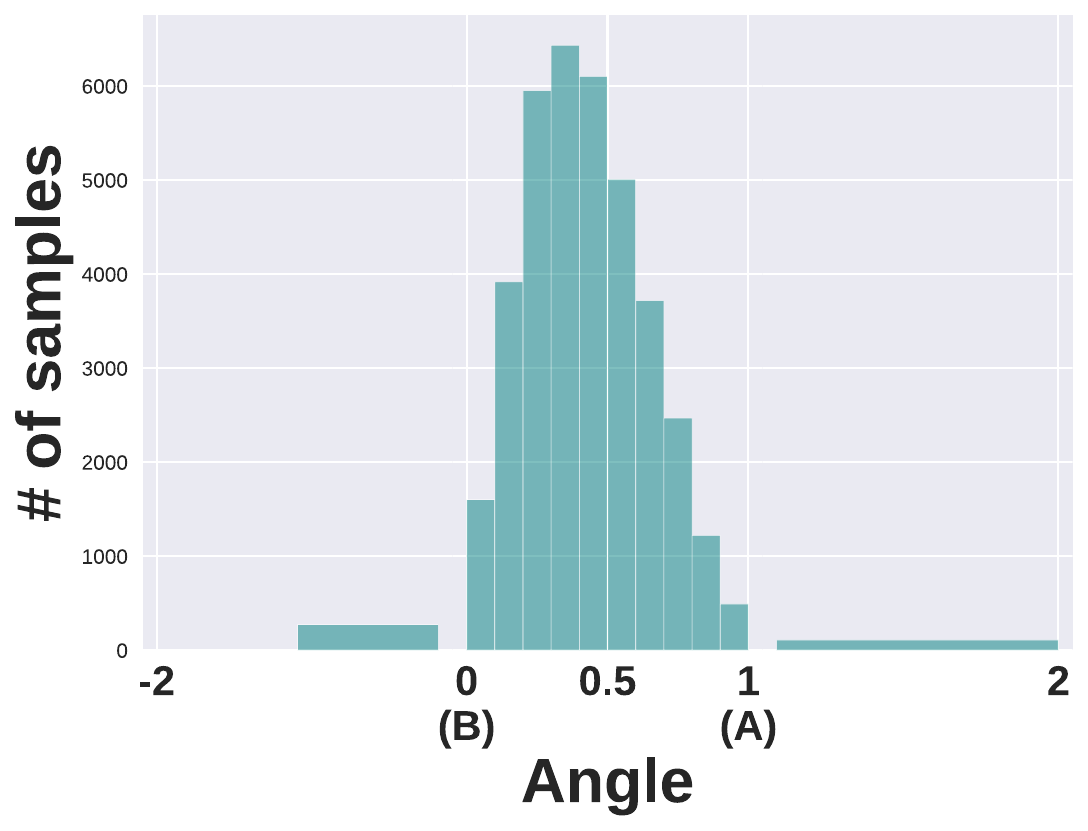}
         \label{fig:llama32_overlap}
     \end{subfigure}
    ~
     \begin{subfigure}[b]{0.32\textwidth}
         \centering
         \begin{tikzpicture}
            \node[anchor=north] at (current bounding box.north) {\textbf{TextDifference}};
        \end{tikzpicture}
        \includegraphics[width=1.0\textwidth]{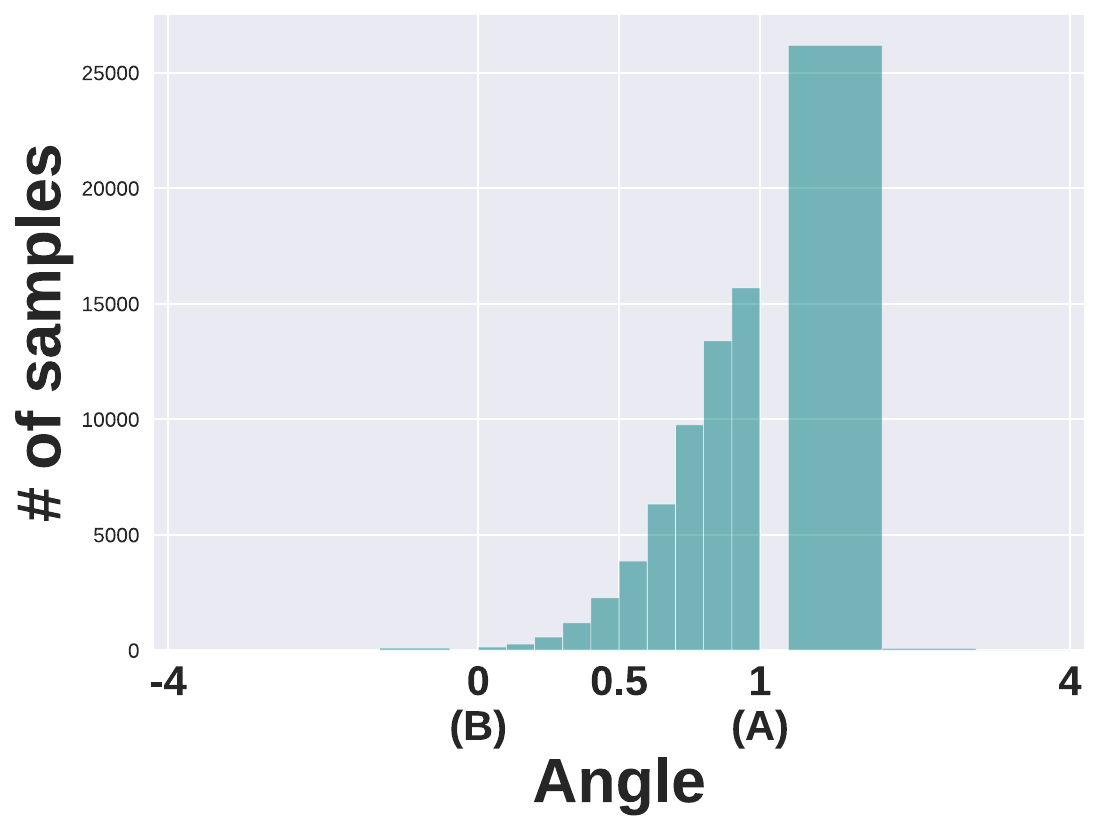}
         \label{fig:llama32_difference}
     \end{subfigure}
    ~
     \begin{subfigure}[b]{0.32\textwidth}
         \centering
         \begin{tikzpicture}
            \node[anchor=north] at (current bounding box.north) {\textbf{TextUnion}};
        \end{tikzpicture}
         \includegraphics[width=\textwidth]{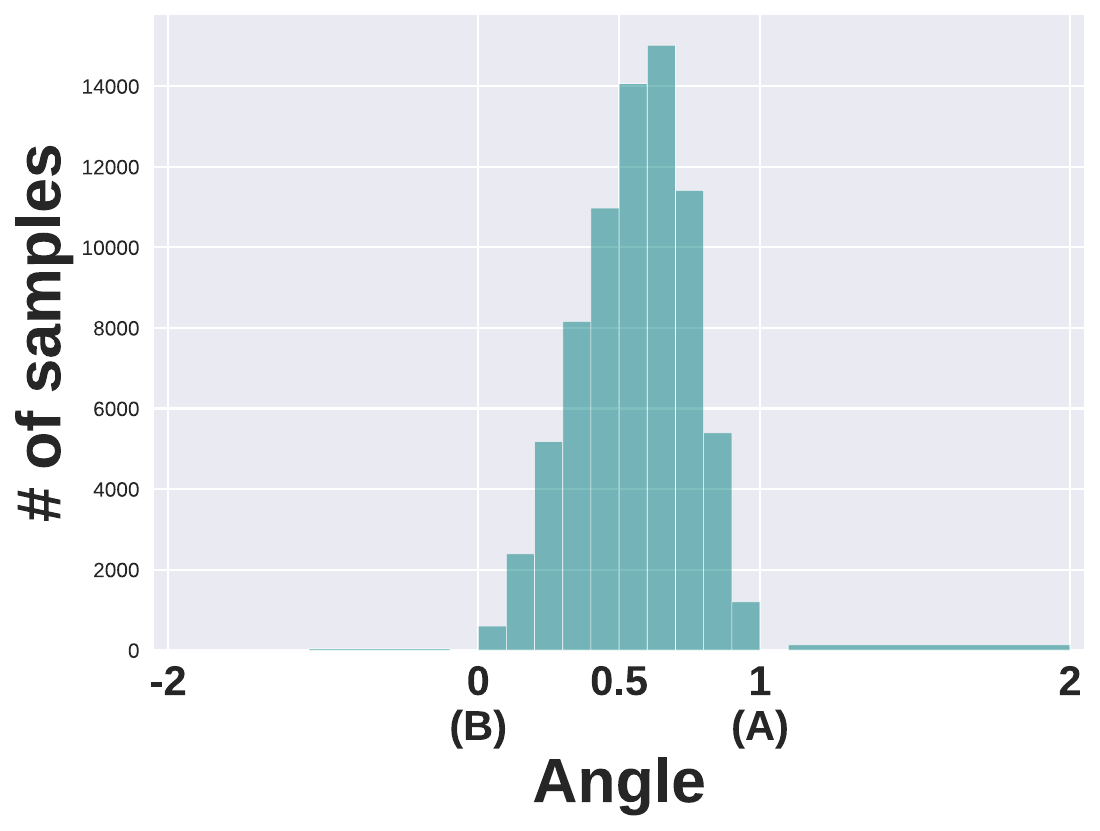}
         \label{fig:llama32_union}
     \end{subfigure}
     \vspace{-4mm}
     \subcaption{\textbf{LLaMA3.2}}
    \label{fig:llama32_projection}
     \end{subfigure}
     \hfill
     
    \begin{subfigure}[b]{1.0\textwidth}     \centering
    \begin{subfigure}[b]{0.32\textwidth}
         \centering
         \includegraphics[width=1.0\textwidth]{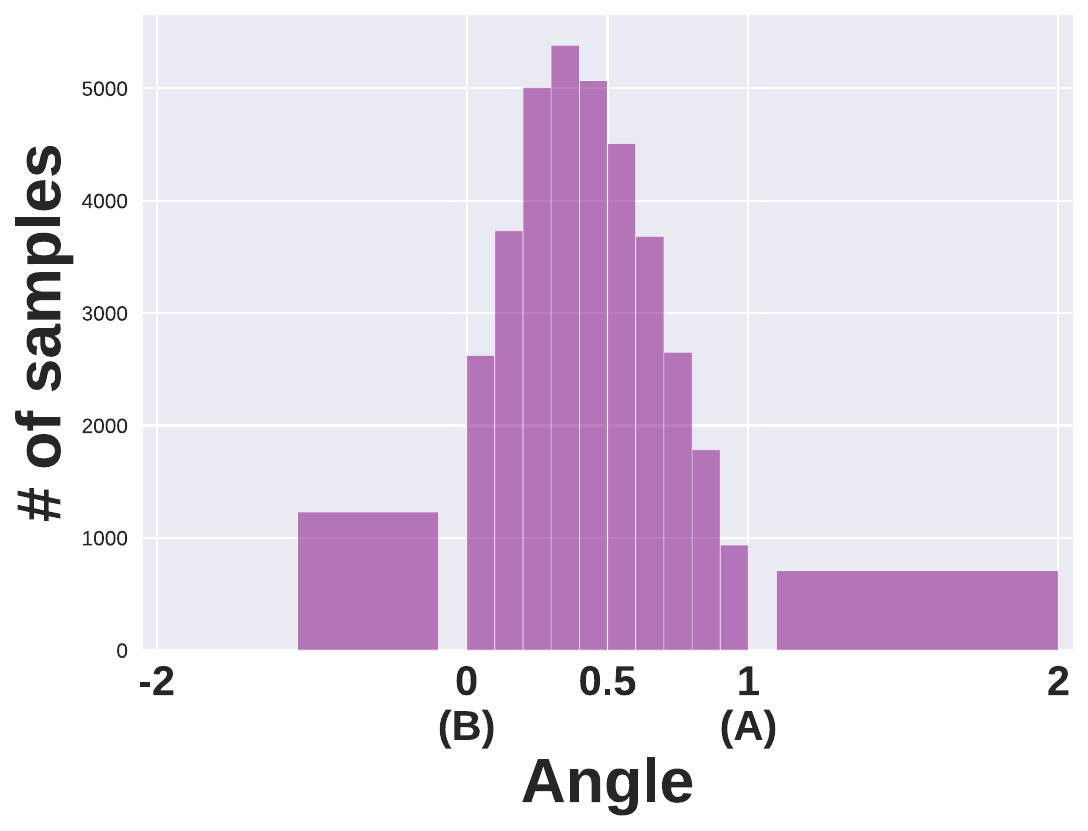}
         \label{fig:qwen_overlap}
     \end{subfigure}
    ~
     \begin{subfigure}[b]{0.32\textwidth}
         \centering
        \includegraphics[width=1.0\textwidth]{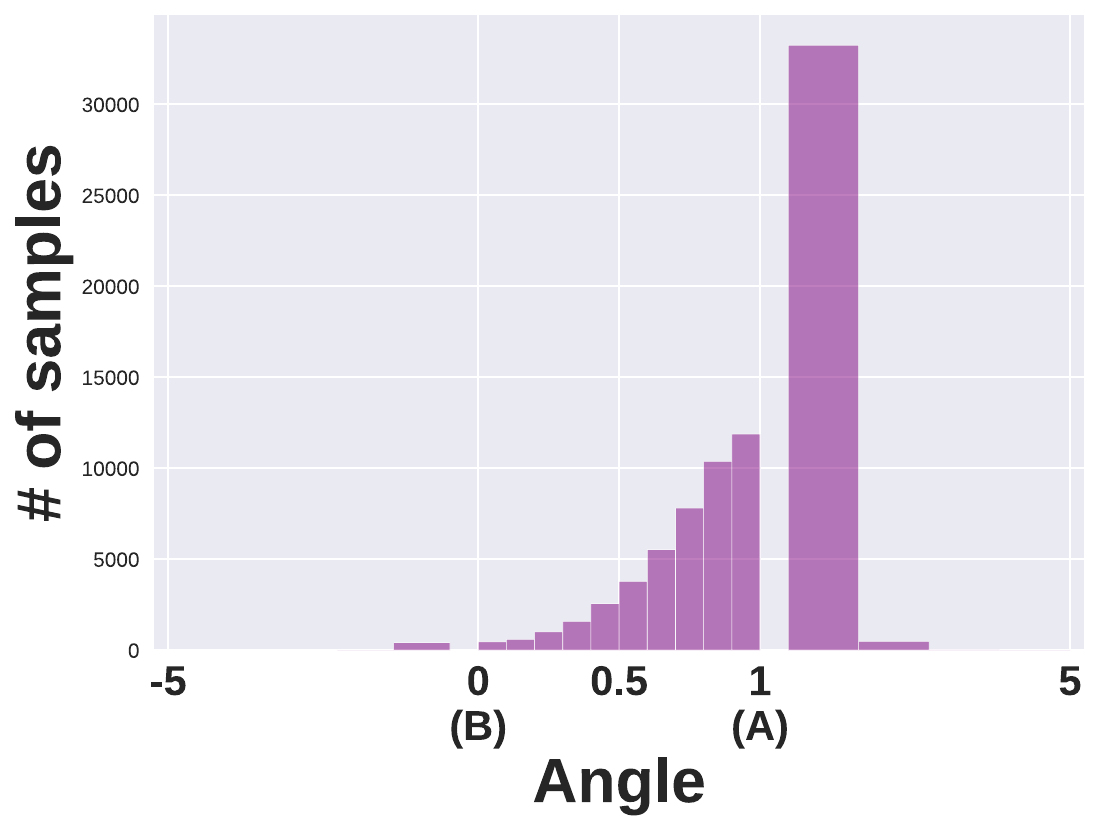}
         \label{fig:qwen_difference}
     \end{subfigure}
    ~
     \begin{subfigure}[b]{0.32\textwidth}
         \centering
         \includegraphics[width=\textwidth]{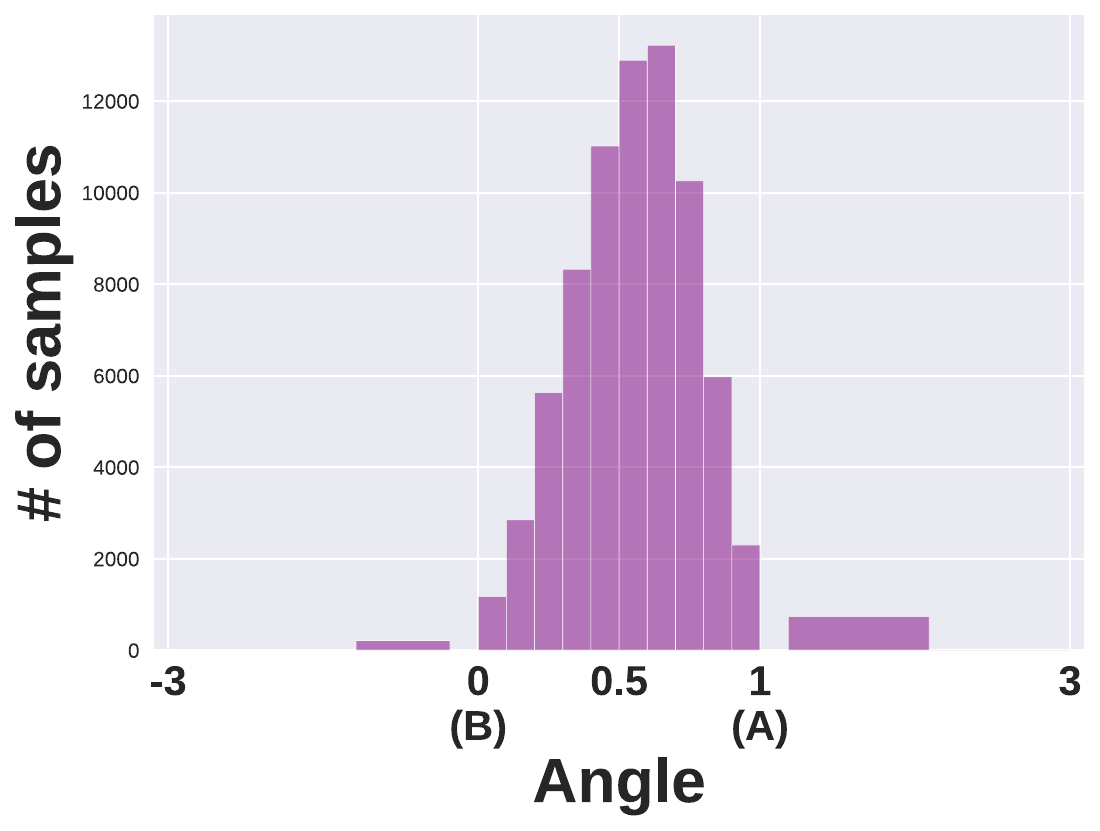}
         \label{fig:qwen_union}
     \end{subfigure}
     \vspace{-4mm}
     \subcaption{\textbf{Qwen}}
    \label{fig:qwen_projection}
     \end{subfigure}
     \hfill

    \begin{subfigure}[b]{1.0\textwidth}     \centering
    \begin{subfigure}[b]{0.32\textwidth}
         \centering
         \includegraphics[width=1.0\textwidth]{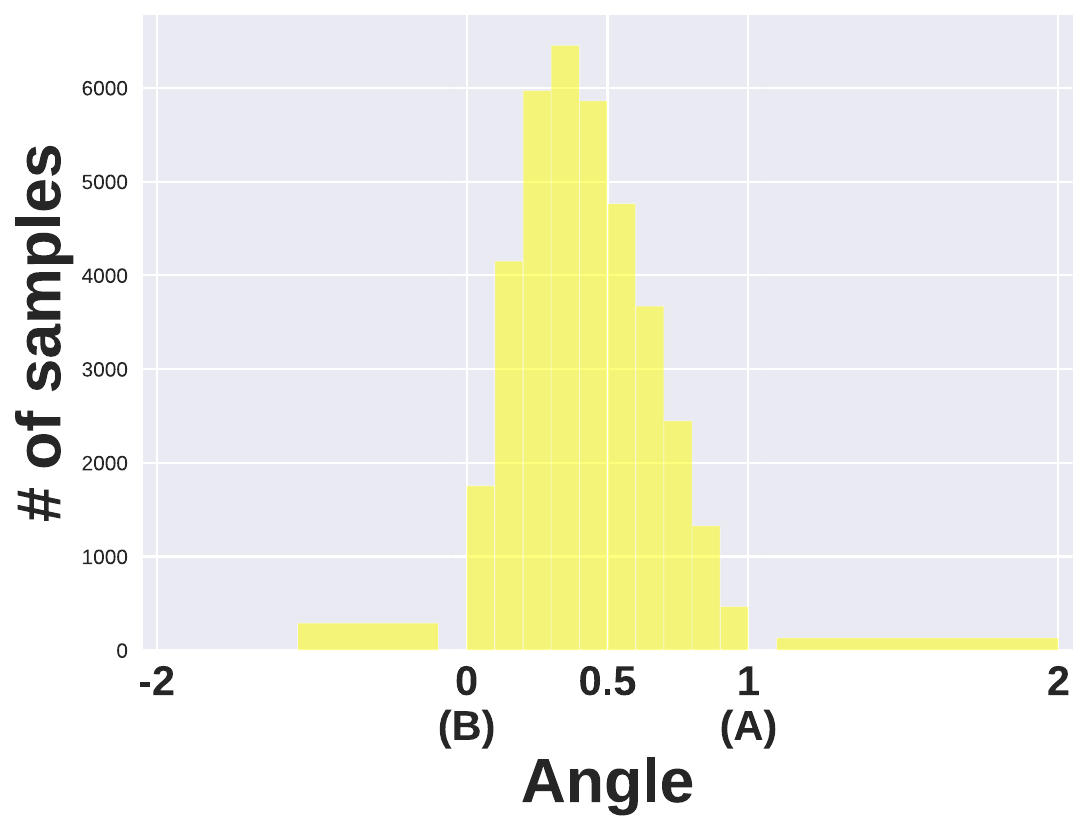}
         \label{fig:gemma_overlap}
     \end{subfigure}
    ~
     \begin{subfigure}[b]{0.32\textwidth}
         \centering
        \includegraphics[width=1.0\textwidth]{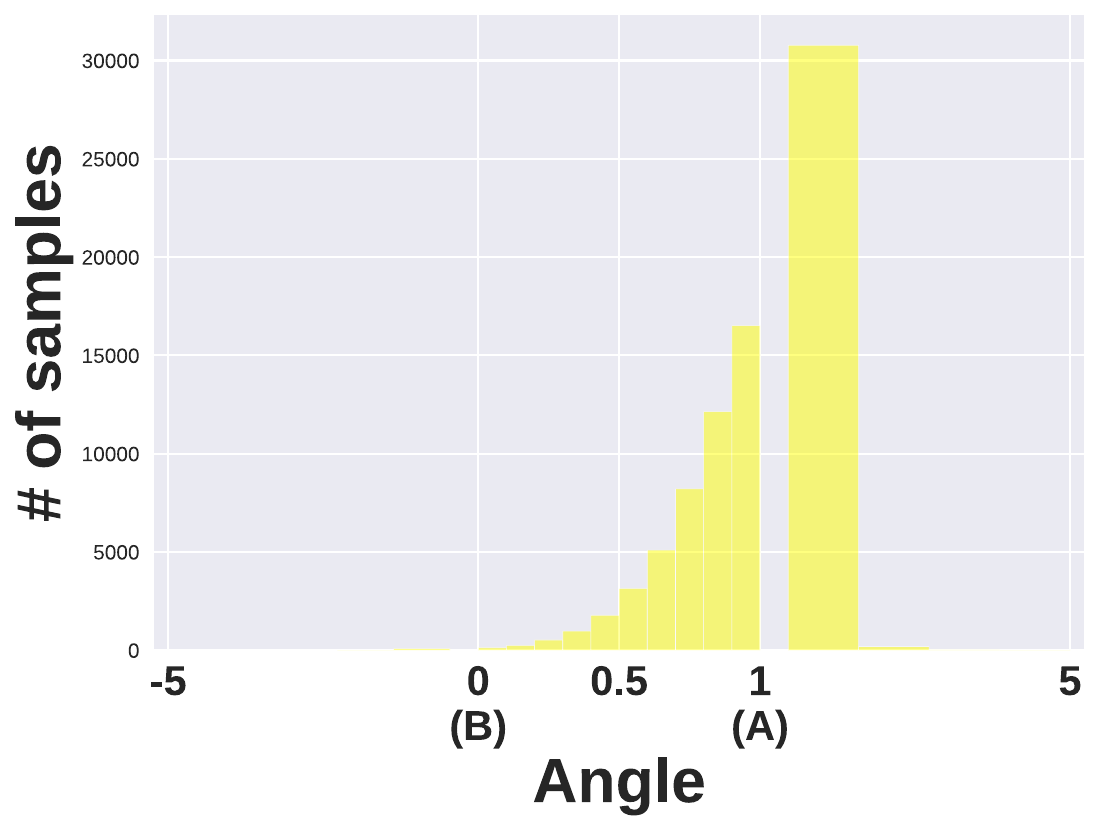}
         \label{fig:gemma_difference}
     \end{subfigure}
    ~
     \begin{subfigure}[b]{0.32\textwidth}
         \centering
         \includegraphics[width=\textwidth]{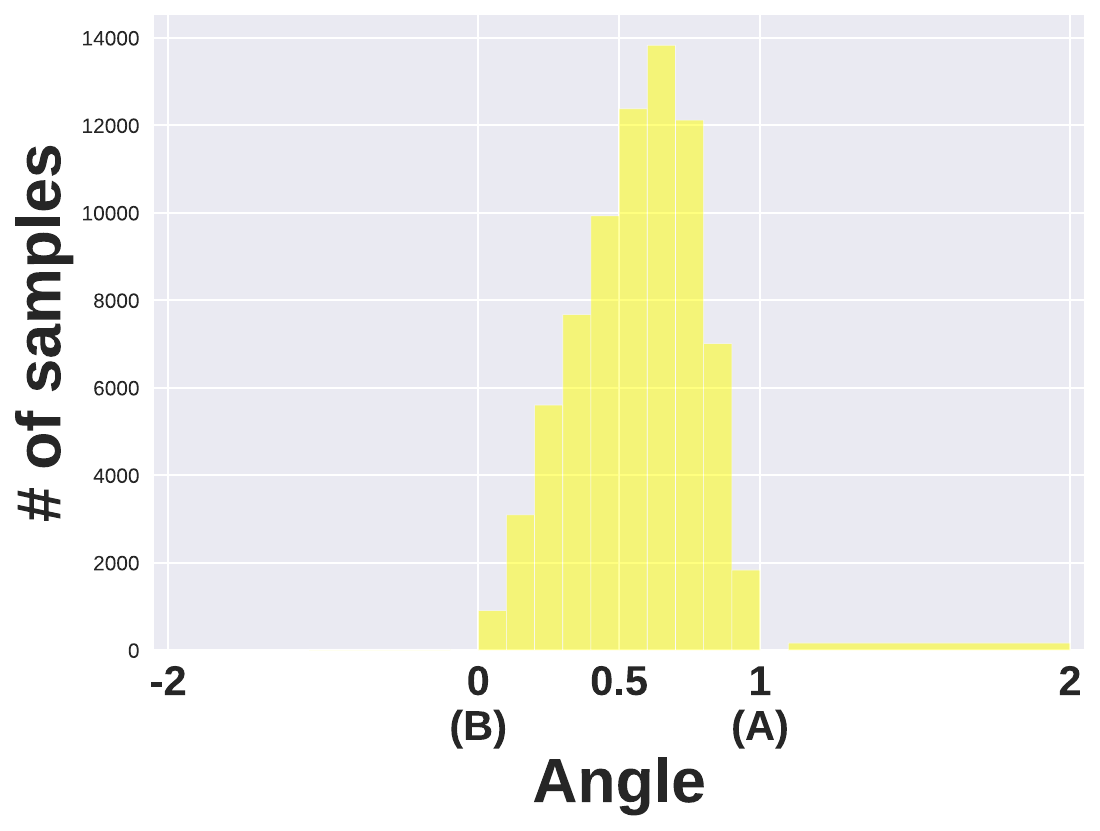}
         \label{fig:gemma_union}
     \end{subfigure}
     \vspace{-4mm}
     \subcaption{\textbf{Gemma}}
    \label{fig:gemma_projection}
    \end{subfigure}
    
    \caption*{Figure~\ref{fig:all_projection_results}: Histogram of the angle between the projection of the target sentence embedding (onto the plane of the embeddings of the input sentences $A$ and $B$) and the embedding of sentence $B$. The target sentence embedding can be:\\
        \textbf{TextOverlap (Left)}:
        The projection embedding mostly lie in the ``middle'' of the embeddings of the input sentences as described in criterion C2 \ref{para:h2_results} and follows  our expectation shown in figure \ref{fig:projection_expectation}b \\
        \textbf{TextDifference (Middle)}: The projection embedding is mostly bounded by a small angle around the embedding of the input sentence $A$ (refer C5\ref{para:h5_results}).
        This follows our expectation shown in figure \ref{fig:projection_expectation}a \\
        \textbf{TextUnion (Right)}: The projection embedding mostly lie in ``middle'' of the input sentence embeddings (refer \ref{para:h6_results}). This follows  our expectation shown in figure \ref{fig:projection_expectation}c-middle.\\
        We normalize this angle such that the angle between the embeddings of sentences $A$ and $B$ is consistently 1 (refer \ref{para:h2_results} for details).}
\end{figure*}

\clearpage

\begin{figure*}[h]
    \centering
    \begin{subfigure}[b]{0.24\textwidth}
        \includegraphics[width=\textwidth]{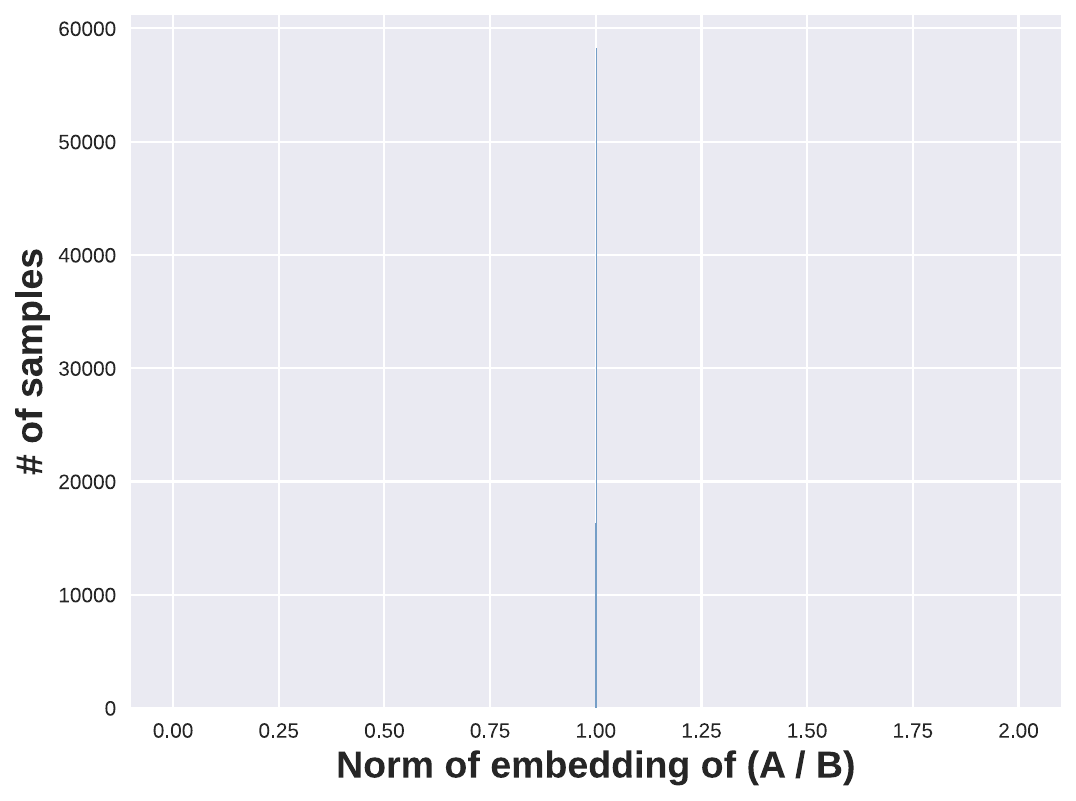}
        \subcaption{SBERT-mini}
    \end{subfigure}
    \hfill
    \begin{subfigure}[b]{0.24\textwidth}
        \includegraphics[width=\textwidth]{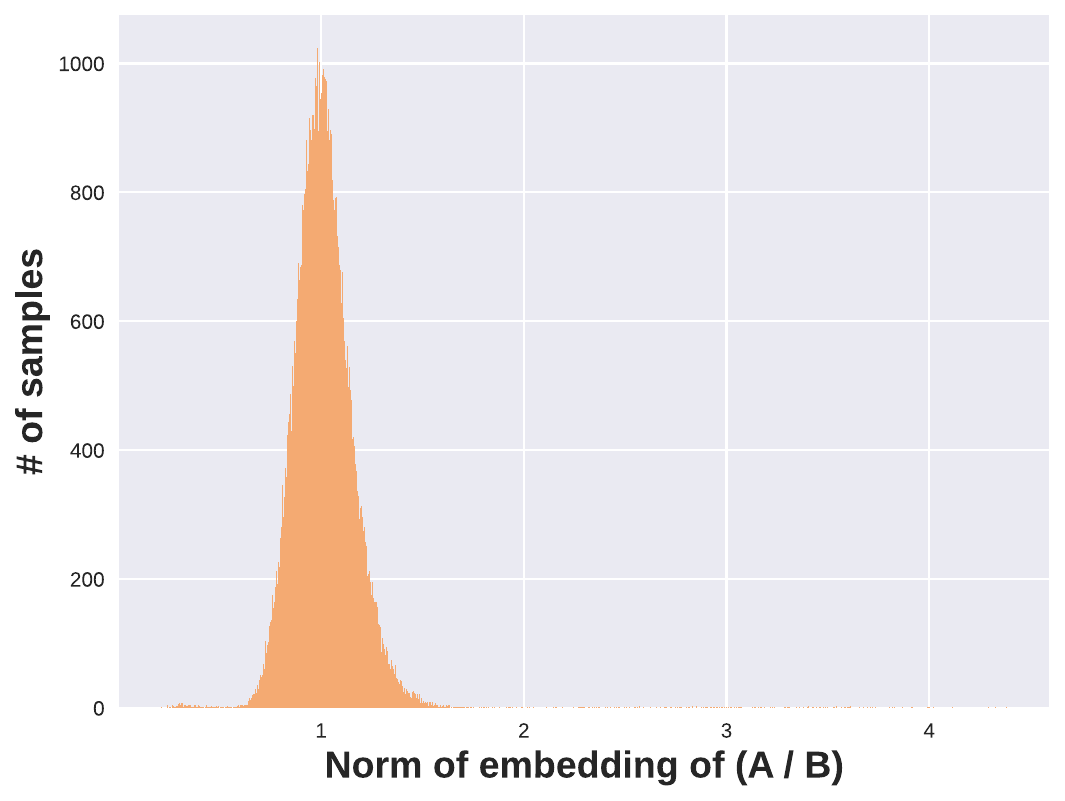}
        \subcaption{LASER}
    \end{subfigure}
    \hfill
    \begin{subfigure}[b]{0.24\textwidth}
        \includegraphics[width=\textwidth]{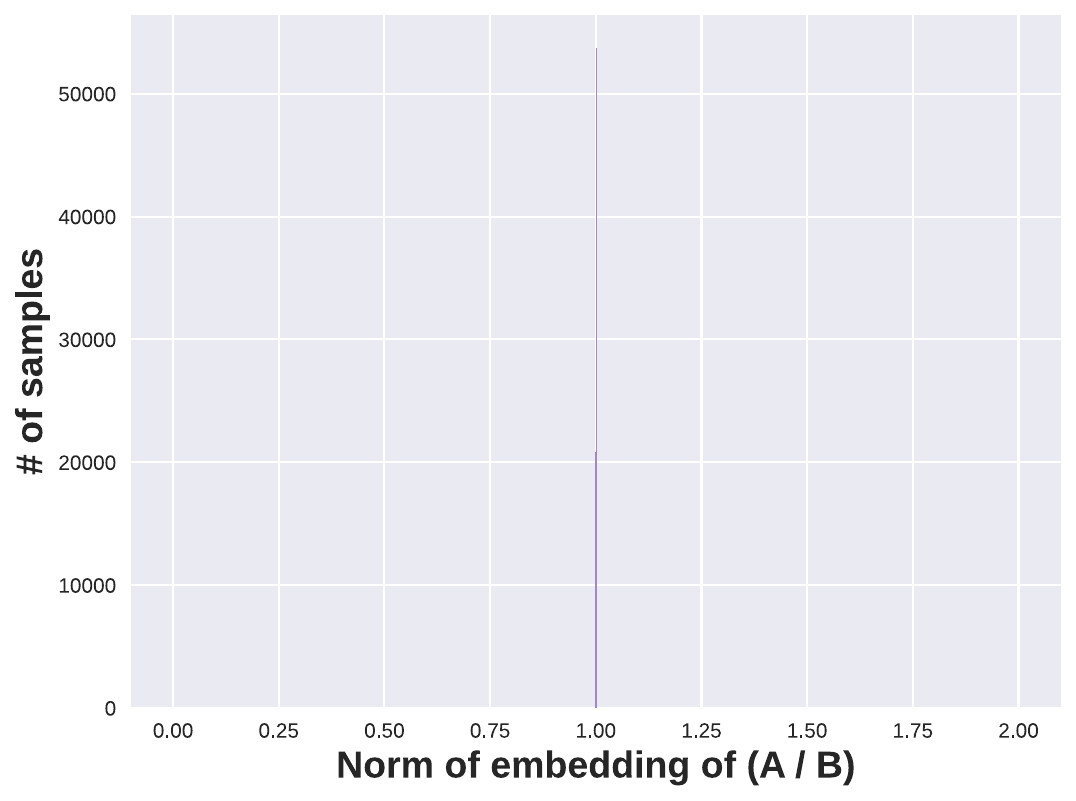}
        \subcaption{USE}
    \end{subfigure}
    \hfill
    \begin{subfigure}[b]{0.24\textwidth}
        \includegraphics[width=\textwidth]{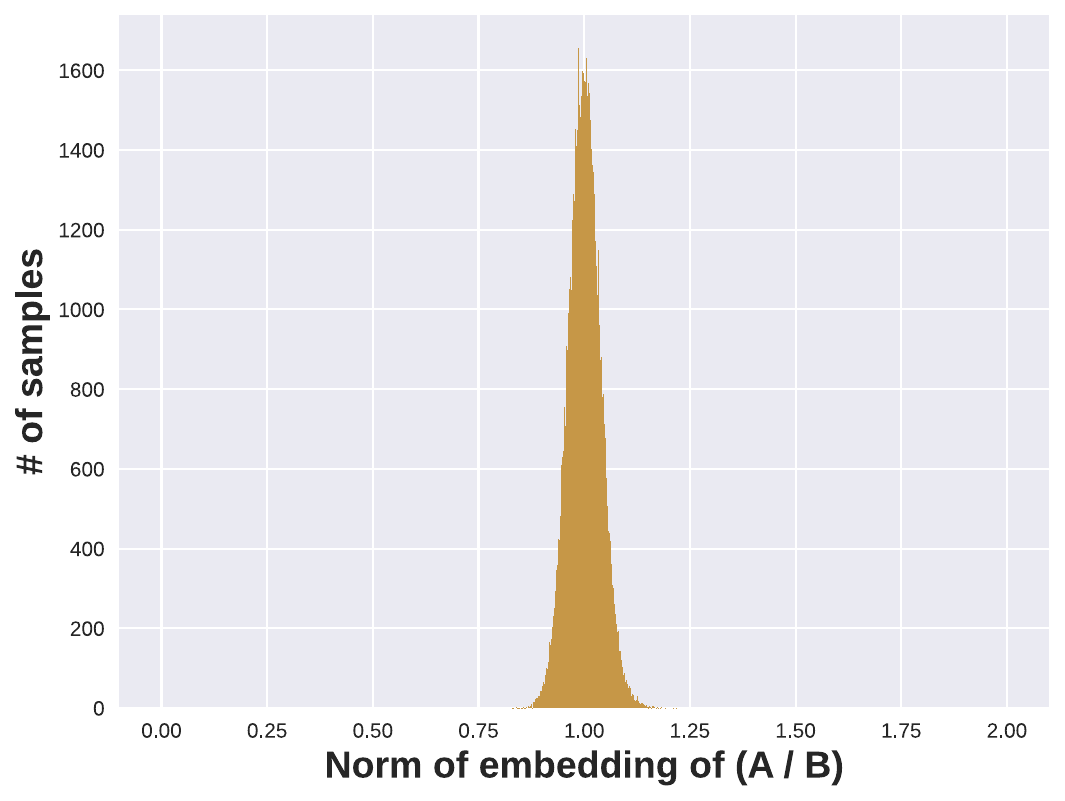}
        \subcaption{RoBERTa}
    \end{subfigure}
    
    \begin{subfigure}[b]{0.24\textwidth}
        \includegraphics[width=\textwidth]{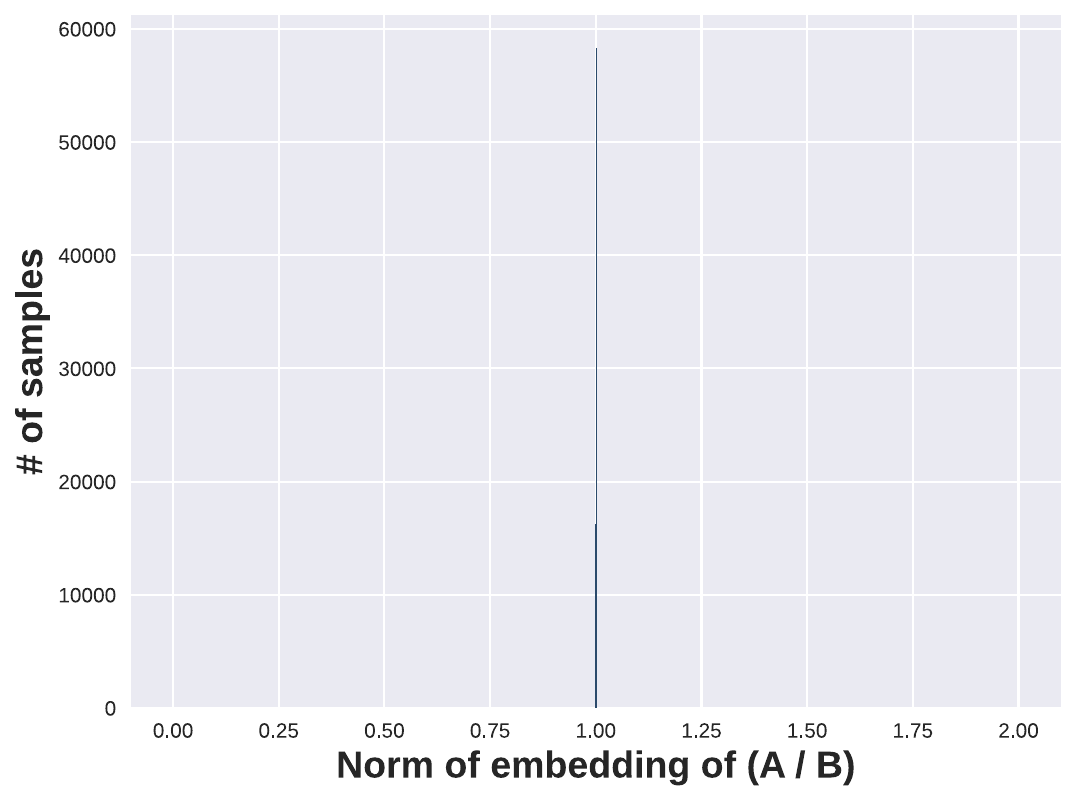}
        \subcaption{SBERT-L}
    \end{subfigure}
    \hfill
    \begin{subfigure}[b]{0.24\textwidth}
        \includegraphics[width=\textwidth]{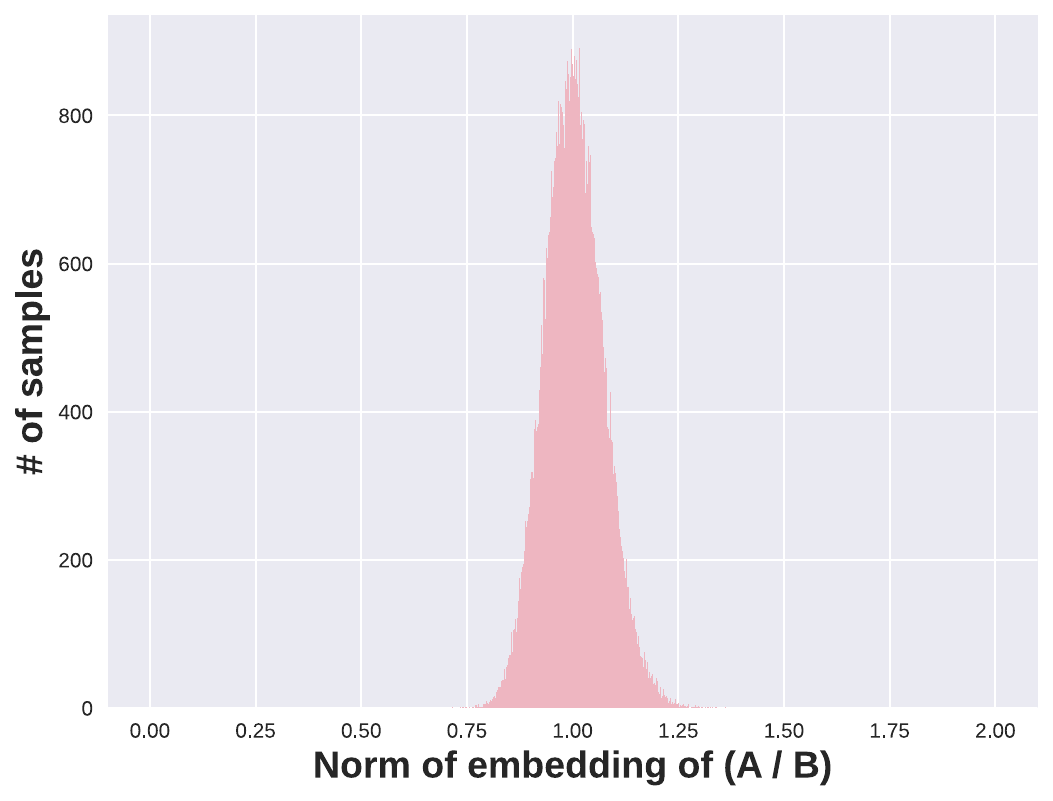}
        \subcaption{SimCSE}
    \end{subfigure}
    \hfill
    \begin{subfigure}[b]{0.24\textwidth}
        \includegraphics[width=\textwidth]{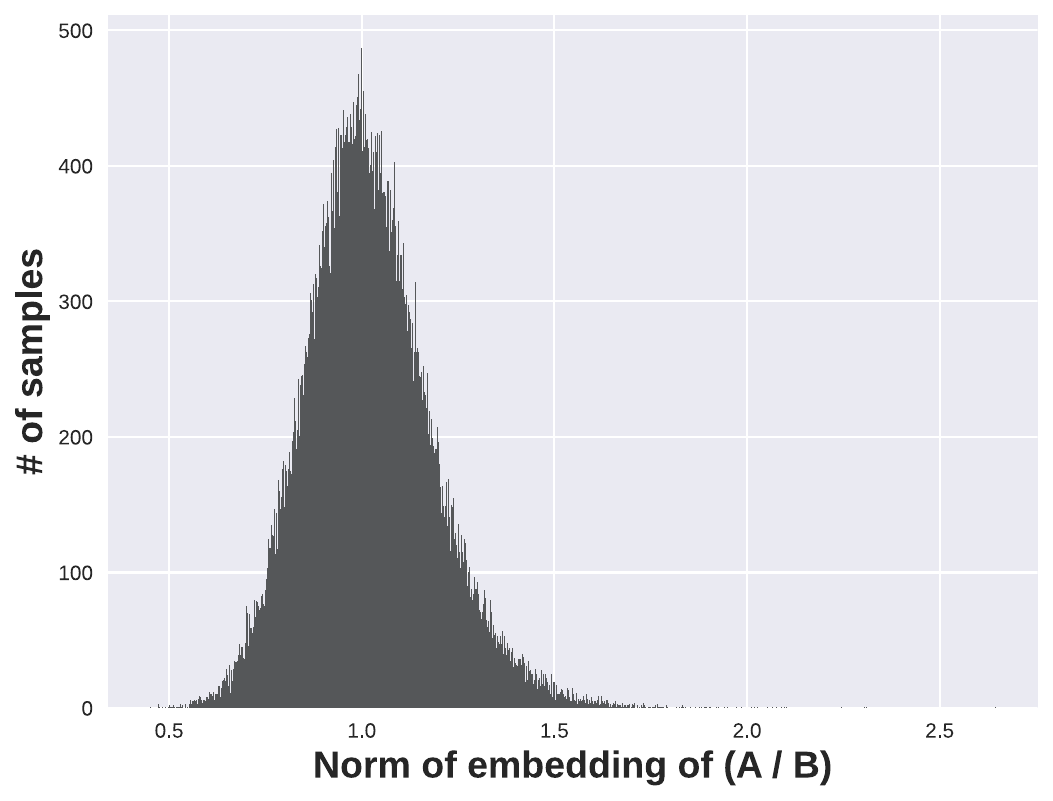}
        \subcaption{InferSent}
    \end{subfigure}
    \hfill
    \begin{subfigure}[b]{0.24\textwidth}
        \includegraphics[width=\textwidth]{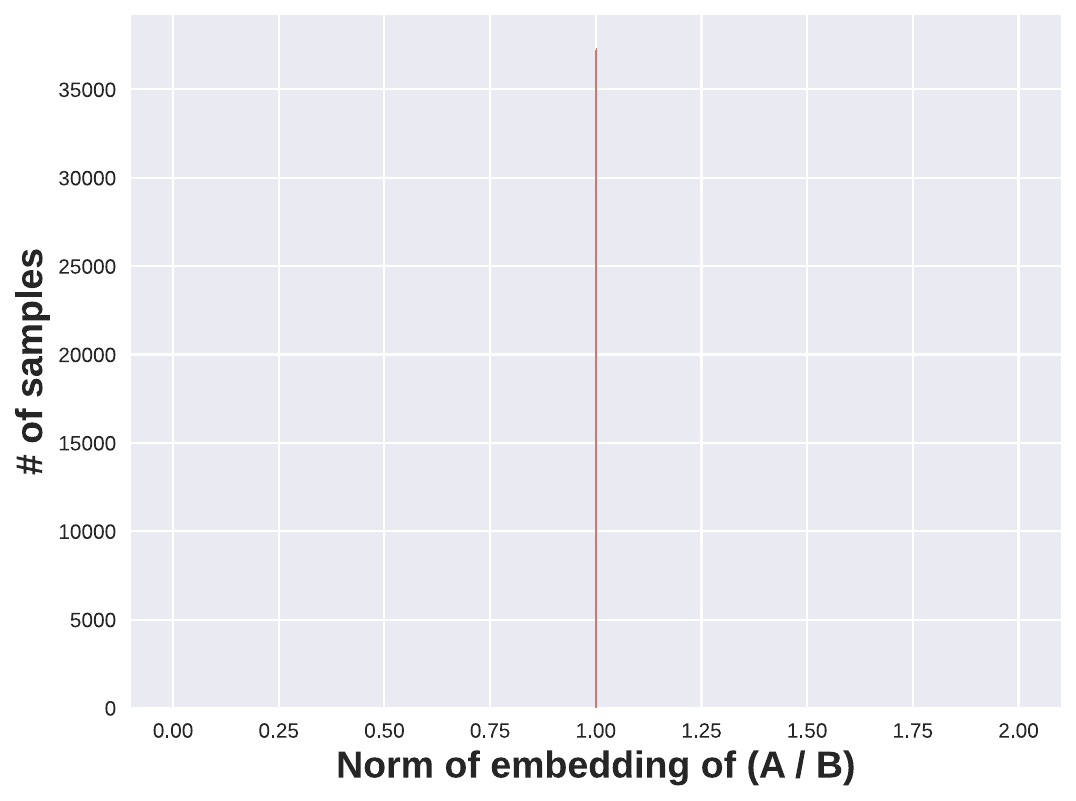}
        \subcaption{GPT3}
    \end{subfigure}
    
    \begin{subfigure}[b]{0.24\textwidth}
        \includegraphics[width=\textwidth]{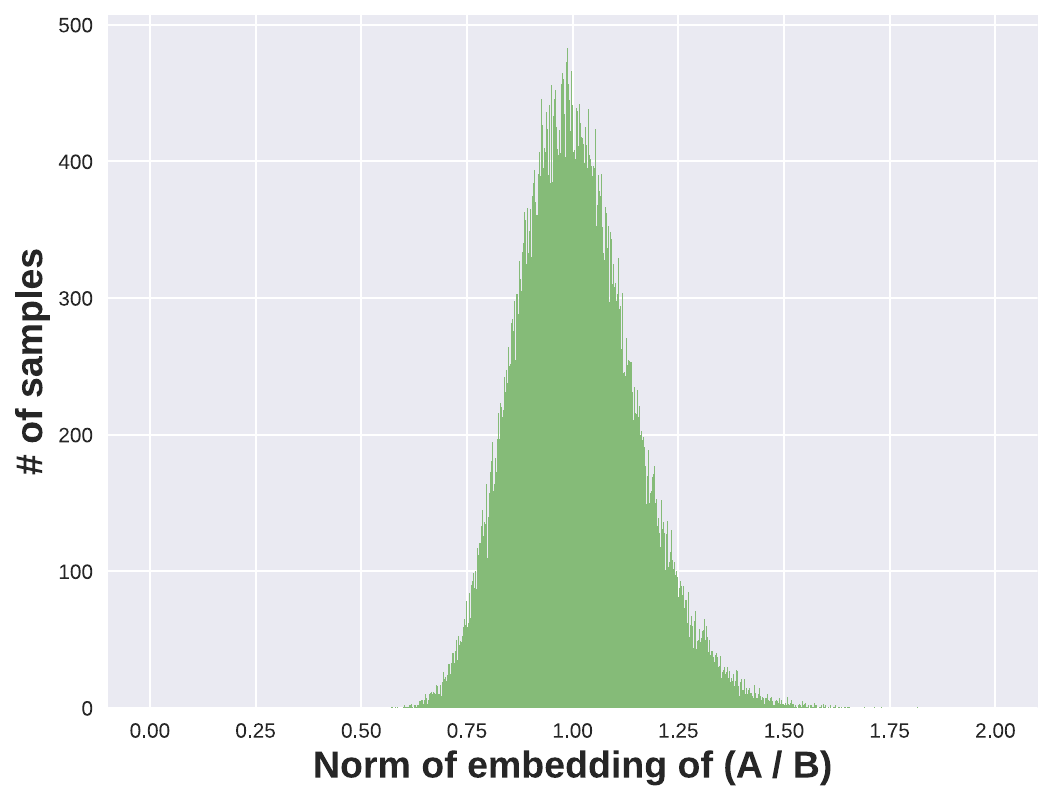}
        \subcaption{LLaMA2}
    \end{subfigure}
    \hfill
    \begin{subfigure}[b]{0.24\textwidth}
        \includegraphics[width=\textwidth]{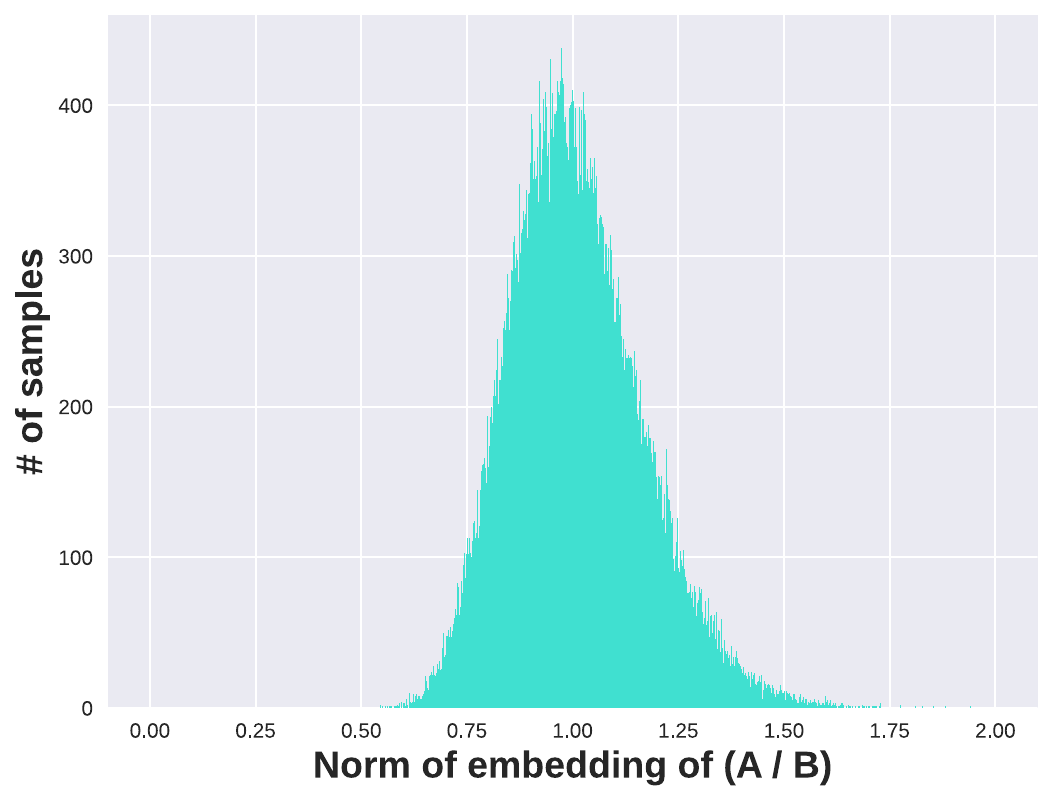}
        \subcaption{Mistral}
    \end{subfigure}
    \hfill
    \begin{subfigure}[b]{0.24\textwidth}
        \includegraphics[width=\textwidth]{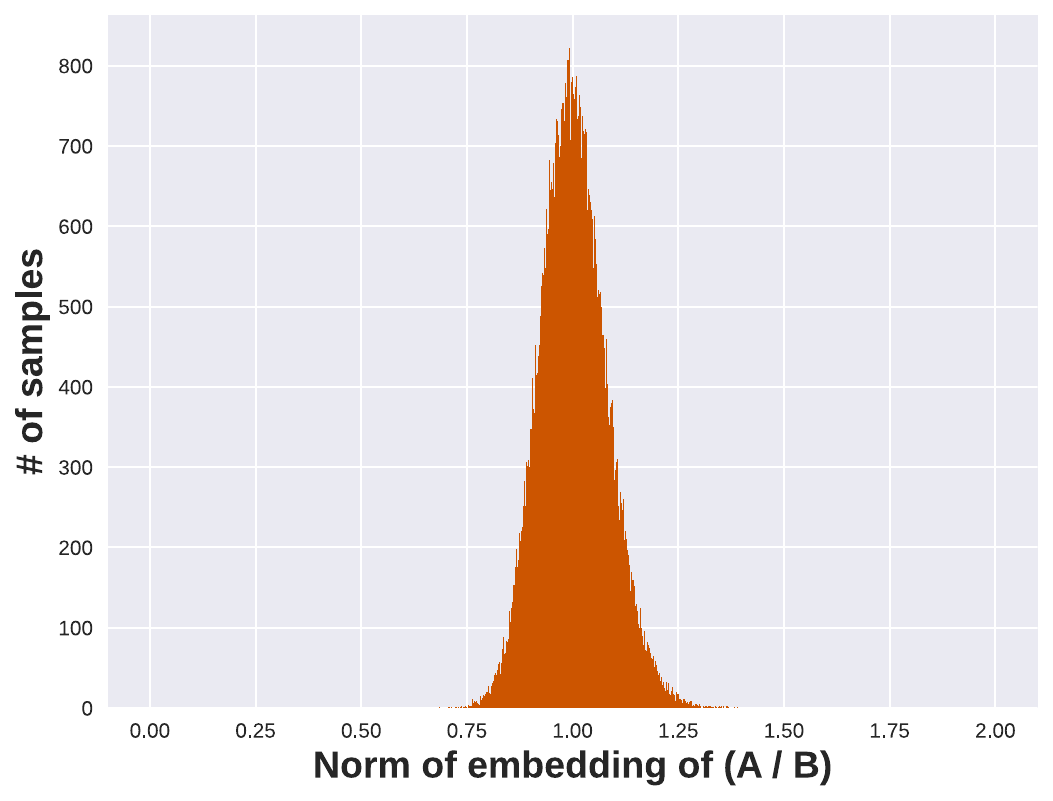}
        \subcaption{LLaMA3}
    \end{subfigure}
    \hfill
    \begin{subfigure}[b]{0.24\textwidth}
        \includegraphics[width=\textwidth]{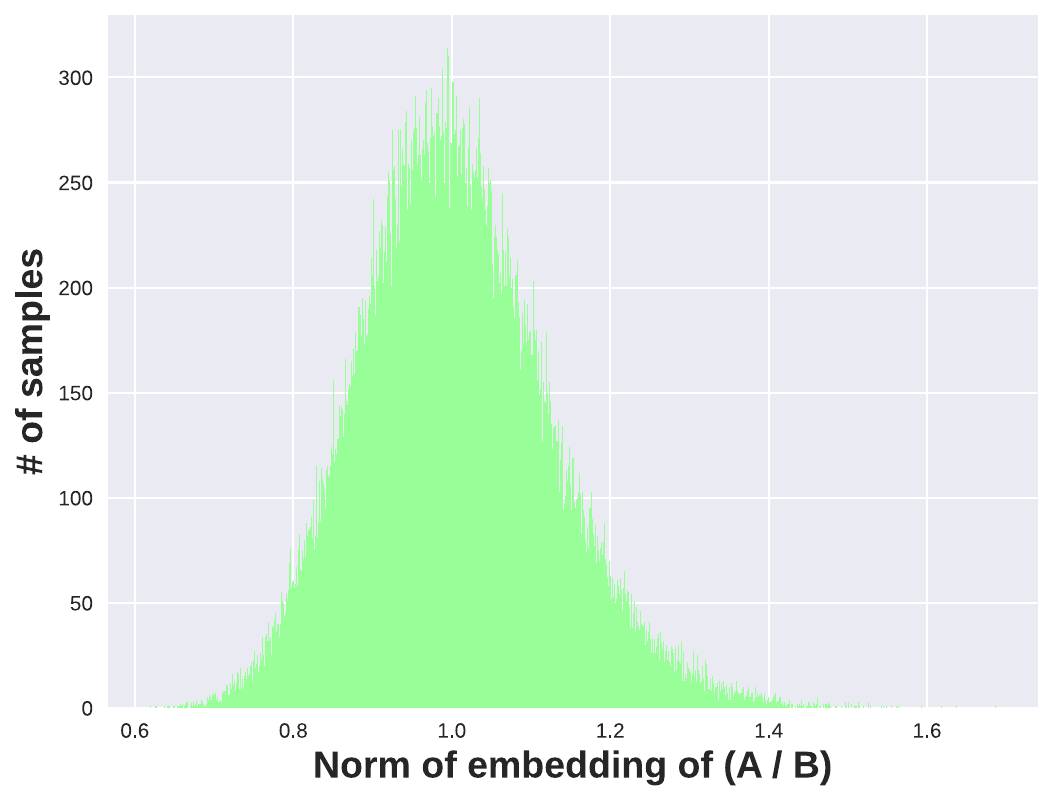}
        \subcaption{OLMo}
    \end{subfigure}
    
    \begin{subfigure}[b]{0.24\textwidth}
        \includegraphics[width=\textwidth]{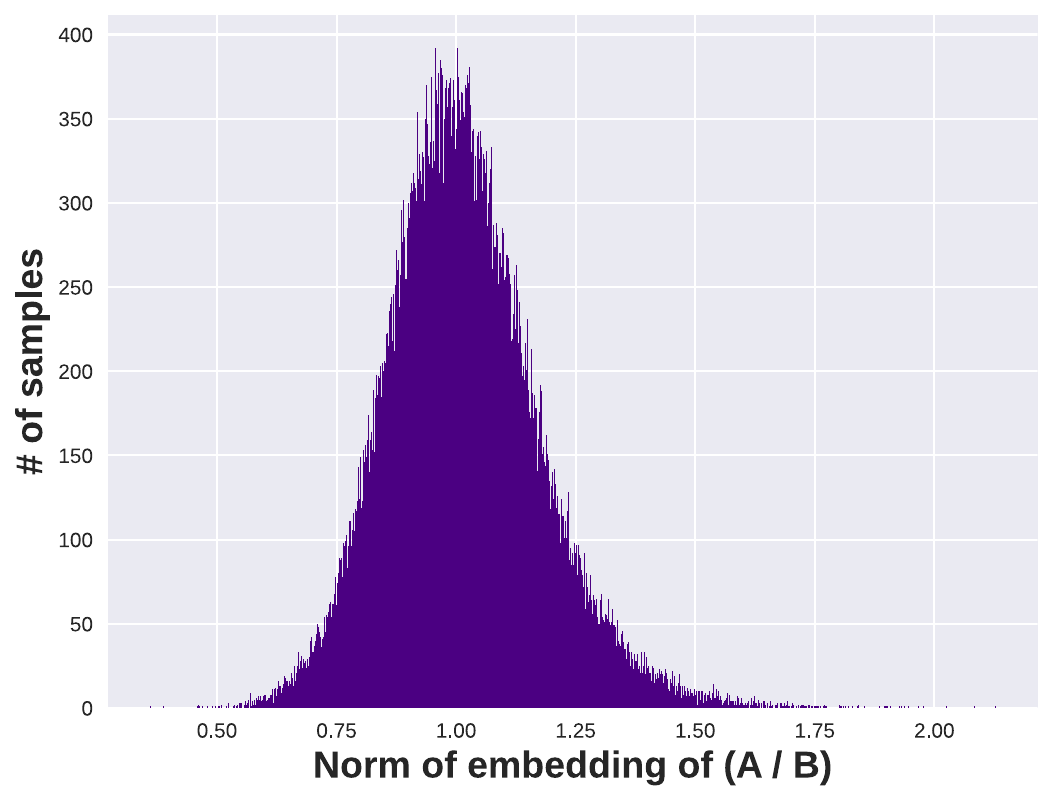}
        \subcaption{OpenELM}
    \end{subfigure}
    \hfill
    \begin{subfigure}[b]{0.24\textwidth}
        \includegraphics[width=\textwidth]{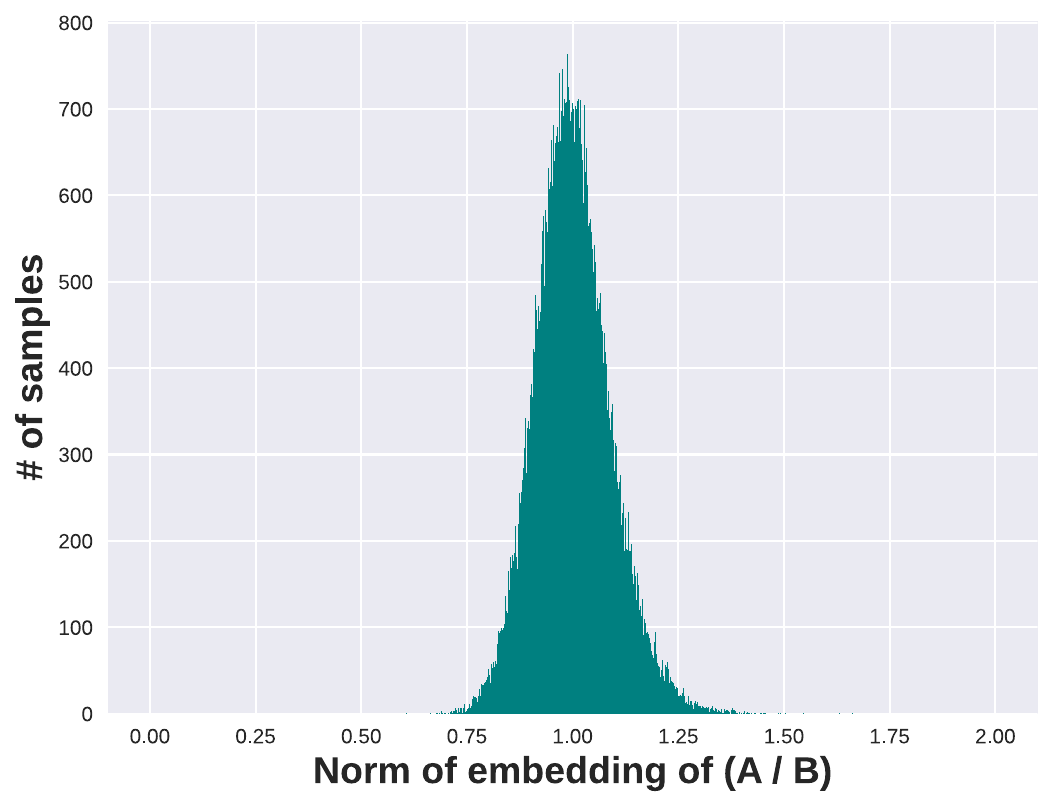}
        \subcaption{LLaMA3.2}
    \end{subfigure}
    \hfill
    \begin{subfigure}[b]{0.24\textwidth}
        \includegraphics[width=\textwidth]{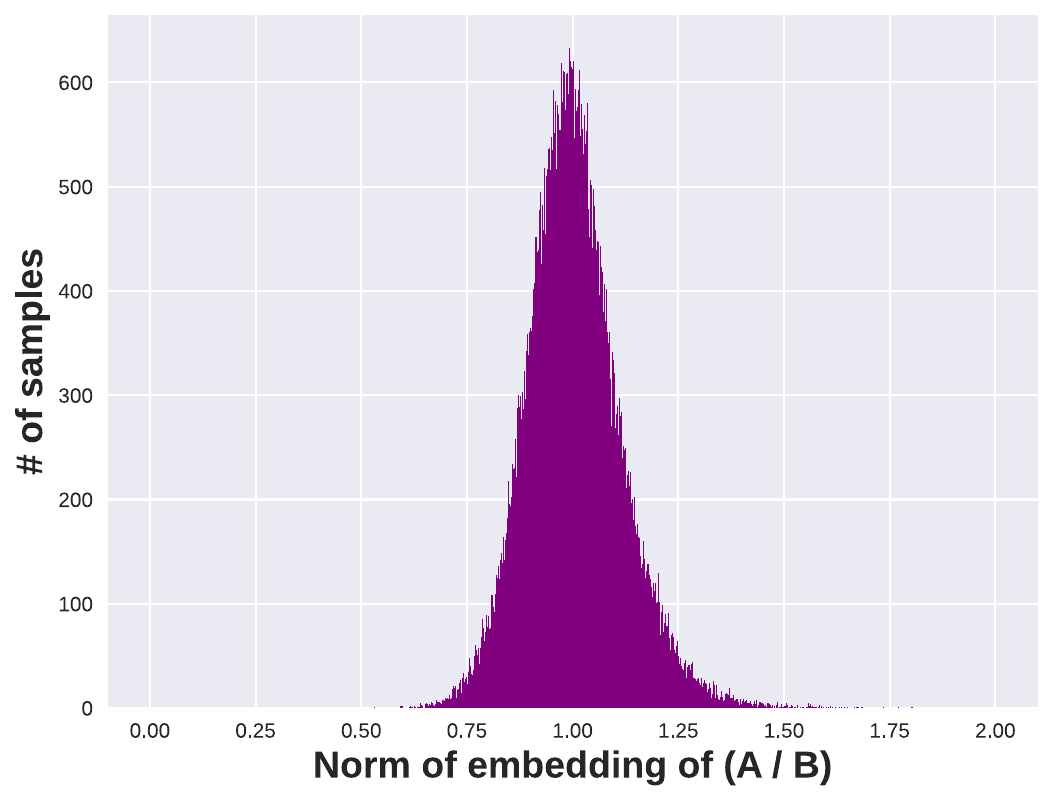}
        \subcaption{Qwen}
    \end{subfigure}
    \hfill
    \begin{subfigure}[b]{0.24\textwidth}
        \includegraphics[width=\textwidth]{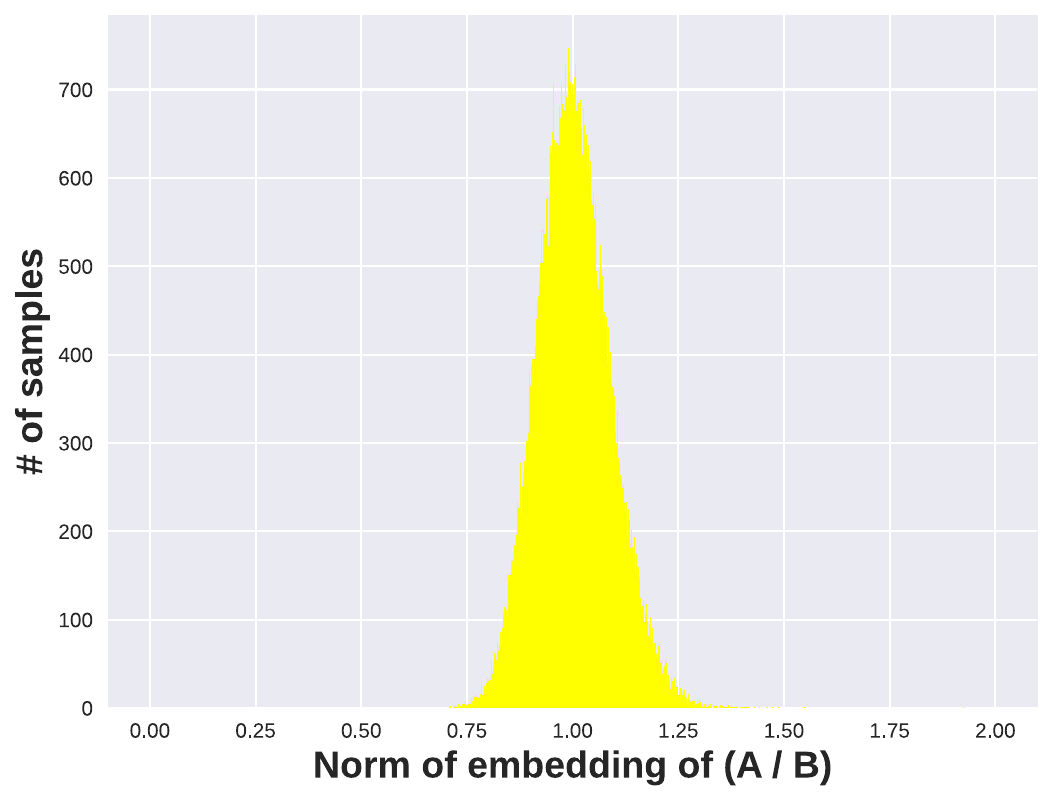}
        \subcaption{Gemma}
    \end{subfigure}

    \caption{Histogram of the embedding norm of sentence A ($E_A$) by the norm of the embedding of sentence B ($E_B$). For details, refer to Section \ref{para:h6} and \ref{para:h6_results}.
    }
    \label{fig:combined_norm_plots}
\end{figure*}

\clearpage





\section*{Supplementary material}

# Annotation Guidelines

## Problem Statement of the Project:

Develop a ML model for the following tasks -

**Intersection:** Given two input sentences, we have to figure out common information between them.
**Difference:** Given sentence 1 and sentence 2, we have to find the information that is present in sentence 1 but not in sentence 2.
**Union:** Given sentence 1 and sentence 2, we have to find the information that is present in either sentence 1 or sentence 2 or both of them.

Here are some examples for each task –

| Task | Sentence 1 | Sentence 2 | Output |
|---|---|---|---|
| Intersection | John decided to go school after his New York vacation. | John is back to school after Thanksgiving | John is back to school |
| Difference | | | John spent his vacations in New York |
| Union | | | John went back to school after spending his Thanksgiving vacations in New York. |

## Annotation Task:

You will be given one excel file corresponding to one of the above-mentioned tasks. Each file has 100 samples (100 rows) in the following format:
**S1:** Sentence 1
**S2:** Sentence 2
**Output:** Output sentence for the corresponding problem/task
**Annotation: ?**

Your job is to annotate these samples on a Likert scale showing whether the `Output` is the correct output for the corresponding problem with regards to the input sentences (`S1` and `S2`). For the Likert scale, you'll annotate as follows:
Strongly Disagree: 0
Disagree: 1
Neutral: 2

Agree: 3
Strongly Agree: 4
(Only use numbers between 0-4 inclusive.)

For example, you are assigned the file, `Union.xlsx`. So, your task is to evaluate the samples for the `Union` problem. So given a sample, if you feel the `Output` is the `Union` of sentences `S1` and `S2`, and you `Agree` with it, you mark it as `3` (Agree) in the `Annotation` column.




\end{document}